\pdfoutput=1

\documentclass[10pt,twocolumn,twoside,journal]{IEEEtran}

\usepackage{amsmath,epsfig}
\usepackage{times}
\usepackage{amssymb}
\usepackage{verbatim}

\usepackage[lined, ruled]{algorithm2e}
\usepackage{multirow}
\usepackage{subfigure}
\usepackage{graphicx,cs,mathsymb,xspace,colortbl} 
\usepackage{sidecap}

\def\x{{\mathbf x}}
\def\y{{\mathbf y}}

\graphicspath{{./}{./Figs/}}

\usepackage[sort&compress]{natbib}
\usepackage[scriptsize]{caption}
\usepackage{bibspacing}

\bibpunct{[}{]}{,\!}{n}{}{;}

\def\CSAcknowledgement{NICTA is funded by the Australian Government as represented by the
Department of Broadband, Communications and the Digital Economy and the
Australian Research Council through the ICT Center of Excellence program.
}

\begin{document}

\title{Real-time Visual Tracking Using Sparse Representation}

\author{Hanxi Li, Chunhua Shen, and Qinfeng Shi
\thanks
{
H. Li and C. Shen is with NICTA, Canberra Research Laboratory,
Canberra, ACT 2601, Australia,
and also with the Australian National University, Canberra,
ACT 0200, Australia
(e-mail: \{hanxi.li, chunhua.shen\}@nicta.com.au).

Q. Shi is with University of Adelaide, Adelaide, SA 5000, Australia
(e-mail: qinfeng.shi@ieee.org).

Correspondence should be addressed to C. Shen. 
}
\thanks{ 
\CSAcknowledgement
}
}

\mark{November 2010}

\maketitle

\begin{abstract}

  The $\ell_1$ tracker obtains robustness by seeking a 
  sparse representation of the tracking
  object via $\ell_1$ norm minimization \cite{Xue_ICCV_09_Track}. 
  However, the high computational complexity
  involved in the $ \ell_1 $ tracker restricts its further applications in real time
  processing scenario. Hence we propose a Real Time Compressed Sensing Tracking (RTCST)
  by exploiting the signal recovery power of Compressed Sensing (CS).  Dimensionality
  reduction and a customized Orthogonal Matching Pursuit (OMP) algorithm are adopted to
  accelerate the CS tracking.  As a result, our algorithm achieves a real-time speed
  that is up to $6,000$ times faster than that of the $\ell_1$ tracker.  Meanwhile,
  RTCST still produces competitive (sometimes even superior) tracking accuracy comparing
  to the existing $\ell_1$ tracker. Furthermore, for a stationary camera, a further refined
  tracker is designed by integrating a CS-based background model (CSBM). This
  CSBM-equipped tracker coined as RTCST-B, outperforms most state-of-the-arts with
  respect to both accuracy and robustness. Finally, our experimental results on various
  video sequences, which are verified by a new metric---Tracking Success Probability
  (TSP), show the excellence of the proposed algorithms.

\end{abstract}


\section{Introduction}
\label{sec:intro}
  Within Bayesian filter framework, the representation of the likelihood model is
  essential. In a tracking algorithm, the scheme of object representation determines how
  the concerned target is represented and how the representation is updated. A promising
  representation scheme should accommodate noises, occlusions and illumination changes in
  various scenarios. In the literature, a few representation models have been proposed to
  ease these difficulties \cite{Cootes_PAMI_01_AAM, Comaniciu_PAMI_03_KMS,
  Yilmaz_PAMI_04_Contour, Avidan_PAMI_04_SVT, Serby_ICPR_04_Prob, Shen_CSVT_10_Gener}.
  Most tracking algorithms represent the target by a single model, typically built on
  extracted features such as color histogram \cite{Doucet_SC_00_SMC, Shen_ICCV_05_GKMS},
  textures \cite{Cascia_PAMI_00_Head} and correspondence points \cite{Sha_ICCV_03_PC}.
  Nonetheless, these approaches are usually sensitive to variations in target appearance
  and illumination, and a powerful template update method is usually needed for
  robustness. Other tracking algorithms train a classifier off-line
  \cite{Avidan_PAMI_04_SVT, Williams_PAMI_05_RVMT} or on-line \cite{Shen_CSVT_10_Gener}
  based on multiple target samples. These algorithms benefit from the robust object model,
  which is learned from labeled data by sophisticated learning methods.

  Recently, Mei and Ling proposed a robust tracking algorithm using $\ell_1$ minimization
  \cite{Xue_ICCV_09_Track}. Their algorithm, referred to as the \emph{$\ell_1$ tracker}, is
  designed within Particle Filter (PF) framework \cite{Sanjeev_TSP_02_PF}. There
  a target is expressed as a \emph{sparse} representation of multiple predefined
  templates.  The $\ell_1$ tracker demonstrates promising robustness compared with existing
  trackers \cite{Comaniciu_PAMI_03_Kernel, Porikli_CVPR_2006_Cov, Zhou_TIP_04_AAPF}.
  However, it has following problems: Firstly, $\ell_1$ minimization in their work is
  slow;  Secondly, they use an over-complete dictionary (an identity matrix) to represent
  the background and noise. This dictionary, in fact, can also represent any objects
  (including the user interested tracking objects) in video. Hence it may not discriminate
  the objects against background and noise. 

  Although the $\ell_1$ tracker \cite{Xue_ICCV_09_Track} is inspired by the face recognition
  work using \emph{sparse representation classification} (SRC)\cite{Wright_PAMI_09_Face},
  it doesn't make use of the sparse signal recovery power of Compressed Sensing (CS) used
  in \cite{Wright_PAMI_09_Face}. CS  is an emerging topic originally proposed in signal
  processing community \cite{Donoho_TIT_06_CS, Candes_CPAM_05_Stable}.  It states that
  sparse signals can be exactly recovered with fewer measurements than what the
  Nyquist-Shannon criterion requires with overwhelming probability. It has been applied to
  various computer vision tasks \cite{Wright_PAMI_09_Face, Volkan_ECCV_08_Background,
  Ali_ICASSP_08_Shape}.

  Inspired by the $\ell_1$ tracker and motivated by their problems, we propose two CS-based
  algorithms termed \emph{Real-Time Compressed Sensing Tracking} (RTCST) and
  \emph{Real-Time Compressed Sensing Tracking with Background Model} (RTCST-B)
  respectively. The new tracking algorithms are tremendously faster than the standard
  $\ell_1$ tracker and serve as {\em better} (in terms of both accuracy and robustness)
  alternatives to existing visual object trackers such as those in
  \cite{Sanjeev_TSP_02_PF, Comaniciu_PAMI_03_Kernel, Shen_CSVT_10_Gener}.

  The key contributions of this work can be summarized as follows.
  \begin{enumerate}
    \item
      We make use of the sparse signal recovery power of CS to reduce the computational
      complexity significantly. That is we hash or random project the original features to a
      much lower dimensional space to accelerate the CS signal recovery procedure for
      tracking. Moreover, we propose a customized \emph{Orthogonal Matching Pursuit} (OMP)
      algorithm for real-time tracking. Our algorithms are up to about $6,000$ times faster
      than the standard $\ell_1$ tracker of \cite{Xue_ICCV_09_Track}. In short, {\em we make
      the tracker real-time by using CS}. 
    \item
      We propose background template rather than the over-complete dictionary in
      \cite{Xue_ICCV_09_Track}. This further improves the robustness of the tracking,
      because the representation of the objects and background are better separated. This
      new tracker, which is referred to as RTCST-B in this work, outperforms most
      state-of-the-art visual trackers with respect to accuracy while achieves even higher
      efficiency compared with RTCST.
    \item
      Finally, we propose a new metric called \emph{Tracking Success Probability} (TSP) to
      evaluate trackers' performance. We argue that this new metric is able to measure tracking
      results quantitatively and demonstrate the robustness of a tracker. Consequently, all
      the empirical results are assessed by using TSP in this work.
  \end{enumerate}

  For ease of exposition, symbols and their denotations used in this paper are
  summarized in Table~\ref{tab:notations}.

\begin{table*}[t]
  \caption{Notation}
    \centering
  {    %
    \begin{tabular}{c|p{0.8\textwidth}}
    \hline
    \hline
    Notation & Description \\
    \hline
      $\s_k$     & A dynamic state vector at time $k$ \\
      $\s_k^i$   & A dynamic state vector at time $k$ corresponding to the \\
                 & $i$th particle \\
      $A$        & The measurement matrix or the collection of templates \\
      $\y$       & The observed target, a.k.a, observation \\
      $\x$       & The signal to be recovered in  compressed 
                  sensing. For CS-based pattern recognition or tracking, 
                  it is the coefficient vector for the sparse representation \\
      $\Phi$   & The projection matrix, could be either a random matrix 
                  or a hash matrix in this work\\
      $T, E, B$  & The collection of target, noise and background templates \\
      $\x_t,~\x_e,~\x_b$     & The coefficient vector associated with target, noise and 
                              background templates respectively\\
      $N_t, N_b$             & The number of target templates and background templates \\
      $d_0, d$               & The dimensionality of original and reduced feature space \\
     \hline
     \hline
    \end{tabular}
    }
  \label{tab:notations}
\end{table*}

  The rest of the paper is organized as follows. We briefly review the related literature
  background in the next section. In Section \ref{sec:rcst}, the proposed RTCST algorithm
  is presented.  We present the RTCST-B tracker in Section \ref{sec:rtcstb}. We verify our
  methods by comparing them against existing visual tracking methods in Section
  \ref{sec:exp}.  Conclusion and discussion can be found in the last section.

\section{Related work}
  In this section, we briefly review theories and algorithms closest to our work.
\subsection{Bayesian Tracking and Particle Filters}
  From a Bayesian perspective, the tracking problem is to calculate the posterior
  probability $p(\mathbf{s}_k|\y_k)$ of state $\mathbf{s}_k$ at time $k$, where $\y_k$ is
  the observed measurement at time $k$ \cite{Sanjeev_TSP_02_PF}. In principle, the
  posterior PDF is obtained recursively via two stages: prediction and update. The
  prediction stage involves the calculation of prior PDF:
  \begin{equation}
    p(\mathbf{s}_k|\y_{k-1}) = \int p(\mathbf{s}_k|\mathbf{s}_{k-1}) p(\mathbf{s}_{k-1}|\y_{k-1})
    d  \s_{k-1}.
    \label{equ:bayes_prediction}
  \end{equation}
  In the update stage, the prior is updated using Bayes' rule
  \begin{equation}
    p(\mathbf{s}_k|\y_k) = \frac{p(\y_k|\mathbf{s}_k) 
    p(\mathbf{s}_k|\y_{k-1})}{p(\y_k|\y_{k-1})}.
    \label{equ:bayes_update}
  \end{equation}

  The recurrence relations \eqref{equ:bayes_prediction} and \eqref{equ:bayes_update} form
  the basis for the optimal Bayesian solution. Nonetheless, the solution of above problem
  can not be analytically solved without further simplification or approximation.
  Particle Filter (PF) is a Bayesian sequential importance sampling technique for
  estimating the posterior distribution $p(\mathbf{s}_k|\y_k)$. By introducing the
  so-called \emph{importance sampling distribution} \cite{Doucet_SC_00_SMC}: 
  \begin{equation}
      \mathbf{s}_i \sim ~ q(\mathbf{s}),\; i = 1, \dots, N_s,
    \label{equ:important_sample}
  \end{equation}
  the posterior density is estimated by a weighted approximation,
  \begin{equation}
    p(\mathbf{s}_k|\y_k) \approx \sum_{i=1}^{N_s} w_k^i\delta(\mathbf{s}_k - \mathbf{s}_k^i).
    \label{equ:pf_weighted_density}
  \end{equation}
  Here  
  \begin{equation}
    w_k^i \propto w_{k-1}^i
    \frac{p(\y_k|\mathbf{s}_k^i)p(\mathbf{s}_k^i|\mathbf{s}_{k-1}^i)}{q(\mathbf{s}_k^i|\mathbf{s}_{k-1}^i,\:\y_k)}.
    \label{equ:pf_weight_update}
  \end{equation}
  For the sake of convenience, $q(\cdot)$ is commonly formed as 
  \begin{equation}
      q(\mathbf{s}_k|\mathbf{s}_{k-1}^i,\:\y_k) = p(\mathbf{s}_k|\mathbf{s}_{k-1}^i).  
    \label{equ:q_common_choice}
  \end{equation}
  Therefore, \eqref{equ:pf_weight_update} is simplified into 
  \begin{equation}
    w_k^i \propto w_{k-1}^i p(\y_k|\mathbf{s}_k^i)
    \label{equ:pf_weight_update_simple}
  \end{equation}
  The posterior then could be updated only depending on its previous value and observation
  likelihood $p(\mathbf{s}_k|\mathbf{s}_{k-1}^i)$. Plus, in order to reducing the effect of \emph{particle
  degeneracy} \cite{Doucet_SC_00_SMC}, a resampling scheme is usually implemented as
  \begin{equation}
    Pr(\mathbf{s}_k^{i*} = \mathbf{s}_k^j) = w_k^j, \; j = 1, 2, \dots, N_s
    \label{equ:pf_resample}
  \end{equation}
  where the set $\{\mathbf{s}_k^{i*}\}_{i=1}^{N_s}$ is the particles after re-sampling. 
  
  Like the $\ell_1$ tracker, both  RTCST and RTCST-B trackers use PF framework. However,
  they differ in how to seek a sparse representation which consequently lead to different
  observation likelihood $p(\mathbf{s}_k|\mathbf{s}_{k-1}^i)$ estimtation.

\subsection{$\ell_1$-norm Minimization-based Tracking} 

  The underlying conception behind SRC is that in many circumstances, an observation
  belonging to a certain class lies in the subspace that is spanned by the samples belong
  to this class, and the linear representation is assumed to be sparse. Hence,
  reconstructing the sparse coefficients associated with the representation is crucial to
  identify the observation. The coefficients recovery could be accomplished by solving a
  relaxed version of
  \eqref{equ:cs_opt}
  \begin{equation}
      \begin{split}
        \min_{\x} ~ \|\x\|_{1}, \;
        \sst ~ \|A\x & - \y\|_{2} \leq \varepsilon,
      \end{split}
      \label{equ:cs_opt_track_lasso}
  \end{equation}
  where $\x \in \mathbb{R}^n$ is the  coefficient vector of interest; $A = [\mathbf{a}_1,
  $ $ \mathbf{a}_2, \dots, $ $ \mathbf{a}_n] \in \mathbb{R}^{d \times n}$ is sometimes
  dubbed as \emph{dictionary} and composed of pre-obtained pattern samples $a_i \in
  \mathbb{R}^d ~\forall i$; and $\y \in \mathbb{R}^d$ is the query/test observation.
  $\varepsilon$ is error tolerance. Then, the class identity $l(\y)$ is retrieved as
  \begin{equation}
    l(\y) = \argmin_{j \in \{1, \cdots, C\}}{r_j(\y)},
    \label{equ:cs_opt_cv_class}
  \end{equation}
  where $r_j(\y) \doteq \|\y - A\delta_j(\x)\|_{2}$ is the reconstruction residual
  associated with class $i$, $C$ is the number of classes and the function $\delta_j(\x)$
  sets all the coefficients of $\x$ to $0$ except those corresponding to $j$th class
  \cite{Wright_PAMI_09_Face}. 

  Given a target template set $T = [\t_1, \cdots, \t_{N_t}] \in \mathbb{R}^{d_0 \times
  N_t}$ and a noise template set $E = [I,\;-I] \in \mathbb{R}^{d_0 \times 2d_0}$, the
  $\ell_1$ tracker adopts a positive-restricted version of \eqref{equ:cs_opt_cv_lasso} for
  recovering the sparse coefficients $\x$, \ie,
  \begin{equation}
    \setlength{\abovedisplayskip}{0.1cm}
    \setlength{\belowdisplayskip}{0.1cm}
    \begin{split}
        \min ~ \|\x\|_{1}, \;
        \sst ~ \|A\x & - \y\|_{2} \leq \varepsilon, \, 
            ~ \x  \succeq 0. 
    \end{split}
    \label{equ:cs_opt_track}
  \end{equation}
  Here $A \doteq [T, E] \in \mathbb{R}^{d_0 \times (N_t + 2d_0)}$ is the combination of
  target templates and noise templates while $\x \doteq [\x_t^\T, ~\x_e^\T]^\T \in
  \mathbb{R}^{N_t + 2d_0}$ denotes the associated target coefficients and noise
  coefficients. Note that $N_t$ denotes the number of target templates and $d_0$ is the
  original dimensionality of feature space which equals to the pixel number of the initial
  target. The $\ell_1$ tracker tracks the target by integrating \eqref{equ:cs_opt_track}
  and a template-update strategy into the PF framework. Algorithm~\ref{alg:l1_tracker}
  illustrates the tracking procedure. In addition, there is a heuristic approach for
  updating the target templates and their weights in the $\ell_1$ tracker. Refer to
  \cite{Xue_ICCV_09_Track} for more details.

\begin{algorithm}[t]
  \caption{{$\ell_1$ Tracking}}   

  \KwIn
  {
    \begin{itemize}
      \item Current frame $F_k \in \mathbb{R}^{h \times w}$. 
      \item Particles $\mathbf{s}_{k-1}^i,~i = 1, 2, \cdots, N_s$. 
      \item Templates set $A = [T,~E]~\in~ \mathbb{R}^{d_0 \times (N_t + 2d_0)}$.
      \item Templates' weight vector $\mathbf{\alpha}$ associated with $T$.
    \end{itemize}
  }
  \Begin
  {
    Generate new particles $\mathbf{s}_k^i,~i = 1, 2, \cdots, N_s$ within the PF framework\;
    \For{$i \leftarrow 1$ \KwTo $N_s$}
    {
      Obtain observation $\y_i$ corresponding to $\mathbf{s}_k^i$\;
      Obtain $\x$ via solving \eqref{equ:cs_opt_track} with IP-based methods\;
      Calculate residual: $r_i = \|\mathbf{y}_i-T\cdot\x_t\|_2$\;
    }
    $i^* \longleftarrow \argmin_{1\leq i \leq N_s}(r_i)$\;
    Get the observed target $\mathbf{y}_k \longleftarrow \mathbf{y}_{i^*}$ and its state
    $\s_k \longleftarrow \mathbf{s}_k^{i^*}$\;
    Update templates $T$ and weights $\mathbf{\alpha}$ based on $\x_{i^*}$ as in \cite{Xue_ICCV_09_Track}\;
  }

  \KwOut
  {
    \begin{itemize}
      \item Tracked target $\mathbf{y}_k$.
      \item Updated target dynamic state $\s_k$.
      \item Updated target templates $T$ and their weights $\mathbf{\alpha}$.
    \end{itemize}
  }

\label{alg:l1_tracker}
\end{algorithm}

\subsection{Compressed sensing and its application in pattern recognition}
  CS states that a $\eta$-sparse\footnote{\scriptsize{a signal $\x$ is said $\eta$-sparse if
  there are at most $\eta$ nonzero entries in $\x$.}} signal $\x\in\mathbb{R}^n$ can be
  exactly recovered with overwhelming probability via few measurements
  \begin{equation*}
    \setlength{\abovedisplayskip}{0.1cm}
    \setlength{\belowdisplayskip}{0.1cm}
    y_i =  \Phi_i\x, ~~~i=1,\dots,m \ll n.
  \end{equation*}
  Intuitively, one would achieve $\x$ via
  \begin{equation}
    \setlength{\abovedisplayskip}{0.1cm}
    \setlength{\belowdisplayskip}{0.1cm}
      \begin{split}
          \min_{\x} ~& \|\x\|_{0},
        \;  
        \sst ~ \Phi  \x = \y,
      \end{split}
      \label{equ:ori_cs_opt}
  \end{equation}
  where $\Phi \in \mathbb{R}^{m \times n}$ is the measurement matrix, of which rows are
  the measurement vectors $\Phi_i$ and $\y = (y_1,\dots,y_m)^T$.  $ \|\x\|_{0} $ is the
  number of non-zero elements of $ \x $. Since \eqref{equ:ori_cs_opt} is NP-hard
  \cite{Tropp_TIT_07_OMP}, it is commonly relaxed to 
  \begin{equation}
    \setlength{\abovedisplayskip}{0.1cm}
    \setlength{\belowdisplayskip}{0.1cm}
      \begin{split}
        \min_{\x} ~& \|\x\|_{1}, \; 
        \sst ~ \Phi  \x = \y, 
      \end{split}
      \label{equ:cs_opt}
  \end{equation} which can be casted into a linear programming problem.

  As regards CS-based pattern recognition, to deal with noise, one could alternatively
  solve a Second Order Cone Program:
  \begin{equation}
    \setlength{\abovedisplayskip}{0.1cm}
    \setlength{\belowdisplayskip}{0.1cm}
      \begin{split}
        \min_{\x} ~ \|\x\|_{1}, \;
        \sst ~ \|\Phi\x & - \y\|_{2} \leq \varepsilon,
      \end{split}
      \label{equ:cs_opt_cv_lasso}
  \end{equation} where $ \varepsilon$ is a pre specified tolerance.

\section{Real-time compressed sensing tracking}
\label{sec:rcst}
  In this section, we present the proposed real-time CS tracking. 
\subsection{Dimension reduction}
\label{subsec:dim_reduct}
  The biggest problem of $\ell_1$ tracking is the extremely high dimensionality of
  the feature space, which leads to heavy computation.  More precisely, suppose that the
  cropped image of observation is $I \in \mathbb{R}^{h \times w}$, the dimensionality $d_0
  = h\cdot w$ is typically in the order of $10^{3} \sim 10^{5}$, which prevents tracking
  from real-time.     

  Fortunately, in the context of compressed sensing (ignoring the non-negativity
  constraint on $\x$ for now), it is well known that if the measurement matrix $\Phi$ has
  Restricted Isometry Property (RIP) \cite{Candes_CPAM_05_Stable}, then a sparse signal
  $\x$ can be recovered from  
  \begin{equation}
    \setlength{\abovedisplayskip}{0.1cm}
    \setlength{\belowdisplayskip}{0.1cm}
    \begin{split}
        \min ~ \|\x\|_{1}, \;
        \sst ~ \|\Phi A\x & - \Phi\y\|_{2} \leq \varepsilon. 
    \end{split}
    \label{equ:cs_opt_track_cs}
  \end{equation}
  A typical choice of such measurement matrix is random gaussian matrix 
  \begin{equation*}
    \setlength{\abovedisplayskip}{0.1cm}
    \setlength{\belowdisplayskip}{0.1cm}
      R  \in \mathbb{R}^{d \times n},\quad R_{i,j} \sim \mathcal{N}(0,\;1).
    \label{equ:def_rand_matrix}
  \end{equation*}

  Besides random projection, there are other means that guarantee RIP.  Shi \etal
  \cite{Shi_JMLR_09_Hash} proposed a hash kernel to deal with the issue of computational
  efficiency.  Let $h_s(j,d)$ denotes a hash function (\ie, the hash kernel)
  $h_s:\mathbb{N} \to \{1,\dots,d\}$ drawn from a distribution of pairwise independent
  hash functions, where $s \in \{1,\dots,S\}$ is the seed. Different seed gives different
  hash function. Given $h_s(j,d)$, the hash matrix $H$ is defined as
  \begin{align}
    H_{ij}:= \left\{{\begin{array}{*{20}c}
                     2h_s(j,2)-3, & {h_s(j,d) = i},\forall s\in\{1,\dots,S\}  \\
                     0, & \text{otherwise}.  \\
                    \end{array} } \right.
  \end{align} 
  Obviously, $H_{ij} \in \{0,\pm 1\}$. The hash kernel generates hash matrices more
  efficiently than conventional random matrices while maintains the similar random
  characteristics, which implies good RIP. 

  In this work, the dimensionality of feature space is reduced by matrix $\Phi \in
  \mathbb{R}^{d \times d_0}$ (which could be either random matrix $R$ or hash matrix $H$)
  from $d_0$ to $d$ where $d \ll d_0$. This significantly speeds up solving equation
  \eqref{equ:cs_opt_cv_lasso}, for its complexity depends on $d$ polynomially. 

\subsection{Customized orthogonal matching pursuit for real-time tracking}
\label{subsec:omp}
\subsubsection{Orthogonal matching pursuit}
\label{subsubsec:omp}
    Before the compressed sensing theory was proposed, numerous approaches had been
    applied for sparse approximation in the literature of signal processing and statistics
    \cite{Mallat_TSP_93_MP, Pati_ACSSC_93_OMP, Davis_OE_94_Adap}.  Orthogonal Matching
    Pursuit (OMP) is one of the approaches and solves \eqref{equ:ori_cs_opt} in a greedy
    fashion. Tropp and Gilbert \cite{Tropp_TIT_07_OMP} proved OMP's recoverability and
    showed its higher efficiency compared with linear programming which is adopted by the
    original $\ell_1$ tracker of \cite{Xue_ICCV_09_Track}. Be more explicit, given that $
    A \in \mathbb{R}^{d \times n}$ the computational complexity of linear programming is
    around $O(d^2n^{\frac{3}{2}})$, while OMP can achieve as low as $O(dn)$\footnote{Here,
    however, we do not employ the trick  that Tropp and Gilbert mentioned for the
    least-squares routine. As a result, the OMP's complexity is higher than $O(dn)$ but
    still much lower than that of linear programming.}.  We implement the sparse recovery
    procedure of the proposed tracker with OMP so as to accelerate the tracking process.

    The number of measurements required by OMP is $O(\eta log(n))$ for $\eta$-sparse
    signals, which is slightly harder to achieve compared with that in $\ell_1$
    minimization. However, it is merely a theoretical bound for signal
    recovering, no significant impact of OMP upon the tacking accuracy is observed in
    our experiments (see Section~\ref{sec:exp}).

\subsubsection{Further acceleration---OMP with early stop}
\label{susubbsec:early_stop_omp}
    
\begin{figure}[h]
      \includegraphics[width=0.45\textwidth]{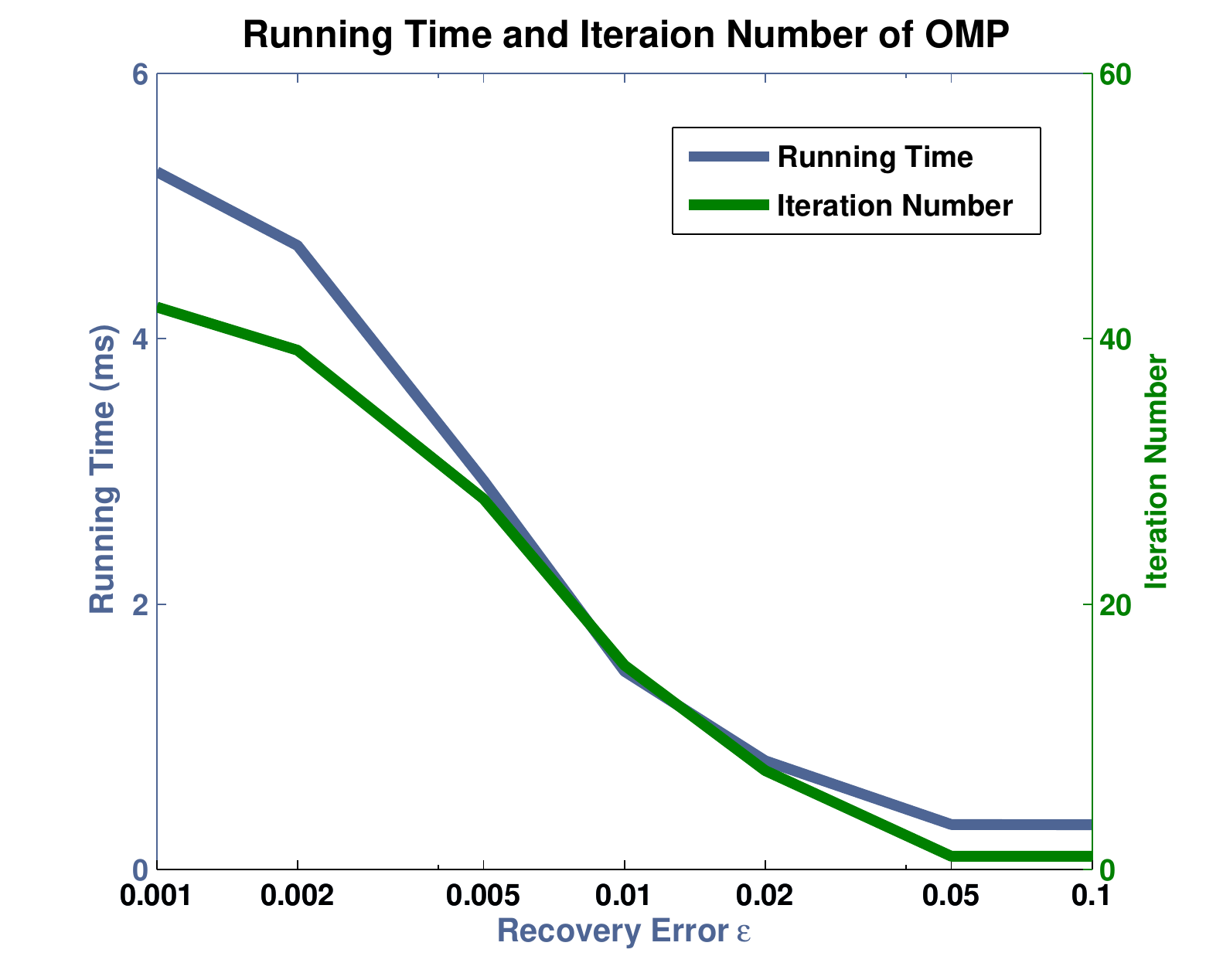}
    \caption{
            The decreasing tendency of running time and iteration numbers of the OMP procedure
            with different residual thresholds. The result is produced from a Matlab-based
            experiment on video ``Cubicle'', with the feature dimension of $50$. Both the running
            time and iteration numbers are the average result over all the frames and
            particles.
         }
         \label{fig:time_iter}
    \end{figure}

    The OMP algorithm was proposed for recovering sparse signal exactly (see Equation
    \eqref{equ:ori_cs_opt}), and the perfect recovery is also guaranteed within $d$ steps
    \cite{Pati_ACSSC_93_OMP}. However, in the realm of pattern recognition, we argue that
    there is no requirement for perfect recovery for many applications.  For example, for
    classification problems, test accuracy is of interest and exact recovery does not
    necessarily translate into high classification accuracy.  So on the contrary, an
    appropriate recovery error may even improve the accuracy of recognition
    \cite{Wright_PAMI_09_Face}. We introduce a residual based stopping criterion into OMP
    by modifying \eqref{equ:ori_cs_opt} as
    \begin{equation}
      \begin{split}
        \min_{\x} ~ \|\x\|_{0}, \;
        \sst ~ \|A\x & - \y\|_{2} \leq \varepsilon. 
      \end{split}
      \label{equ:omp_opt_residual}
    \end{equation}
    Moreover, the procedure of OMP could be accelerated remarkably if the above stopping
    criterion is enforced. To understand this, let us assume that OMP follows the MP algorithm
    \cite{Mallat_TSP_93_MP} with respect to the convergence rate\footnote{Although the
    convergence rate for MP algorithm is $O(1/\sqrt{t})$, the convergence rate
    for OMP remains unclear.}, \ie,
    \begin{equation}
      r_k = \frac{K}{\sqrt{t}}, \; t < n,
      \label{equ:mp_convergence}
    \end{equation}
    where $K$ is a positive constant and $r_k = \|A\x_k -\y \|_2$ is the recovery residual
    after $t$ steps. Given that we relax the stopping criterion $\varepsilon$ by $10$ times
    \begin{equation}
      \varepsilon' = 10\varepsilon,
      \label{equ:change_stop_residual}
    \end{equation}
    then the required step $t_{\rm stop}$ is reduced to be
    \begin{equation}
      \begin{split}
        t_{\rm stop}' = & ~K^{2} / {\varepsilon'}^{2} \\
        = & ~10^{-2}K^{2} / \varepsilon^{2} \\
        = & ~10^{-2}t_{\rm stop}. 
      \end{split}
      \label{equ:change_stop_step}
    \end{equation}
    Considering that the complexity of OMP is at least proportional to $t$, the algorithm
    could be accelerated by $100$ times theoretically. Figure \ref{fig:time_iter} shows the
    empirical influence of the terminating criterion upon the running iterations and running
    time. In our algorithm, we empirically set the stopping threshold $\varepsilon =
    0.01$,  which draws a balance between speed and accuracy.

\subsubsection{Tracking with a large number of templates}
\label{susubbsec:inf_rcst}

  One noticeable advantage of the SRC-based tracker is the exploitation of multiple
  templates obtained from different frames. However, for the $\ell_1$ tracker, the number
  of templates $n$ should be curbed into strictly because it equals to the dimensionality
  of the optimization variable $\x$. To design a good $\ell_1$ tracker, a trade-off
  between $n$ and the optimization speed is always required. Fortunately, this dilemma
  dose not exist when the tracker is facilitated with OMP and a carefully-selected
  sparsity $\eta$.

  The computational burden of OMP consists of two steps: one is for selecting the maximum
  correlated vector from matrix $A \in \mathbb{R}^{d \times n}$, and the other is for
  solving the least squares fitting. In step $t ~(t < d)$, it is trivial to compute the
  complexity of the first step is $O(dn)$ and that for least-square fitting is $O(d^3 +
  td^2 + td)$.  Accordingly, the running time of OMP is dominated by solving the
  least-squares problem, which is independent of the number of templates, $n$. In other
  words, {\em within a certain number of iterations, the amount of templates would not
  affect the overall running time significantly}.  This is an important and desirable
  property in the sense that we might be able to employ a large amount of templates. 

\begin{figure}[h!]
    \includegraphics[width=0.45\textwidth]{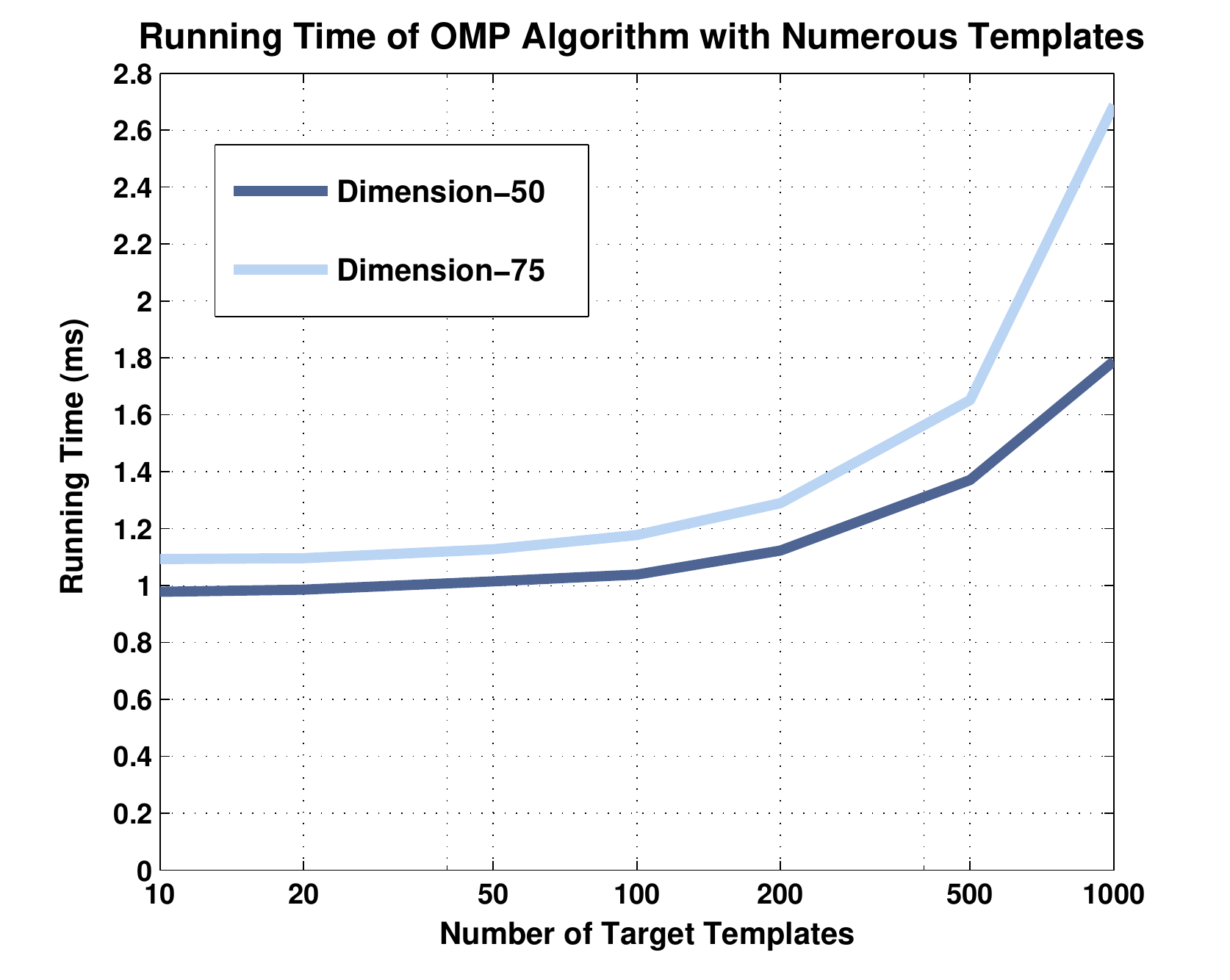}
  \caption{
          Running time of OMP with various numbers of target templates. The
           experiment is carried out on video sequence ``Cubicle'' with
          reduced dimensions $50$ and $75$. The recorded running time is the average time
          consumption for one OMP procedure which calculates the observation likelihood
          for a particle. Note that the $x$-axis only indicates target templates' number,
          and the number of trivial templates is not counted.
          The sparsity $\eta = 15$. 
       }
  \label{fig:time}
  \end{figure}

  Admittedly, larger $n$ might lead to more iterations. However, if we impose a maximum
  sparsity $\eta$, the OMP procedure would only last for $\eta$ steps in the worst
  scenario. From this perspective, a preset $\eta \ll n$ is capable to eliminate the
  influence of a large $n$ upon the running iterations. Figure~\ref{fig:time} depicts the
  change tendency of running time with increasing $n$, given that $d \in \{50, 75\}$,
  $\eta = 15$.  As can be seen, the elapsed time is only doubled when $n$ is raised by
  $10^2$ times. 

  Inspired by this valuable finding, we aggressively set the number of target templates to
  $100$ which is $10$ times larger than that in X. Mei's paper. We try to harness the
  enormous target templates to accommodates the variation of illumination, gesture and
  occlusion and consequently improve the tracking accuracy. As regards the sparsity, we
  elaborately set $\eta = 0.5\cdot d$ for RTCST and $\eta = 15$ for RTCST-B which is
  introduced in Section~\ref{sec:rtcstb}. We believe the numbers are sufficiently large
  for the representations.

  Hereby, we sum up all the adjustments to OMP mentioned in
  Algorithm~\ref{alg:customized_omp}. Note that here we use the inner product rather than
  its absolute value to verify the correlation. This heuristic manner is used to make the
  recovered coefficient vector $\x \succeq 0$, approximately. For RTCST-B introduced in
  next section, the absolute value of inner product is re-employed due to the absence of
  the positive constraint.

\begin{algorithm}[h]
  \caption{Customized OMP for Tracking}   
  \KwIn
  {
    \begin{itemize}
      \itemsep -2pt
      \item A normalized observation $\y \in \mathbb{R}^d$.
      \item A mapped templates set $\Phi A = [\a_1, \cdots, \a_n] \in \mathbb{R}^{d \times n}$.
      \item A recovery residual $0 < \varepsilon \ll 1$.
      \item A sparsity $0 < \eta \ll n$.
    \end{itemize}
  }
  \Begin
  {
    Initialize the residual $\r_0 = \y$, index set $\Lambda_0 = \varnothing$ and selected
    template set $\Psi_0 = \varnothing$\;
    \For{$t \leftarrow 1$ \KwTo $\eta$}
    {
      $\lambda_t = \argmax_{j = 1, \ldots, n}\langle r_{t-1}, \a_j\rangle$\;
      $\Lambda_{t} = \Lambda_{t-1} \cup \{\lambda_t\}$\;
      $\Psi_t = [\Psi_{t-1}\;\;\boldsymbol{\a}_{\lambda_t}]$\;
      Solve the least-squares problem: \\
      \quad \quad \quad $\x_t = \argmin_{\x}\|\Psi_t\x - \y\|_2$\;
      Calculate the new residual: \\
      \quad \quad \quad \quad $\r_t = \y - \Psi_t\x_t$ \;
      \lIf{$\|\r_t\|_2 < \varepsilon$}{break}\;
    }
    Retrieve signal $\x$ according to $\x_t$ and $\Lambda_t$\; 
  }
  \KwOut
  {
    \begin{itemize}
      \item Recovered coefficients $\x \in \mathbb{R}^n$
    \end{itemize}
  }
\label{alg:customized_omp}
\end{algorithm}

\subsection{Minor modifications}
\label{subsec:minor_modifications}
    Besides the dimension reduction methods and OMP, modifications to the original
    $\ell_1$ tracker are proposed in this section to achieve a even higher tracking
    accuracy. 
\subsubsection{Update templates according to sparsity concentration index} 
\label{susubbsec:sci}
    In the $\ell_1$ tracker, the template set is updated when a certain threshold of similarity 
    is reached, \ie, 
    \begin{equation}
      {\rm sim}(\y, \mathbf{a}_i) < \tau, 
      \label{equ:cst_update_sim}
    \end{equation}
    where $i = \argmax(x_i)$ and $ {\rm sim}(\y, \mathbf{a})$ is the function for evaluating the
    similarity between vectors $\y$ and $\a$. It can be the angle between two vectors or SSD
    between them. However, Wright \etal proposed a better approach to validate the
    representation. The approach, which utilizes the recovered $\x$ itself rather
    than the similarity, is termed \emph{Sparsity Concentration Index}
    (SCI) \cite{Wright_PAMI_09_Face}.
    Particularly, in the context of RTCST, class
    number is $1$ if the noise is not viewed as a class, 
    then we obtain a simplified SCI measurement  for the target class, which writes 
    \begin{equation}
        \text{SCI}_{t}(\x) = \|\x_t\|_1 / \|\x\|_1  \in  [0, 1],
        \label{equ:rtcst_sci}
    \end{equation}
    where $\x_t = \x(1 : N_t)$. In the presented RTCST algorithm, $\text{SCI}_{t}$ is
    employed instead of \eqref{equ:cst_update_sim}. 

\subsubsection{Abandoning the template weight} 
\label{susubbsec:abandon_weight}
    The original $\ell_1$ tracker enforces a template re-weighting scheme to distinguish templates
    by \cite{Xue_ICCV_09_Track}, their importance. Nonetheless, following their scheme the
    weight of each target template is always smaller than that of noise templates (see
    Algorithm~ \ref{alg:l1_tracker}). This does not make much sense.  Actually, it may be
    intractable to design an ideal template re-weighting scheme that works in all the
    circumstances. 
    A poorly-designed re-weighting scheme could even deteriorate the tracking performance.
    We abandon the template weight because the importance of templates be easily exploited
    by the compressed sensing procedure. Without template weights, the tracker becomes
    simpler and less heuristic. The empirical result also shows better tracking accuracy
    when template weight is abandoned.

\subsubsection{MAP and MSE} 
\label{susubbsec:mse}
    In Mei and Ling's framework \cite{Xue_ICCV_09_Track}, the new state ${\s}_k$ is
    corresponding to the particle with the largest observation likelihood. This method is
    known as the Maximum A Posterior (MAP) estimation. It is also known that for the
    particle filtering framework, Mean Square Error (MSE) estimation is usually more
    stable than MAP. As a result, we adopt MSE in our real-time tracker, namely, 
    \begin{equation}
      {\s}_k = \frac{\sum^{N_s}_{i=1}({\s}_k^i \cdot l_i)}{\sum^{N_s}_{i=1}l_i},
        \label{equ:mse}
    \end{equation}
    where ${\s}_k^i$ is the $i$th particle at time $k$ and $l_i$ is the corresponding
    observation likelihood.

\subsection{The Algorithm}
\label{subsec:rcst}
  In a nutshell, for each observation, we utilize Algorithm~\ref{alg:customized_omp} to
  recover the coefficient vector $\x$ by solving the problem 
  \begin{equation}
      \min_{\x}\|\x \|_{0},~\sst\; \|\Phi A\x - \Phi \y\|_{2} \leq \varepsilon,~
       \x \succeq 0
  \label{equ:rtcst_opt}
  \end{equation}
  where $\x = [\x_t, ~\x_e]$, $A = [T,~E]$. The residual is then obtained by  
  \begin{equation}
    r = \| \Phi \y - \Phi A\x_t\|_2.
    \label{equ:rtcst_residual}
  \end{equation}
  Finally the likelihood of this observation is updated as 
  \begin{equation}
     l = \exp(-\lambda\cdot r),~\lambda > 0.
    \label{equ:rtcst_lklhd}
  \end{equation}
    
  The procedure of Real-Time Compressed Sensing Tracking algorithm is summarized in
  Algorithm~\ref{alg:rtcst_tracker}. Our template update scheme is demonstrated in
  Algorithm~\ref{alg:rtcst_temp_up}. As can be seen, the proposed update scheme is much
  conciser than that in the $\ell_1$ tracker \cite{Xue_ICCV_09_Track} thanks to the
  abandonment of template weight. The empirical performance of RTCST is verified in
  Section~\ref{sec:exp}. 

\begin{algorithm}[t!]
  \caption{Real-Time Compressed Sensing Tracking}   
  \KwIn
  {
    \begin{itemize}
      \itemsep -2pt
      \item Current frame $F_k \in \mathbb{R}^{h \times w}$. 
      \item Particles $\mathbf{s}_{k-1}^i,~i = 1, 2, \cdots, N_s$ 
      \item A dimension-reduction matrix $\Phi ~\in~\mathbb{R}^{d \times d_0}$.
      \item A Templates set $A = [T,~E]~\in~ \mathbb{R}^{d_0 \times (N_t + 2d)}$.
      \item A preset parameter $\lambda > 0$.
    \end{itemize}
  }
  \Begin
  {
    Normalize every column of $\Phi A$\;
    Generate new particles $\mathbf{s}_k^i,~i = 1, 2, \cdots, N_s$\;
    \For{$i \leftarrow 1$ \KwTo $N_s$}
    {
      Obtain mapped observation $\Phi \y_i$ corresponding to $\mathbf{s}_k^i$\;
      Get $\x$ via solving \eqref{equ:rtcst_opt} with Algorithm~\ref{alg:customized_omp}\;
      Calculate residual $r_i$ via \eqref{equ:rtcst_residual}\;
      Calculate observation likelihood $l_i = \exp(-\lambda\cdot r_i)$
    }
    Calculate target dynamic state $\boldsymbol{\s}_k$ via \eqref{equ:mse} and then get the
    target $\y_k$\;
    Recalculate $\x_k$ for $\y_k$ via solving \eqref{equ:rtcst_opt}\;
    Update templates $T$ based on $\x_k$ and \eqref{equ:rtcst_sci}\;
  }

  \KwOut
  {
    \begin{itemize}
      \itemsep -2pt
      \item Tracked target $\y_k$.
      \item Updated target dynamic state $\boldsymbol{\s}_k$.
      \item Updated target templates $T$.
    \end{itemize}
  }

\label{alg:rtcst_tracker}
\end{algorithm}

\begin{algorithm}[h]
  \caption{Template Update Scheme for RTCST}   
  \KwIn
  {
    \begin{itemize}
      \item Sparse coefficient $\x = \x_k$ in Alg.~\ref{alg:rtcst_tracker}.
      \item Observed target $\mathbf{y}_k$.
      \item Target templates set $A=[\a_1, \a_2, \cdots, \a_{N_t}]~\in~ \mathbb{R}^{d_0 \times N_t}$.
      \item A preset parameter $0 < \tau < 1$.
    \end{itemize}
  }
  \Begin
  {
  \If{$\text{SCI}_{t}(\x) < \tau$}
    {
      $j^* \longleftarrow \argmin_{1\leq j \leq N_t}(\x_j)$\;
      $\a_{j^*} \longleftarrow \mathbf{y}_k$, where $\a_{j^*}$
      is the $j^*$th target template;
    }
  }

  \KwOut
  {
    \begin{itemize}
      \item Updated target templates $A$.
    \end{itemize}
  }

\label{alg:rtcst_temp_up}
\end{algorithm}

\section{RTCST-B: More Robust and Efficient RTCST with background model}
\label{sec:rtcstb}

  To some extent, visual tracking is viewed as object detection task with prior
  information. Similar to object detection, which is sometimes treated as a classification
  problem, visual tracking also distinguishes the foreground (target) from background. In
  detection applications, the background class is usually considered without distinct
  feature because it could follow any pattern. Quite the contrary, in the context of visual
  tracking, the background is much more limited with respect to appearance variation.
  Particularly, for the stationary camera, the background is nearly fixed. Under these
  assumptions, it is worthwhile exploiting the background information for tracking.  And
  appropriate incorporations of background model indeed improve the tracking
  performance\cite{Stauffer_CVPR_99_Back, Isard_ICCV_01_Blob, Zhao_PAMI_08_Segmentation,
  Shen_CSVT_10_Gener}. 

  We hereby propose a novel CS-based background model (CSBM) to facilitate tracking
  algorithm.  The definition of CS-based background model is quite simple. Suppose that
  $\Gamma_i \in \mathbb{R}^{h \times w},\;i = 1, \cdots, N_b$ is the $i$th frame where
  foreground is absent, and $h$ and $w$ are the height and width of the frame respectively,
  we define the background model as
  \begin{equation}
    \mathbb{G} = \{\Gamma_1, \dots, \Gamma_{N_b}\}  
    \label{equ:def_cbm}
  \end{equation}
  or in short, the collection of $N_b$ backgrounds. The background templates are then
  generated from CSBM to cooperate with target templates in our new tracker. 

  Please note that our algorithms is unrelated to the background subtraction manner
  proposed by Volkan \etal \cite{Volkan_ECCV_08_Background}. In their paper, foreground
  silhouettes are recovered via CS procedure but the background subtraction is still
  performed in conventional way. Our CSBM and RTCST-B is entirely different from their
  manner, both in essence and appearance. The details of CSBM and its incorporation with
  RTCST are introduced below. 

\subsection{Building the Optimal CSBM}
\label{subsec:cs_back}
  A good CSBM should only constitute ``pure'' backgrounds and contain sufficiently large
  appearance variation, \eg, illumination changes. Ideally, we could simply select certain
  number of foreground-absent frames from video sequence to build a CSBM. However, the
  ``pure'' background is usually difficult to find and it is even harder to obtain the
  ones cover the main distribution of background appearance. 

\begin{figure}[h]
      \subfigure[]{\includegraphics[width=0.155\textwidth,clip]{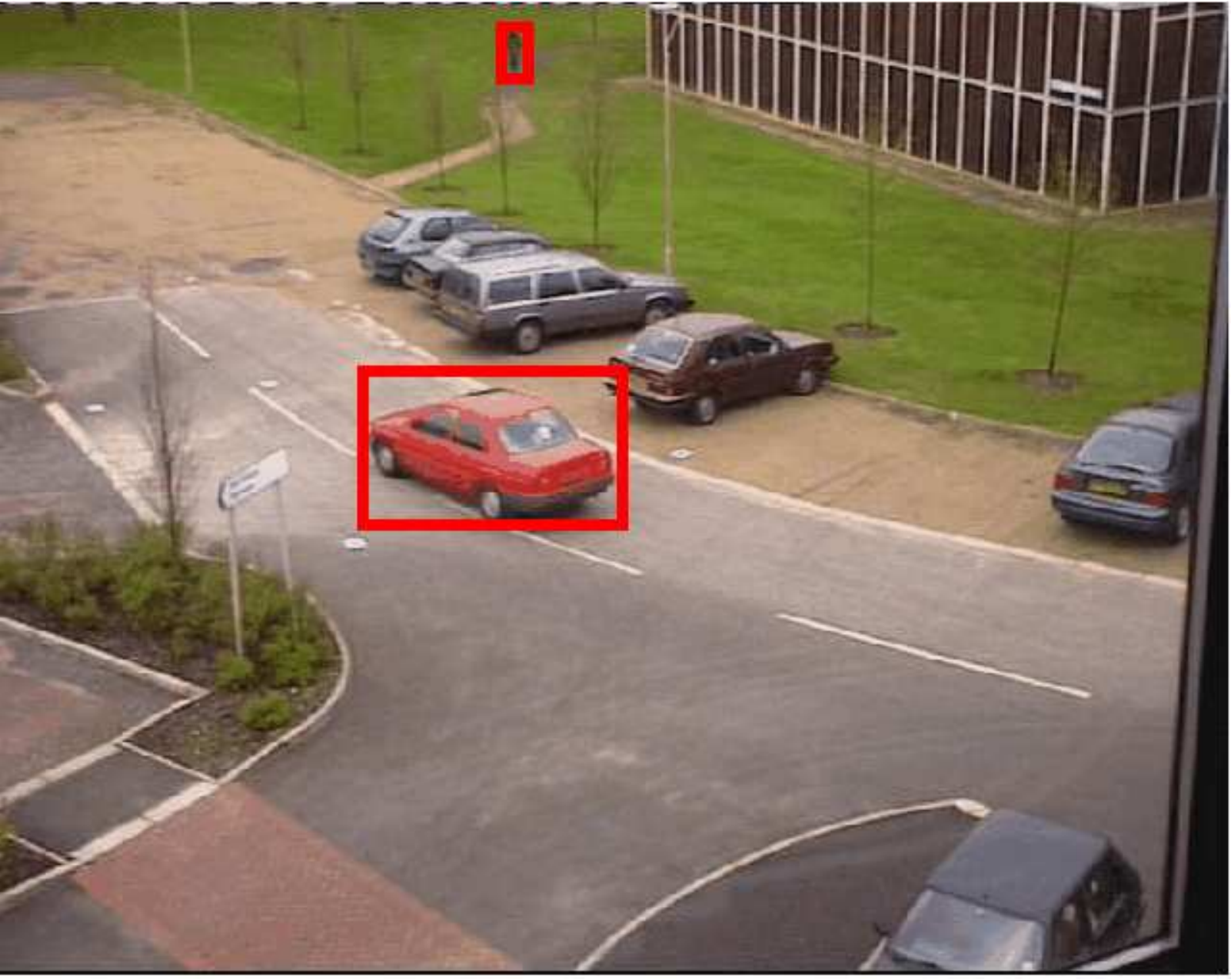}}
      \subfigure[]{\includegraphics[width=0.155\textwidth,clip]{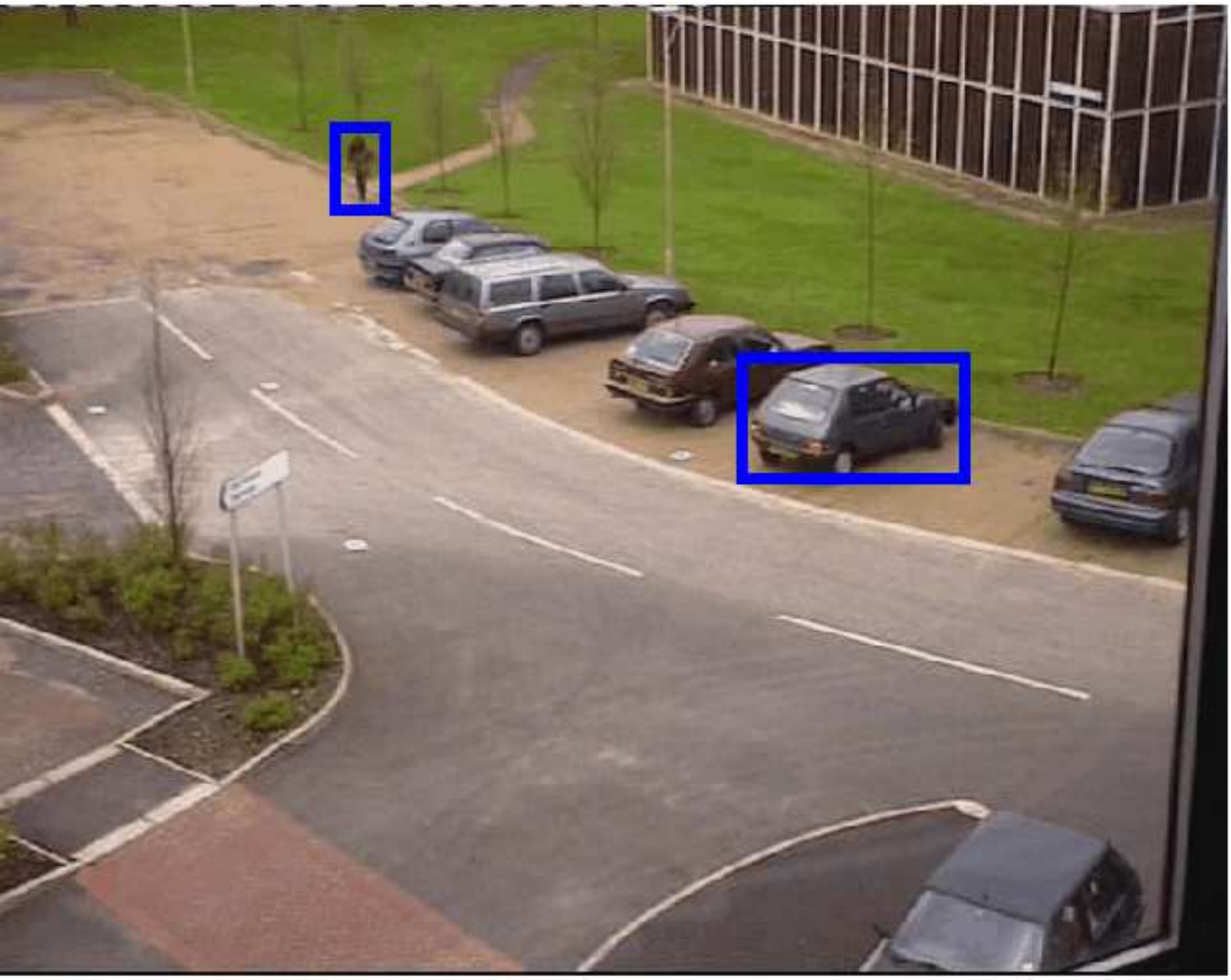}}
      \subfigure[]{\includegraphics[width=0.155\textwidth,clip]{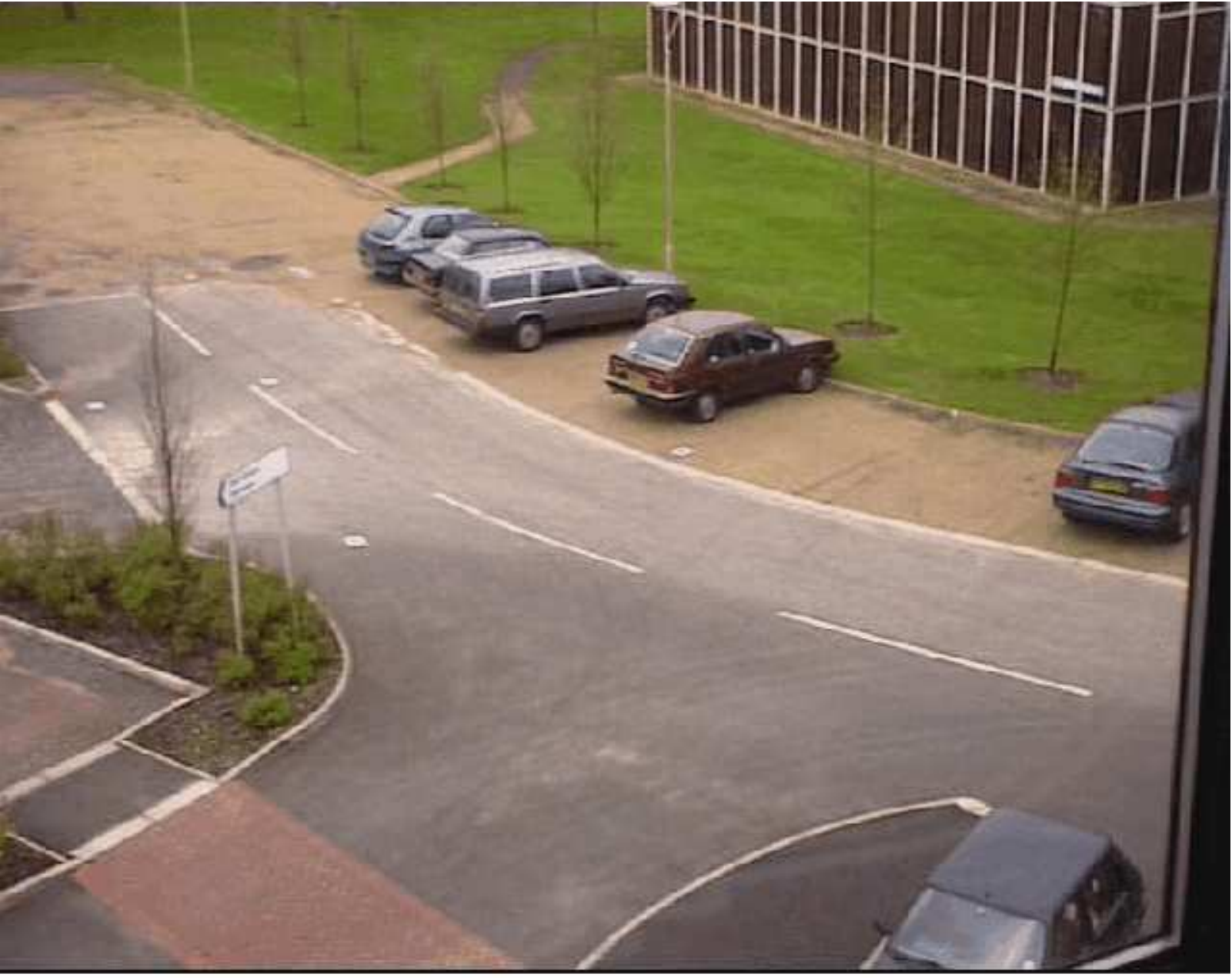}}
  \caption{ An illustration for retrieving background. (a) shows contaminated background
            with foreground regions signed by red rectangles; (b) is the frame where the
            background patches are obtained, note that blue rectangles indicate the
            foregrounds in (b), they are far from the foreground areas in (a); (c)
            demonstrates a retrieved background based on (a) and (b). The frames are
            captured from video sequence \emph{pets2000\_c1}.
           }
       \label{fig:patch}
  \end{figure}

  An intuitive way to obtain a clean background is replacing the foreground of
  one frame with a background patch cropped from another frame. More precisely, let $F
  \in \mathbb{R}^{h \times w}$ denote the frame based on which the background is
  retrieved, and $F'\in \mathbb{R}^{h \times w}$ stand for the frame where the background
  patch is cropped, suppose that the foreground region in $F$ is $F(t : b, l :
  r)$\footnote{In this paper, all the target or foreground is represented as a rectangle
  region}, the patching operation could be described as 
  \begin{equation}
    \Gamma_{i, j} = \left\{{\begin{array}{cc}
      F'_{i, j}, & t \leq i \leq b ~\& ~l \leq j \leq r\\
            F_{i, j}, & \text{otherwise}.  \\
                      \end{array} }\right.
      \label{equ:patch}
  \end{equation}
  where $\Gamma$ is the retrieved background. An illustration of \eqref{equ:patch} is also
  available in Figure~\ref{fig:patch}.

  In practice, multiple foreground regions need to be mended for each ``impure''
  background candidate. Furthermore, a selection approach should be conducted to form the
  optimal combination of the retrieved backgrounds over all the candidates. To achieve
  this goal, we first randomly capture $N' > N_b$ frames from the concerned video
  sequence. Afterwards, every foreground region of the frames are located manually. The
  foreground is then replaced by a clean background region cropped from the nearest frame
  (in terms of frame index). Finally, a $k$-median clustering algorithm is carried out for
  selecting $N_b$ most comprehensive backgrounds. 
  
  It is nontrivial to notice that even some backgrounds are not perfectly retrieved, \ie,
  with minor foreground remains, CSBM can still work well considering that CS is robust to
  the noise in measurements \cite{Candes_CPAM_05_Stable}.  

\subsection{Equiping RTCST with CSBM}
\label{subsec:equip_back}
  We equip the RTCST with CSBM to build a novel visual tracker, \emph{a.k.a Real-Time CS-based
  Tracker with Background Model} (RTCST-B). In RTCST-B,  original noise templates are
  replaced by \emph{background templates} which are generated
  from CSBM. In the context of PF tracking, given a observation position $\Xi$ with $d_0$
  pixels and a CSBM $\mathbb{G}$ defined in \eqref{equ:def_cbm}, the background templates
  set $B$ is obtained by: 
  \begin{equation}
    \begin{split}
      B & = [I_1~I_2~\dots~I_{N_b}] \in \mathbb{R}^{d_0 \times N_b}  \\
      I_i & = \text{CV}(\Gamma_i,~ \Xi) ~\forall i = 1, \dots, N_b
    \end{split}
    \label{equ:def_back_temp}
  \end{equation}
  where function $\text{CV}(\cdot)$ is called \emph{crop-vectorize} operation which first
  crops the region indicated by $\Xi$ from background $\Gamma_i$ and then vectorize it
  into $I_i \in \mathbb{R}^{d_0}$. Eventually, the optimization problem
  for RTCST-B writes:
  \begin{equation}
      \min_{\x}\|\x\|_{0} ~
      \sst \; \|\Phi A\x - \Phi \y\|_{2} \leq \varepsilon, 
  \label{equ:rtcst_b_opt}
  \end{equation}
  where $\x$ is comprised of $\x_t$ and $\x_b$, \ie, the coefficient vectors for target
  and background, $A=[T,~B] \in \mathbb{R}^{d \times (N_t + N_b)}$.
  
  Despite the diverse optimization problem, the calculation for the likelihood remains the
  same as in \eqref{equ:rtcst_lklhd}. To understand this, let $\x_t$ and $\x_b$ denote the
  coefficients associated with target templates and background templates respectively,
  $p(\y_k | \s) = p(\y_k | \x_t) = \exp(-\lambda r)$ be the observation
  likelihood\footnote{It is trivial to prove that the relationship between particle $\s$
  and $\x_t$ is deterministic given a specific frame image}, where $r$ is defined in
  \eqref{equ:rtcst_residual}, then we have:
  \begin{equation}
      p(\y_k | \x_t, \x_b) = p(\y_k | \x_t) = \exp(-\lambda r)
    \label{equ:proof_information}
  \end{equation}
  with the assumption that $\x_t$ and $\x_b$ are deterministic by each other, \ie,
  \begin{equation}
    p(\x_b, \x_t) = p(\x_t) = p(\x_b)
    \label{equ:recovery_approx}
  \end{equation}
  or in other words, the solution of CS procedure is unique.
  \cite{Candes_CPAM_05_Stable}.
  
  In addition, the template update scheme should be changed slightly considering a new
  class is involved in. More precisely, target templates are updated only when 
  \begin{equation}
    \text{SCI}_{tb}(\x) = \frac{\max\{\|\x_t\|_1,
    \|\x_b\|_1\}}{\|\x\|_1} \leq \tau 
    \label{equ:new_sci}
  \end{equation}

  Finally, the positive constraint for $\x$ is removed in \ref{equ:rtcst_b_opt} because 
  background subtraction implies minus coefficients for background templates. It is
  reasonable to not curb the coefficients in RTCST-B.  

  In summary, one just needs to impose following minor modifications on RTCST to transfer it
  into RTCST-B. 
  \begin{enumerate}
    \item Substitute the background templates for noise templates.
    \item Eliminate the positive constraint.
    \item Conduct the CV operation for each observation.
    \item Utilize the new SCI measurement.
  \end{enumerate}
  Apparently, the diversity between RTCST and RTCST-B is not significant with respect to
  formulation. Nevertheless, the seemingly small change makes RTCST-B much more
  superior to its prototypes.

  \subsection{Superiority Analysis}
  \label{subsubsec:rtcst_b_advantages}
  Compared with the $\ell_1$ tracker and RTCST, RTCST-B enjoys three main advantages which are
  described as follows.
  \subsubsection{More Sparse}
  \label{subsubsec:more_strict}
      An underlying assumption behind the $\ell_1$ tracker and RTCST is that, the
      background could be sparsely represented by noise templates in $E$. It is true
      when foreground dominates the observed rectangle. More quantitatively, given
      $\eta_t$ is the sparsity of target coefficient vector $\x_t$, when \[\eta_t +
      \|\x_e\|_0 \leq d / 3\] the representation based on solution $\x$ in
      \eqref{equ:rtcst_opt} is guaranteed to be reliable \cite{Wright_PAMI_09_Face}.
      Nonetheless, the sparse representation is no longer valid when the background
      covers the main part of observation.  Predictably, the incorrect representation
      will deteriorate tracking accuracy. 
          
\begin{figure}[h]
          \subfigure[]{\includegraphics[width=0.24\textwidth]{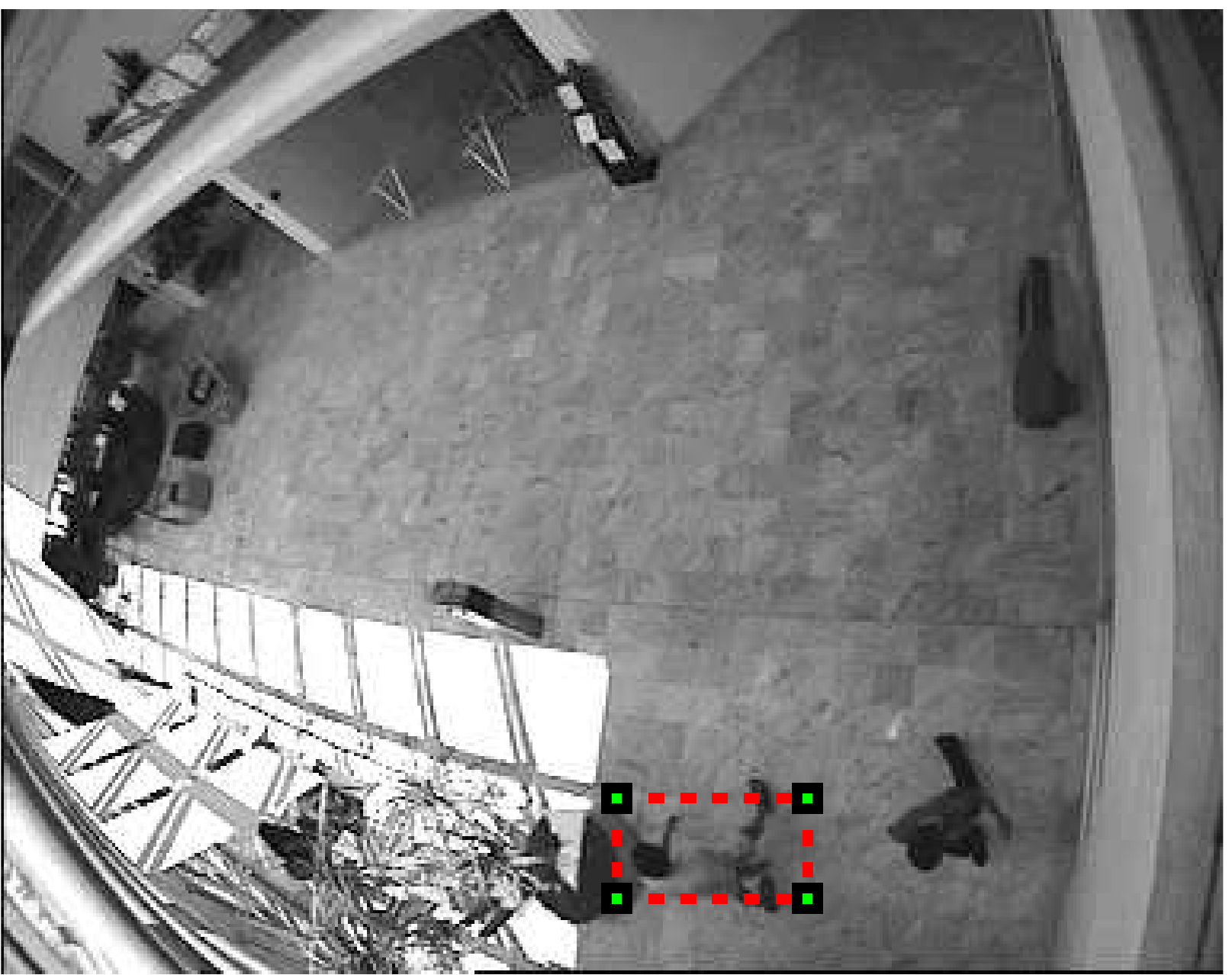}}
          \subfigure[]{\includegraphics[width=0.24\textwidth]{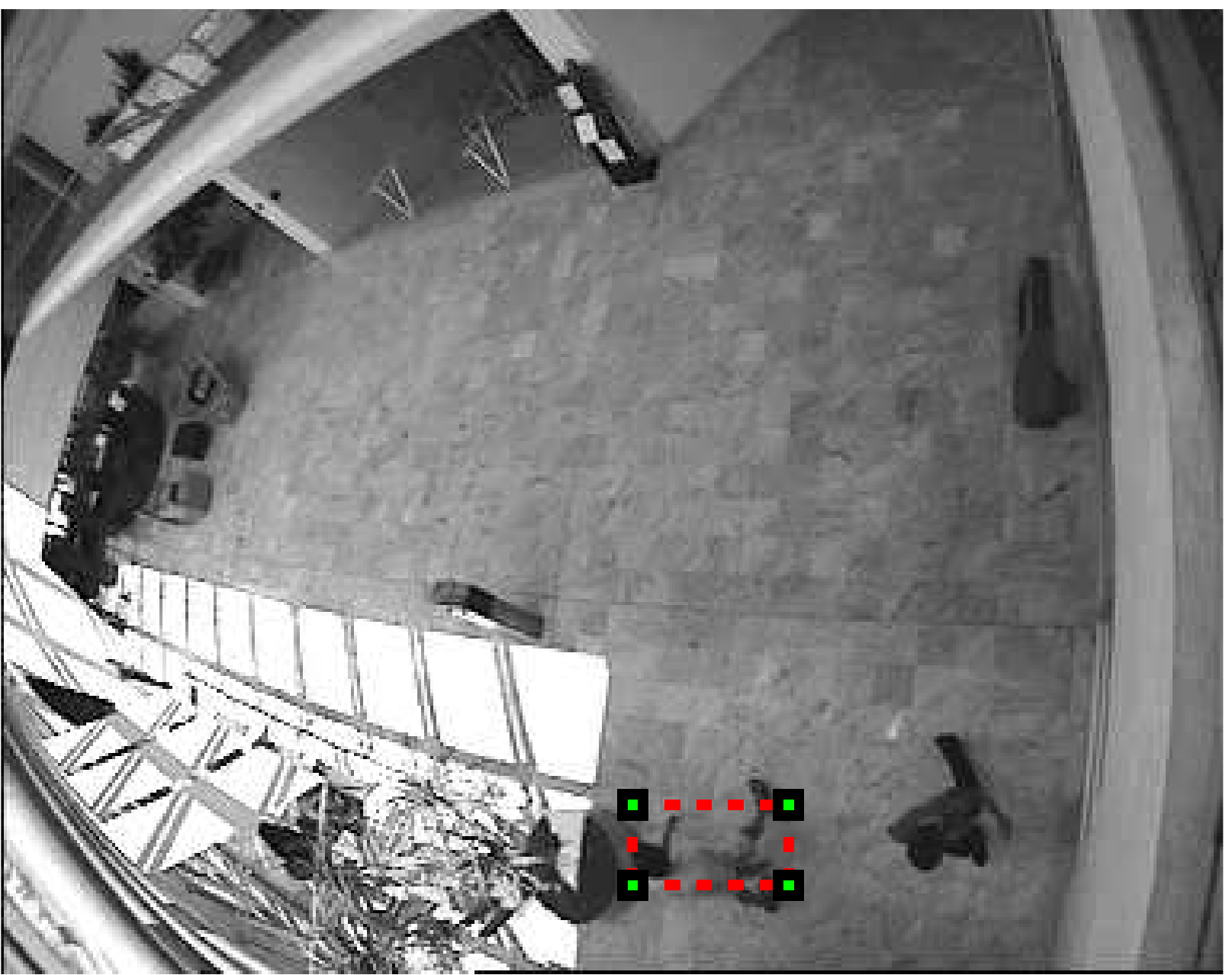}}
          \subfigure[]{\includegraphics[width=0.24\textwidth]{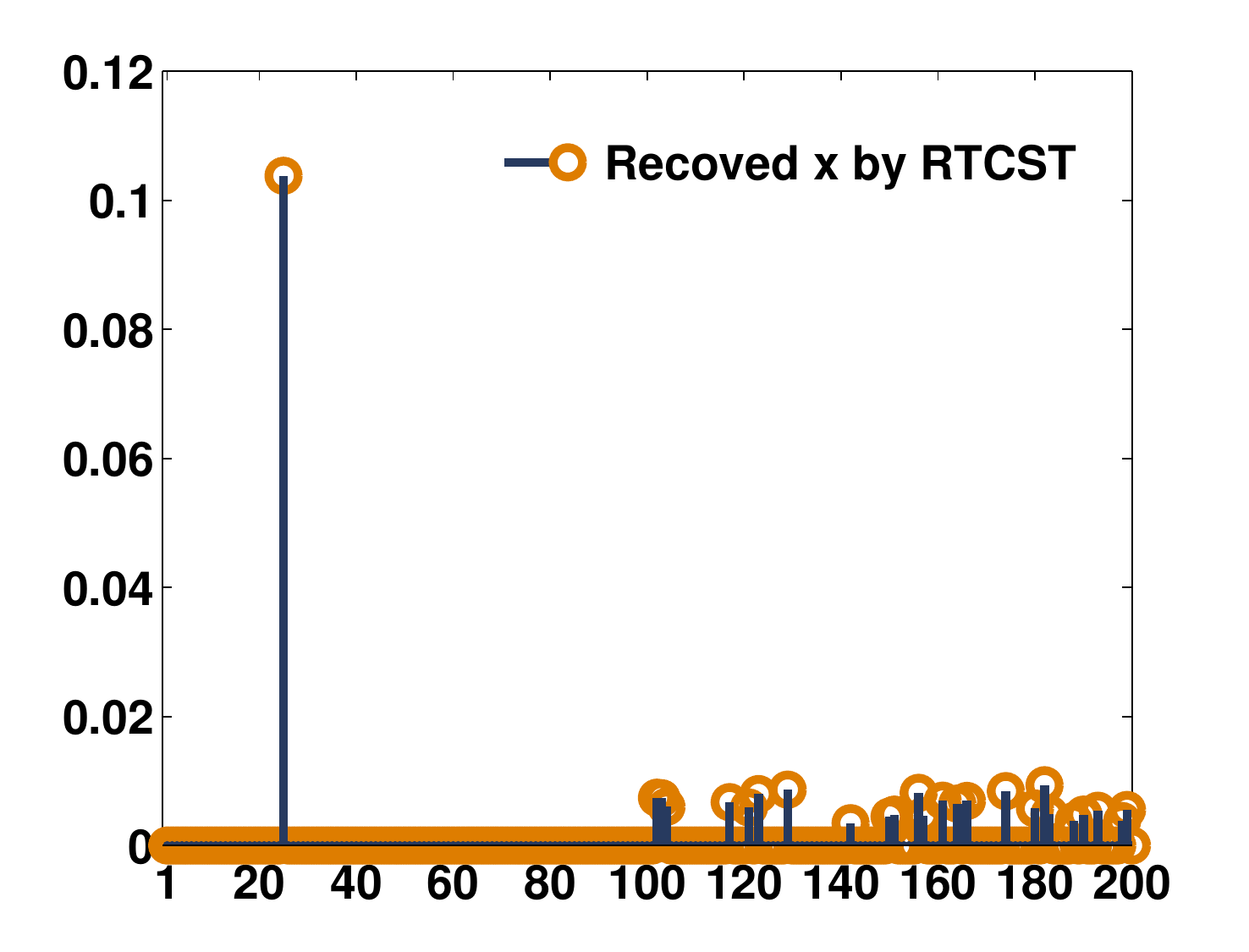}}
          \subfigure[]{\includegraphics[width=0.24\textwidth]{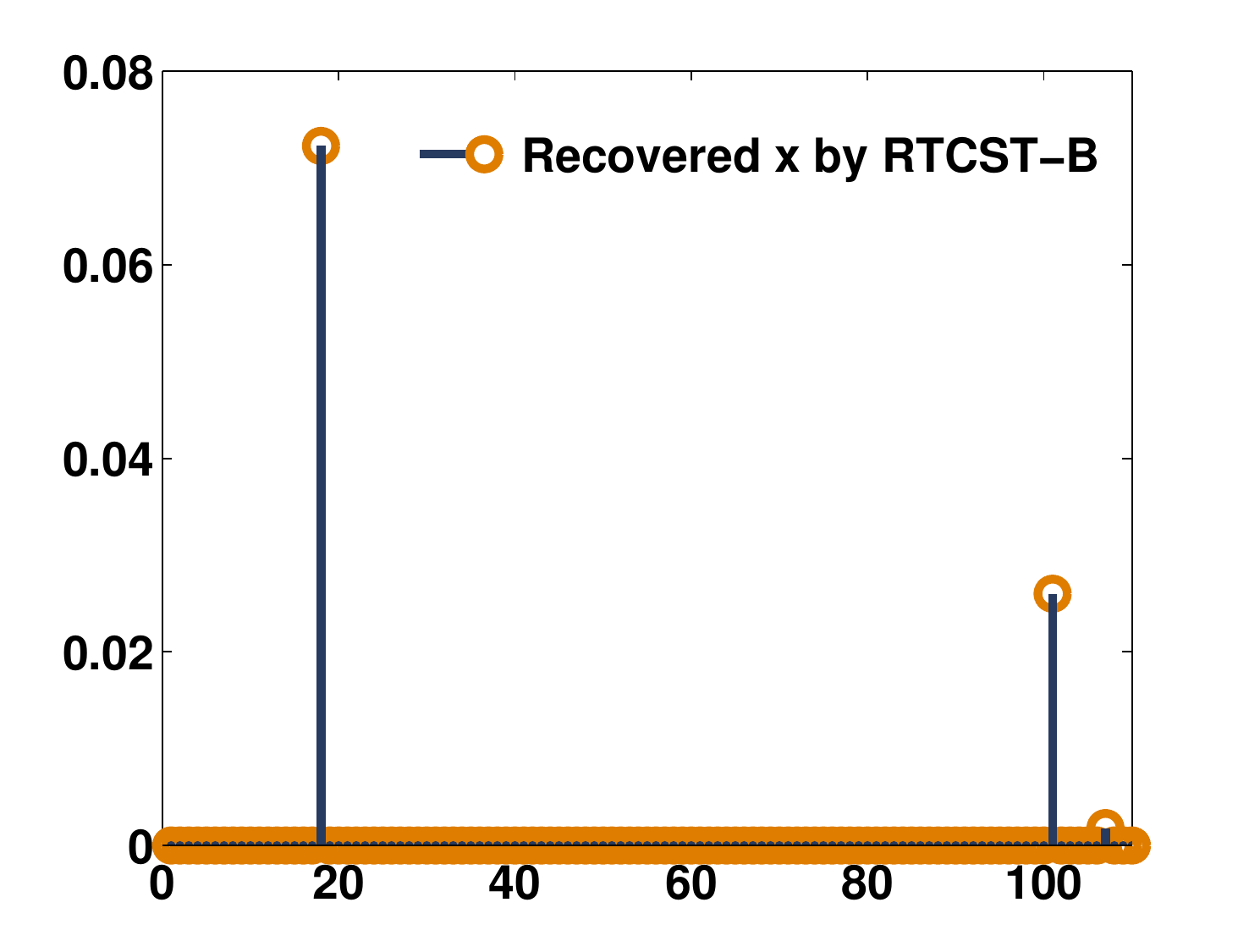}}
      \caption{ A demonstration of the sparse solutions for RTCST and RTCST-B. (a) and
                (b) are the tracking result by RTCST and RTCST-B on the same frame
                (captured from \emph{pets2004\_p1}). (c) and (d) are the recovered signals for
                RTCST and RTCST-B respectively. The representation by RTCST-B is much
                more sparse than that by RTCST. Note that here, $d = 50$, $N_t = 100$
                and $N_b = 10$ for RTCST-B.
               }
           \label{fig:rtcst_b_sparse}
      \end{figure}

      On the other hand, after noise templates being replaced by background templates,
      the aforementioned assumption usually keeps true.
      Figure~\ref{fig:rtcst_b_sparse} give us a explicit demonstration for the
      sparsity of solutions.

  \subsubsection{More Efficiency}
  \label{subsubsec:more_efficient}
      Comparing with existing background models, the computation burden of CSBM is
      extremely trivial. First of all, there is no need to conduct the background
      subtraction or foreground connection in RTCST-B, because these two functions are
      integrated within the CS procedure implicitly. Secondly, if the CSBM is generated
      properly, \ie, can cover the main distribution of background's appearance, to update
      model becomes unnecessary. Thirdly, the sufficient number of background templates is
      much smaller than that of noise template, \ie, \[N_b \ll N_n = 2d\] where $N_n$ is
      the number of noise templates. The reduction of templates' amount will immediately
      speed up the optimization process. The last, and the most important reason is, the
      required sparsity $\eta$ for RTCST-B is much smaller than that for RTCST (see
      Section~\ref{susubbsec:inf_rcst}). This leads to an earlier terminated OMP procedure
      in RTCST-B and hence makes it faster. In conclusion, the introduction of CSBM won't
      impose further computational burden on the algorithm, and just the opposite, the
      tracking procedure will be accelerated to some extent.

  \subsubsection{More Robust}
  \label{subsubsec:more_robust}
      In RTCST and $\ell_1$ tracker, one tries to use noise templates $E = [I -I]$ to
      represent background. However, it is the columns in $I$, which is called
      \emph{standard basis vectors}, doesn't favor background images over targets.
      This character makes RTCST and $\ell_1$ tracker powerless for recognizing
      background and consequently, decreases the tracking accuracy. Differing from the
      prototype, RTCST-B harnesses the discriminant nature of CS-based pattern
      recognition. Both foreground (target) and background are treated as a typical
      class with distinct features. In RTCST-B, target templates compete against
      background templates, who are as powerful as their competitors, to ``attract''
      the observation.  Intuitively, the more discriminative templates will make
      RTCST-B more robust.
          
\begin{figure}[h]
          \subfigure[]{\includegraphics[width=0.118\textwidth,clip]{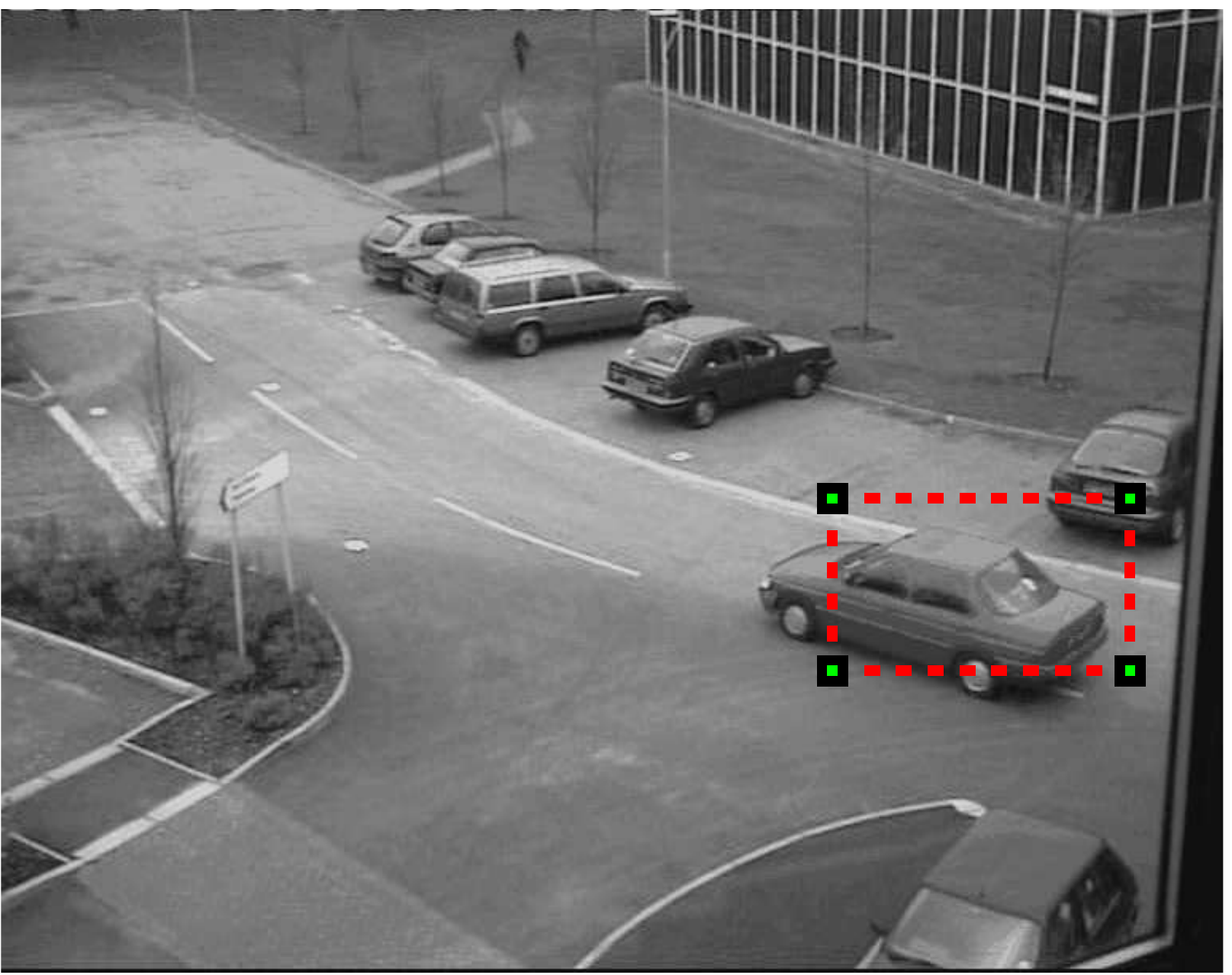}}
          \subfigure[]{\includegraphics[width=0.118\textwidth,clip]{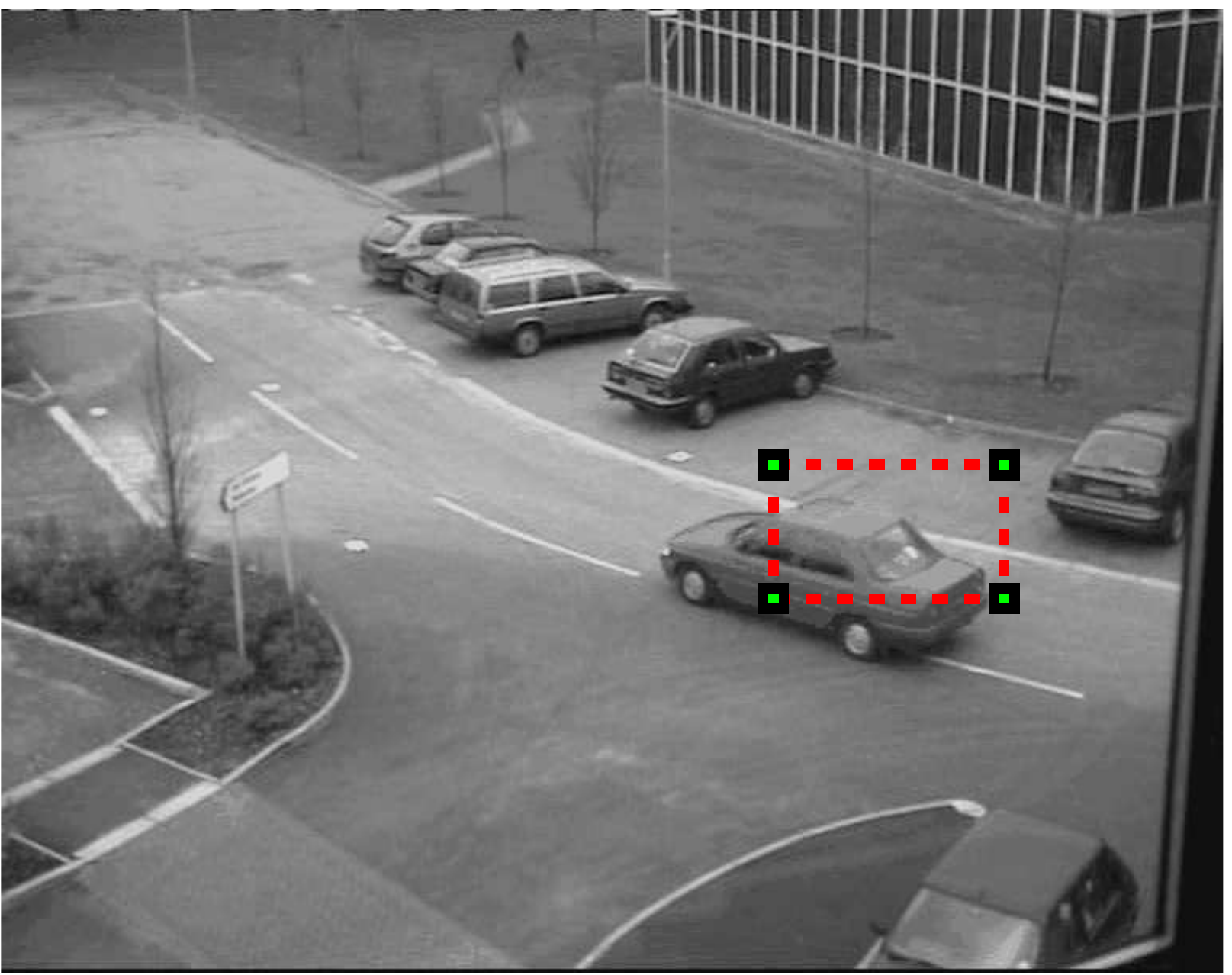}}
          \subfigure[]{\includegraphics[width=0.118\textwidth,clip]{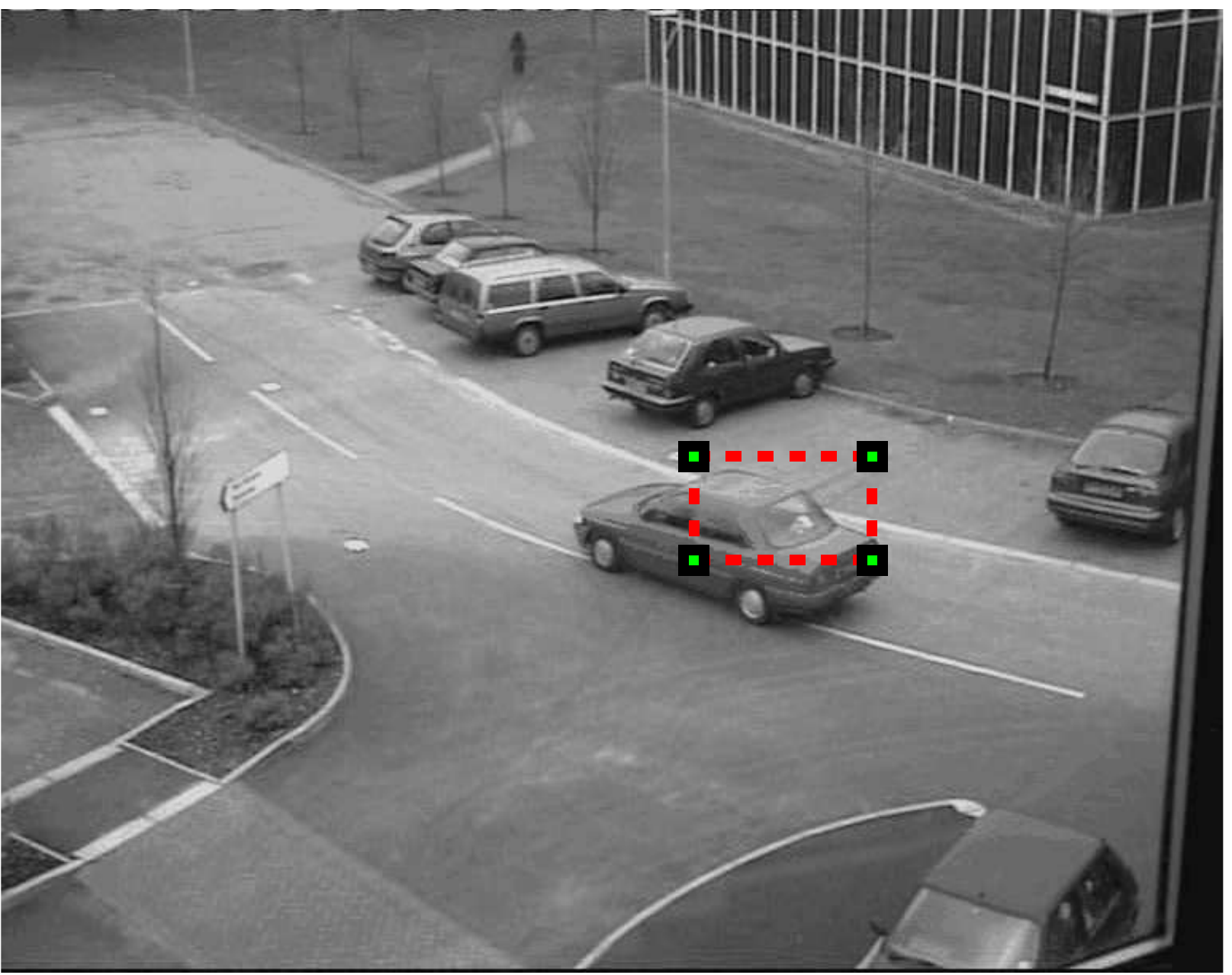}}
          \subfigure[]{\includegraphics[width=0.118\textwidth,clip]{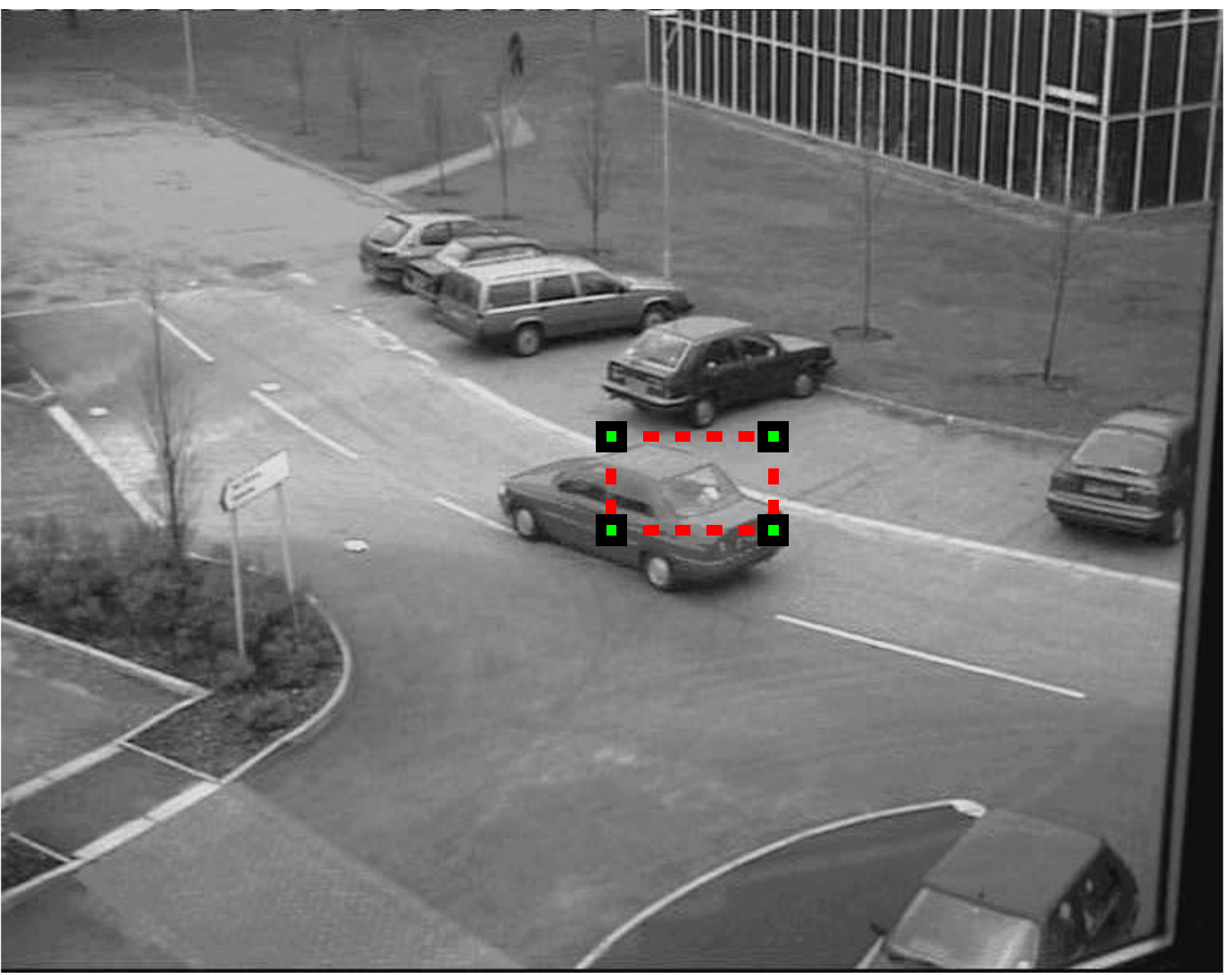}}
          \subfigure[]{\includegraphics[width=0.118\textwidth,clip]{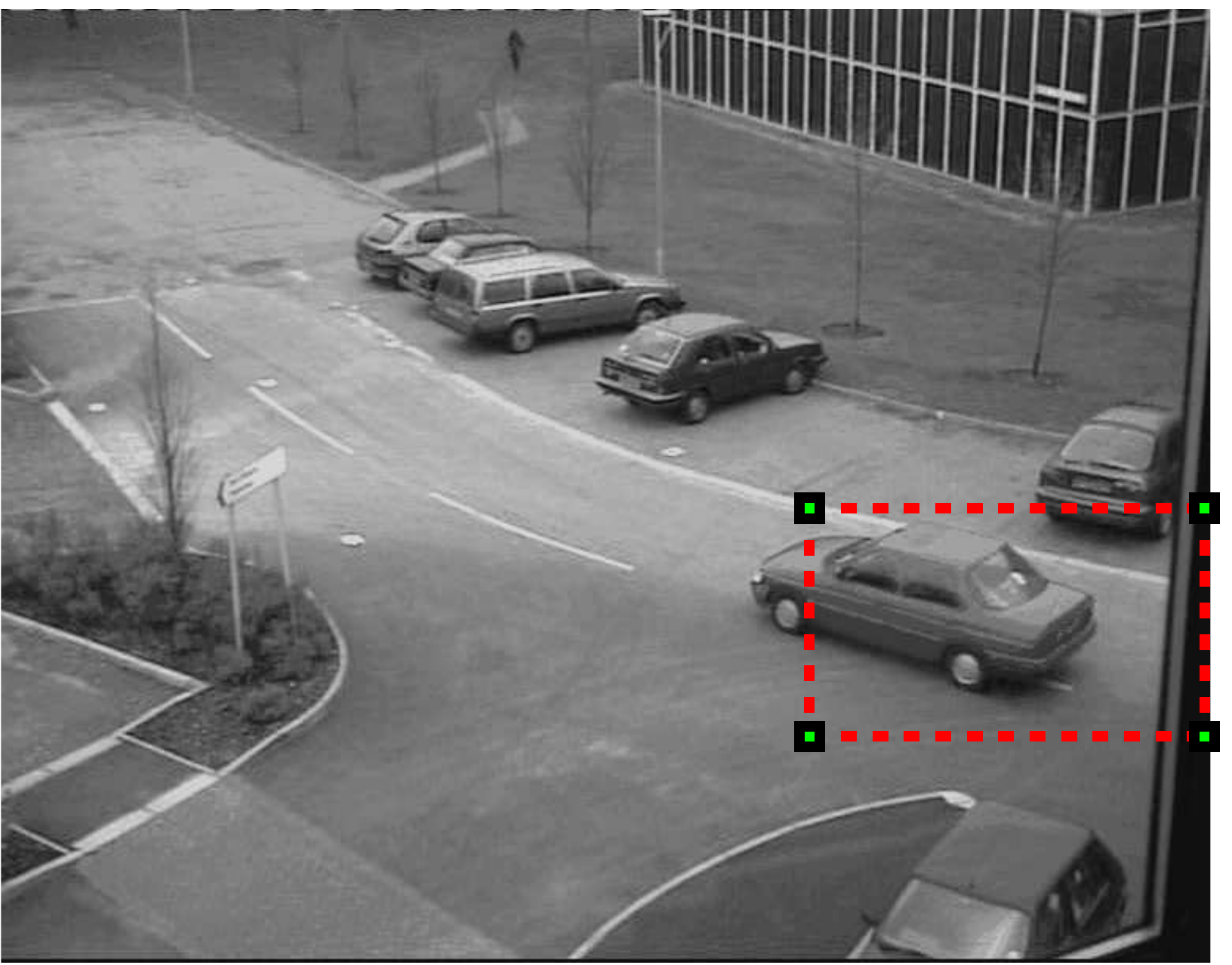}}
          \subfigure[]{\includegraphics[width=0.118\textwidth,clip]{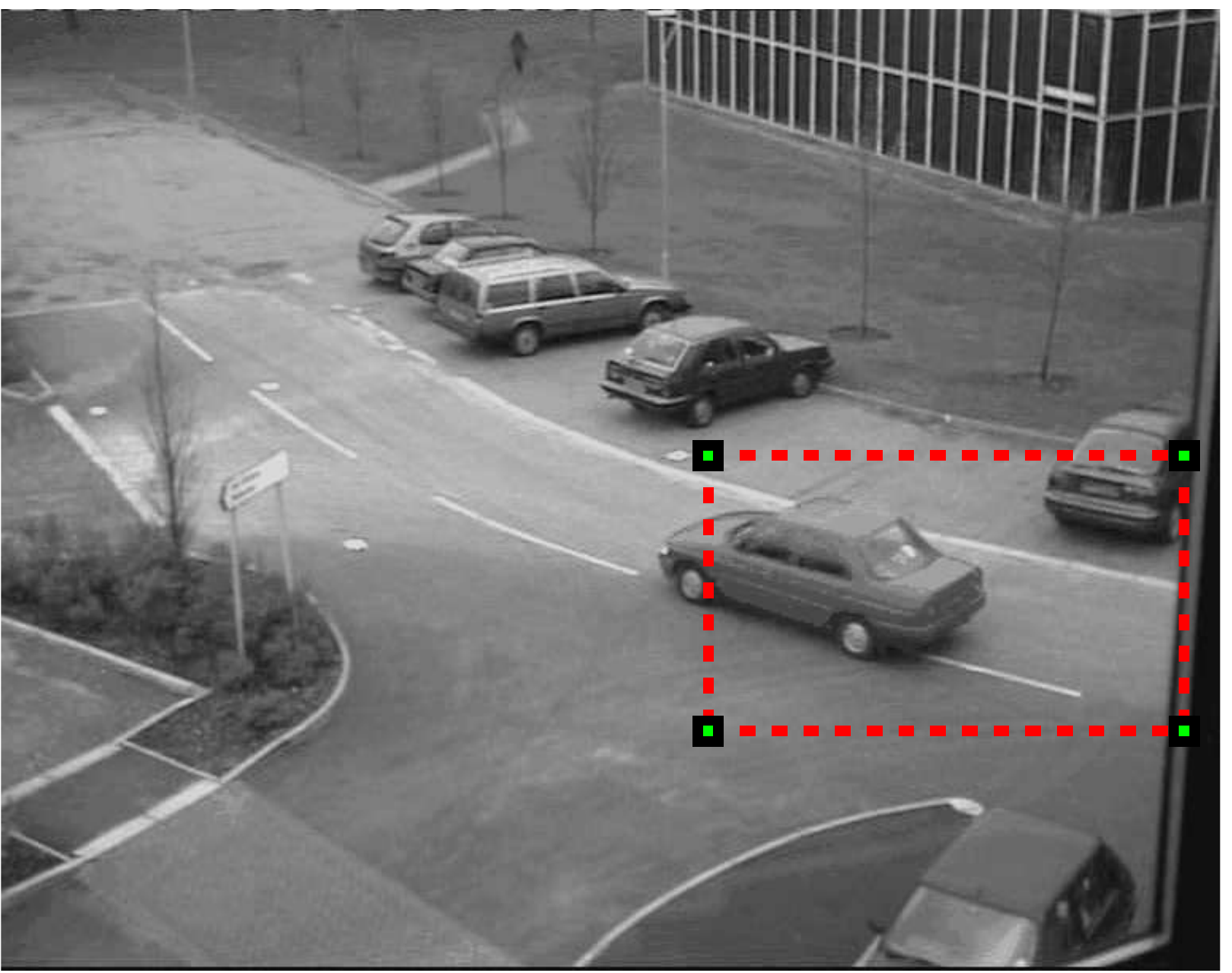}}
          \subfigure[]{\includegraphics[width=0.118\textwidth,clip]{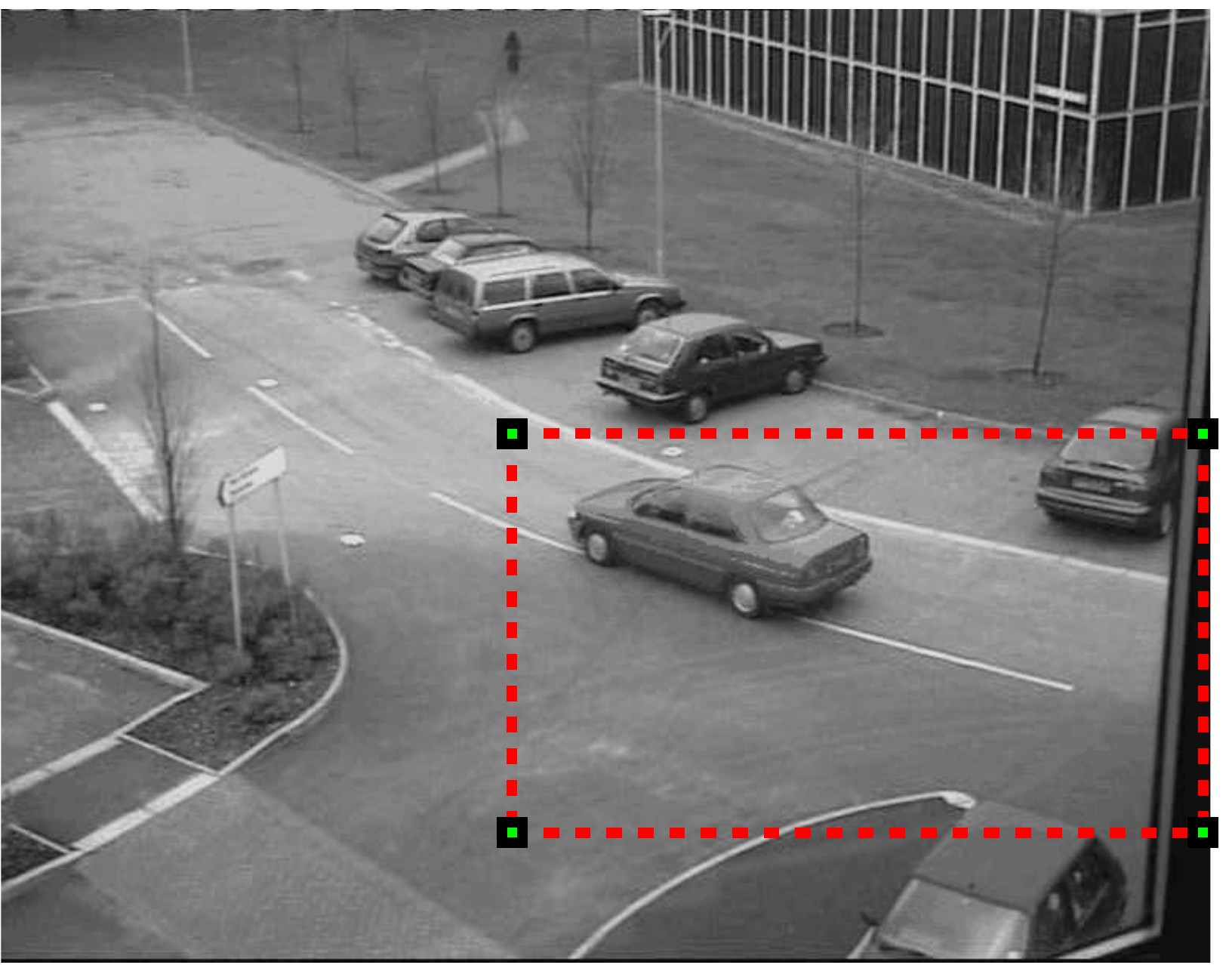}}
          \subfigure[]{\includegraphics[width=0.118\textwidth,clip]{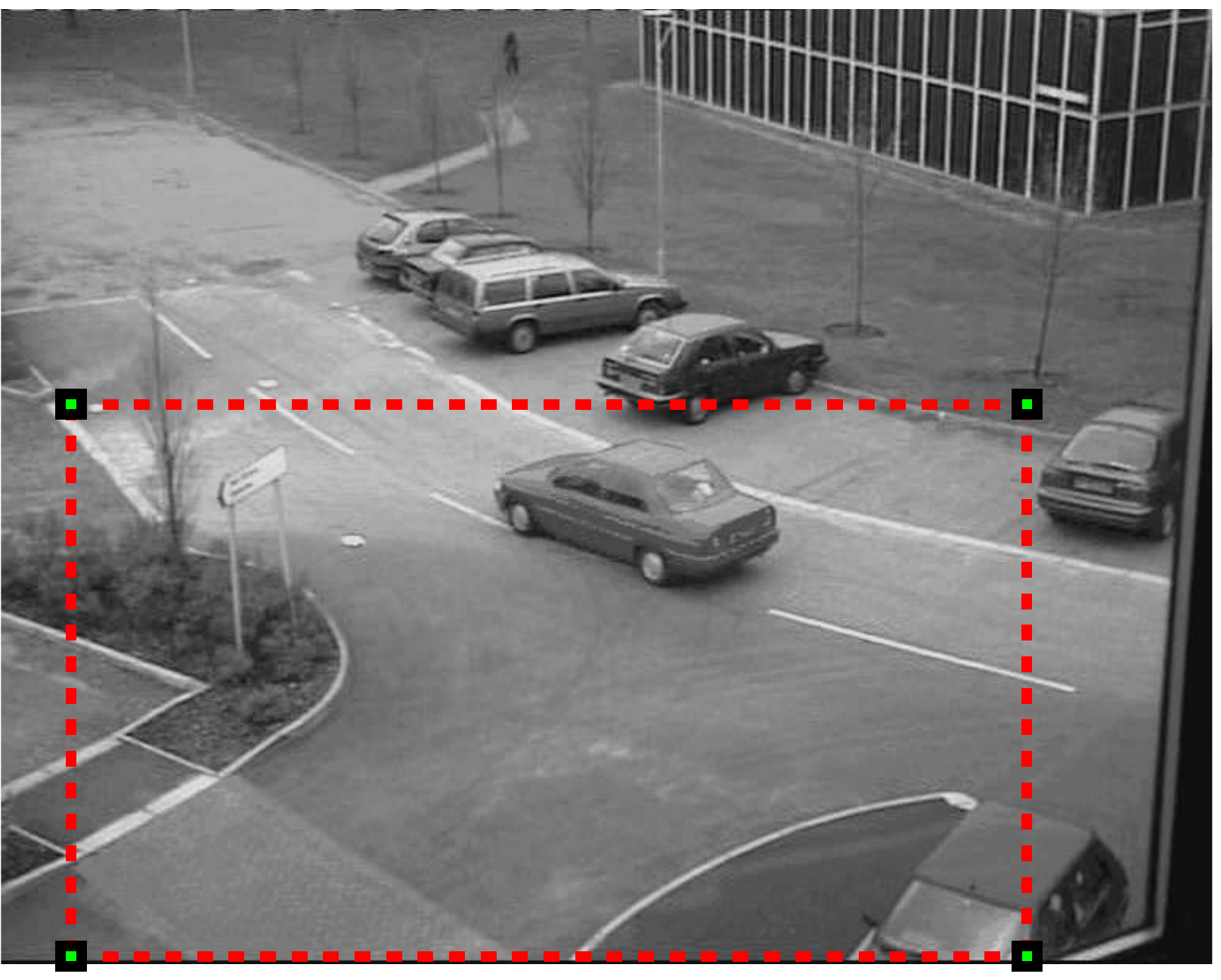}}
      \caption{ An empirical evidence for the robustness of RTCST-B against drift. (a) to
                (d) are the tracking results for RTCST-B compared with image (e) to
                (h) which are the results for RTCST on the same frames from
                \emph{pets2000}.
                The tracked target is signed by red rectangle. We can see that a drift
                tendency shown on (b) is curbed in the successive frames. Quite the
                contrary, in the bottom line, the drift effect grows dramatically.
               }
           \label{fig:rtcst_b_robust}
      \end{figure}

      Moreover, once the tracked region drifts away, background information would be
      brought into target templates via template update (which is almost unavoidable).
      In this situation, for RTCST and $\ell_1$ tracker, some target templates could
      be more similar to background than all the noise templates. This leads to a
      serious classification ambiguity and therefore, poor tracking performance.
      Quite the contrary, RTCST-B could draw back the target to the correct position
      thanks to the capacity of recognizing background. In plain words, RTCST-B always
      tends to locate the target in the region which doesn't \emph{look like}
      background. An empirical evidence for the robustness of RTCST-B is shown in
      Figure~\ref{fig:rtcst_b_robust}.

\section{Experiment}
\label{sec:exp}

  \subsection{Experiment Setting}

  To verify the proposed tracking algorithms, we design a series of experiments for
  examining the tracking algorithm in terms of accuracy, efficiency and robustness. The
  proffered algorithms are conducted on $10$ video sequences comparing with $\ell_1$
  tracker, Kernel-Mean-Shift (KMS) tracker \cite{Comaniciu_PAMI_03_Kernel} and color-based
  PF tracker\cite{Sanjeev_TSP_02_PF}. The details of selected video sequences are list in
  Table~\ref{tab:video_sequence}. Note that we only conduct $\ell_1$ tracker on $5$ videos
  which are \emph{cubicle}, \emph{dp}, \emph{car11}, \emph{pets2001\_c1} and
  \emph{pets2004-2\_p1} respectively. It is because for other videos, the convex
  optimization problem is too slow to be solved (above $5$ minutes per frame). 

\begin{table}[h]
\caption
{
  The details of video sequences which are employed for our experiment. The tracking
  frames refer to the concered frame index for each video; initial position indicates the
  minimum bounding box for the target in the first frame; if ``Yes'' shows in the
  last column, the video is captured from a stationary camera and consequently, it suits
  RTCST-B.
}
  \resizebox{0.5\textwidth}{!}
  {
  \begin{tabular}{ c | l | l | c }

  \hline\hline
  & tracking frames & initial position & stationary camera \\
  \cline{1-4}
  \bf cubicle & $1 \sim 51$ & [56, 24, 90, 67] & No\\
  \bf dp & $1 \sim 66$ & [91, 25, 116, 57] & No\\
  \bf car4 & $1 \sim 300$ & [139, 102, 356, 283] & No\\
  \bf car11 & $1 \sim 393$ & [69, 123, 104, 157]& No\\
  \bf fish & $1 \sim 200$ & [122, 57, 208, 148]& No\\
  \bf pets2000\_c1 & $122 \sim 312$ & [536, 318, 743, 432] & Yes\\
  \bf pets2001\_c1 & $1550 \sim 1635$ & [8, 272, 46, 296] & Yes\\
  \bf pets2002\_p1 & $275 \sim 500$ & [578, 92, 641, 172] & Yes\\
  \bf pets2004\_p1 & $115 \sim 550$ & [193, 258, 251, 287] & Yes\\
  \bf pets2004-2\_p1 & $1 \sim 201$ & [181, 224, 239, 262] & Yes\\
  \cline{1-4}
  \hline\hline
\end{tabular}
}
\label{tab:video_sequence}
\end{table}

  There are two alternative dimension-reduction manners for RTCST and RTCST-B, namely,
  random projection and hash matrix projection. In our experiments, both of them are
  performed with reduced dimension $25$, $50$ and $100$. As regards the particles' number,
  we examine the proposed trackers with $100$ and $200$ particles and the numbers for PF
  tracker is $100$, $200$ and $500$. All the PF-based trackers are run for $20$ times
  except $\ell_1$ tracker which is merely conducted for $3$ times. We perform KMS tracker
  for only $1$ time considering it is a deterministic method. The average values and
  standard errors are reported in this section. The MS tracker, PF tracker and $\ell_1$
  tracker are implemented in C++ while our CS-based trackers are implemented in Matlab. To
  compare the efficiency with the proposed algorithms, there is also a Matlab version of
  $\ell_1$ tracker. All the algorithms are run on a PC with $2.6G$Hz quad-core CPU and
  $4G$ memory (we only use one core of it). As to the software, we use Matlab
  $2009a$ and the linear programming solver is called from Mosek $6.0$\cite{Mosek}.

  It is important to emphasize that in our experiment, \emph{no trick is used for
  selecting the target region in the first frame}. The initial target region is always the
  minimum rectangle $R = [l, r, t, b]$ which can cover the whole target\footnote{Shadows
  are not taken into consideration.}, where $l$, $r$, $t$, and $b$ are the left, right,
  top and bottom boundaries' coordinates (horizontal or vertical) respectively. This rigid
  rule is followed for eliminating the artificial factors in visual tracking and making
  the comparison unprejudiced. 

  \subsection{TSP --- A New Metric of Tracking Robustness}

\begin{figure}[h]
      \subfigure[tracking error]{\label{subfig:dis}\includegraphics[width=0.24\textwidth,
      height=0.18\textwidth]{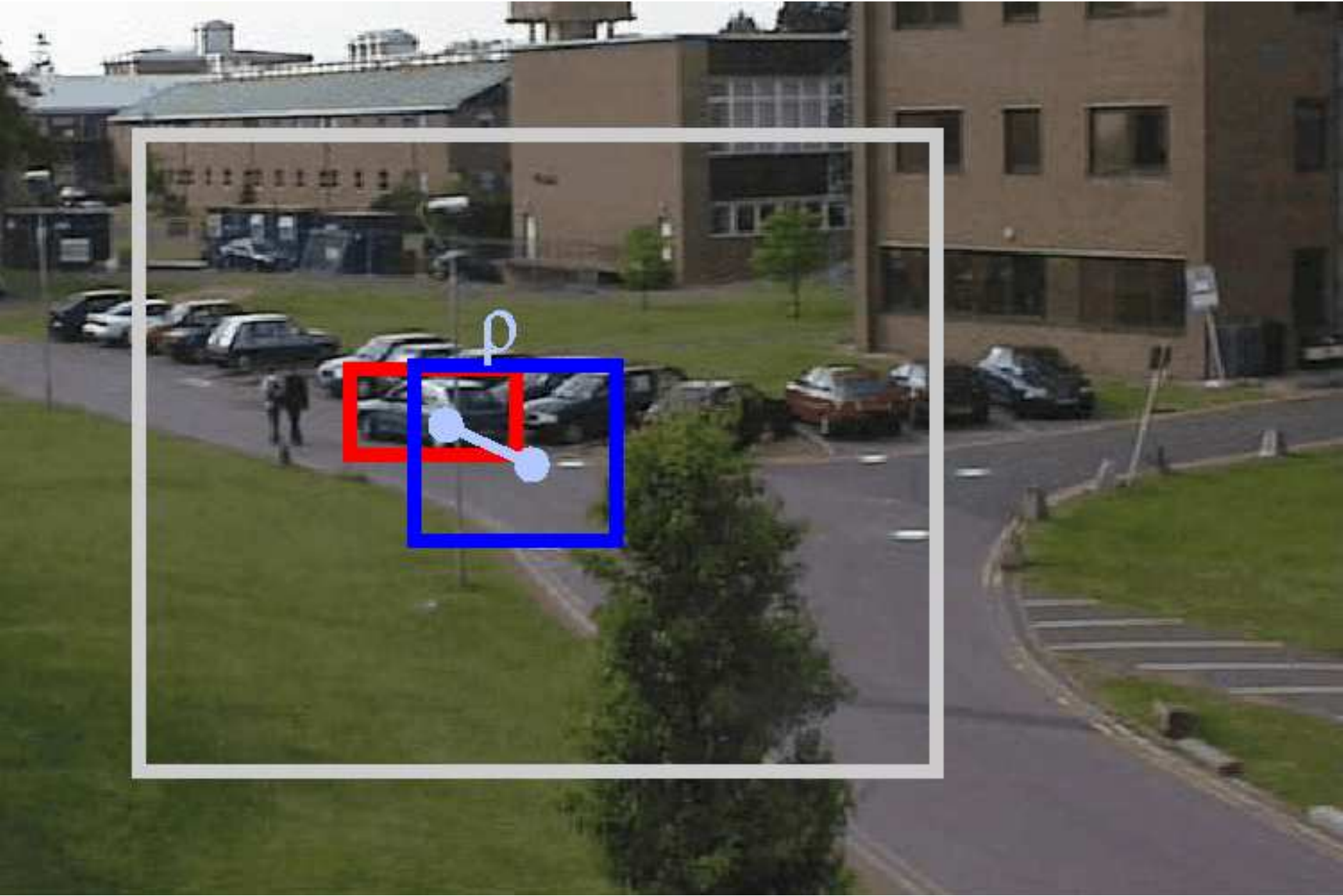}}
      \subfigure[TSP]{\label{subfig:tsl}\includegraphics[width=0.24\textwidth,
      height=0.18\textwidth]{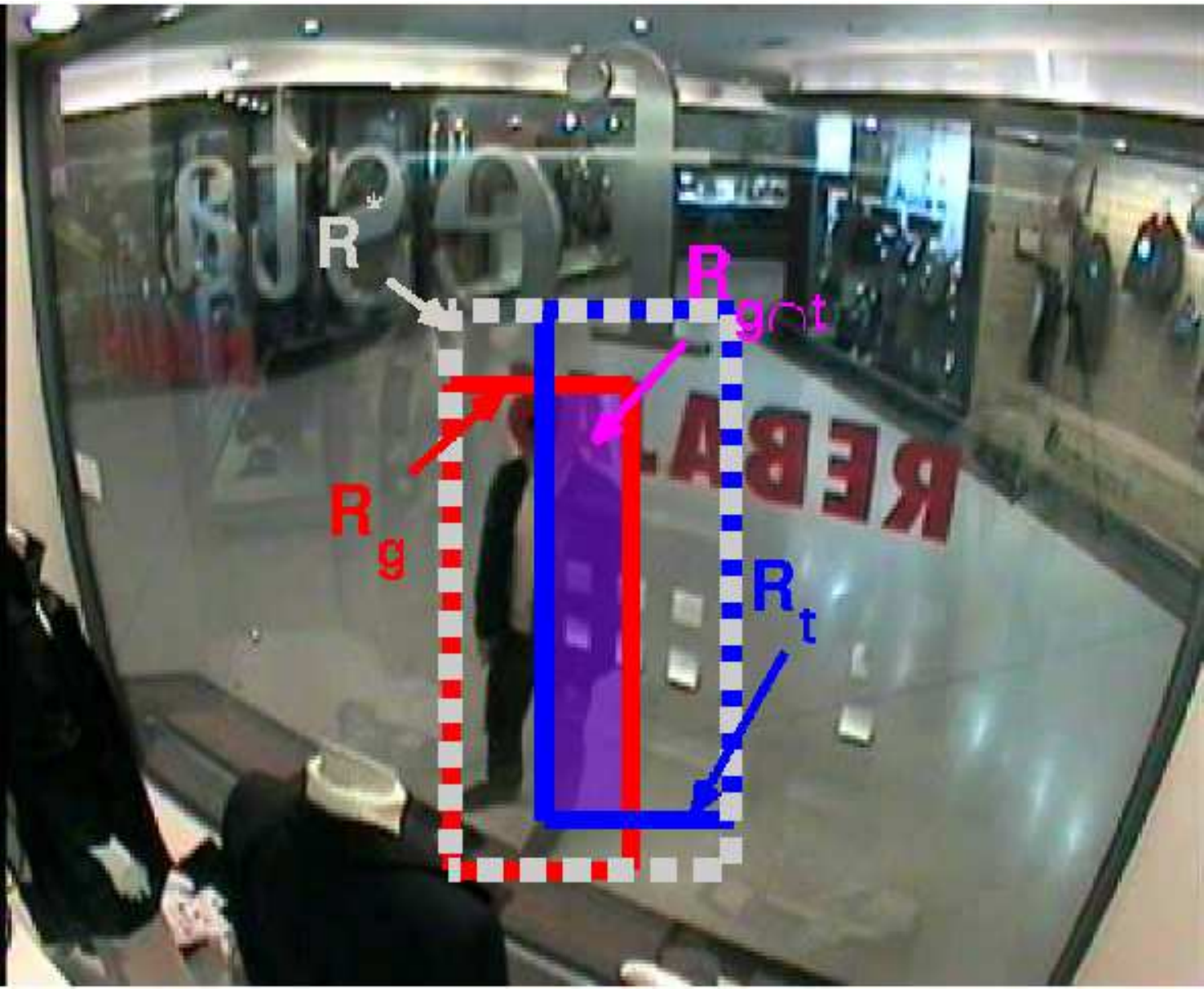}}
    \caption{A demonstration of two measurements of tracking accuracy. (a) shows the poor
            capacity of $\rho$. (b) illustrates the definition of TSP. $R_g$ and $R_t$ are
            illustrated as red and blue rectangles respectively; the region 
            $R^*$ is a gray dashed square in the image while intersection region
            $R_{g\cap t}$ is shown in purple. We can see that in this case, $a(R_g,
            R_d) = R_{g\cap t} / R^*$. These two frames are obtained from video
            sequence \emph{pets2001} and \emph{pets2002} respectively.}
         \label{fig:tsl}
    \end{figure}

    A conventional choice of the manner to verify the tracking accuracy is \emph{tracking
    error}. Specifically, given that the centroid of ground
    truth region is $\c_g$ while that of tracked region is $\c_t$, the tracking error
    $\rho$ is defined as
    \begin{equation}
    \setlength{\abovedisplayskip}{0.1cm}
    \setlength{\belowdisplayskip}{0.1cm}
        \rho = \|\c_g - \c_t\|_2,
        \label{equ:pos_error}
    \end{equation}
    \ie, the euclidean distance between two centroids. However, if we take scale variation
    into consideration, $\rho$ is poor to verify tracker's performance. Let's see
    Figure~\ref{subfig:dis} for a example. In the image, red rectangle indicates the
    ground truth for a moving car. The blue and gray rectangles, which are obtained by 
    various tracking algorithms, share the identical centroid. By using tracking error,
    same performance is reported for both two trackers despite the obvious difference on
    tracking accuracy. 
    
    Inspired by the evaluation manner proposed for PASCAL data
    base\cite{Everingham_PASCAL_07_Detection}, we propose a new tracking accuracy
    measurement which is termed \emph{Tracking Success Probability} (TSP). To obtain the
    definition of TSP, firstly let's suppose the bounding box of ground truth region is
    $R_g = [l_g, r_g, t_g, b_g]$, and the one for tracked region is $R_t = [l_t, r_t, t_t,
    b_t]$.  We then design a function $a(R_g, R_t) \in [-1, 1]$ to estimate the
    overlapping state between $R_g$ and $R_t$. Given two distance sets: 
    \begin{equation*}
      \begin{split}
        \mathbb{H} & = \{r_t - l_g, r_g - l_t, r_g - l_g, r_t - l_t\} \\
        \mathbb{V} & = \{b_t - t_g, b_g - t_t, b_g - t_g, b_t - t_t\} \\
      \end{split}
    \end{equation*}
    and a indicator function $s_{tg}$
    \begin{equation}
      s_{tg}:= \left\{{\begin{array}{*{20}c}
        -1,\;\;&R_g \text{ and } R_t \text{ are seperate} \\
        1,\;\;&\text{otherwise}.  \\
                      \end{array} } \right.
      \label{equ:s}
    \end{equation}
    then $a(R_g, R_t)$ writes\footnote{Here, we suppose the origin of image is on the
    left-top corner.}
    \begin{equation*}
      a(R_g, R_t) = s_{tg} \cdot \left|\frac{min(\mathbb{H})\cdot
      min(\mathbb{V})}{max(\mathbb{H})\cdot max(\mathbb{V})}\right|,
    \end{equation*}
    It is easy to find that when two regions overlap each other, $a(R_g, R_t)$ is the
    ratio of the intersection area $R_{g\cap d}$ to the area $R^*$, which is the minimum
    region covering both $R_g$ and $R_d$. See Figure~\ref{subfig:tsl} for an instance.
    Finally, TSP is formulated as 
    \begin{equation}
      \text{TSP}(R_g, R_t) = \frac{\exp(\nu\cdot a(R_g, R_t))}{1 + \exp(\nu\cdot a(R_g,
      R_t))}  \in [0, 1],
        \label{equ:def_tsl}
    \end{equation}
    where $\nu > 0$ is a preset parameter reflects the worst scenario we could
    assure the target is located correctly. In our experiment, $\nu$ is the solution of 
    \begin{equation}
     \frac{\exp(0.25 \nu)}{1 + \exp(0.25\nu)} = 0.95 \Longrightarrow \nu = 11.8.
      \label{equ:def_nu}
    \end{equation}
    In other words, when the overlapped region is larger than $25\%$ part of region $R^*$,
    we are convinced (with the probability of $0.95$) that the tracking is successful. 
    
    Obviously, the larger the TSP is, the more confident we believe this tracking is
    successful. If we apply TSP to the tracking results shown in Figure~\ref{subfig:dis},
    then the TSP of blue rectangle is $0.95$ which is significantly larger than that of
    the gray one (with TSP of $0.55$). The difference implies that TSP is capable to
    accommodate dynamic factors besides displacement. Another merit of TSP is the
    comparability over different video sequences thanks to its fixed value range \ie, $[0,
    1]$.  Considering these advantages, in the current paper, all the empirical results
    are evaluated by TSP. As a reference, tracking error results are also available.

    \subsection{Tracking Accuracy}
    \label{subsec:accuracy}

    Firstly, we examine the tracking accuracy of our trackers comparing with the
    competitors. The average TSP for every experiment is shown in
    Table~\ref{tab:accuracy}. For each video sequence, the optimal accuracy is displayed
    in bold type.

    \begin{table*}[h]
    \centering
    \caption
    {
      TSP values for tracking experiments. The term ``R-D$x$-Rand'' stands for RTCST with
      $x$-dimension features which is generated by random projection while the row
      started with ``RB-\ldots'' refers to the results with RTCSTB. ``PN$x$'' indicates $x$
      particles are used in the tracker. The optimal values for each video sequence is
      illustrated in bold type.
    }
    \resizebox{1.000\textwidth}{!}
    {
    \begin{tabular}{ c | c | l | l | l | l | l | l | l | l | l | l }

    \hline\hline
\multicolumn{2}{c|}{}   & \bf cubicle    & \bf dp    & \bf car4    & \bf car11    & \bf
fish    & \bf pets2000\_c1    & \bf pets2001\_c1    & \bf pets2002\_p1    & \bf
pets2004\_p1    & \bf pets2004-2\_p1   \\
\cline{1-12}
\multicolumn{2}{c|}{\bf KMS}  & $77\pm32.7$  & $\bf 100\pm0.4$  & $24\pm29.5$  & $67\pm40.5$ & $\bf 98\pm1.4$  & $94\pm4.8$  & $23\pm14.6$  & $23\pm37.0$  & $52\pm30.2$  & $26\pm32.8$ \\
\cline{1-12}
{\bf \multirow{3}{*}{PF}}  
& PN100  & $95\pm8.1$  & $98\pm3.2$  & $64\pm30.0$  & $37\pm30.6$  & $90\pm14.2$  & $45\pm31.5$  & $97\pm2.7$  & $24\pm38.2$  & $23\pm25.8$  & $58\pm16.9$  \\
& PN200  & $95\pm8.3$  & $98\pm3.0$  & $65\pm30.3$  & $39\pm31.8$  & $90\pm14.7$  & $44\pm31.7$  & $98\pm2.5$  & $24\pm38.4$  & $23\pm25.9$  & $58\pm16.9$  \\
& PN500  & $95\pm7.9$  & $98\pm2.9$  & $64\pm33.6$  & $39\pm32.7$  & $90\pm15.4$  & $44\pm33.1$  & $98\pm2.5$  & $24\pm38.4$  & $22\pm25.8$  & $58\pm17.0$ \\
\cline{1-12}
{\bf \multirow{2}{*}{R-D25-Rand}}  
& PN100  & $69\pm21.4$  & $66\pm20.1$  & $89\pm11.0$  & $64\pm17.2$  & $63\pm20.1$  & $77\pm8.9$  & $89\pm8.3$  & $54\pm21.6$  & $31\pm25.8$  & $33\pm29.1$  \\
& PN200  & $80\pm15.8$  & $78\pm15.7$  & $95\pm7.5$  & $62\pm20.7$  & $64\pm20.2$  & $80\pm8.5$  & $87\pm10.1$  & $63\pm16.0$  & $28\pm25.6$  & $29\pm31.1$  \\
\cline{1-12}
{\bf \multirow{2}{*}{R-D50-Rand}}  
& PN100  & $73\pm21.5$  & $78\pm16.7$  & $95\pm8.1$  & $64\pm24.1$  & $61\pm21.0$  & $72\pm10.1$  & $86\pm12.5$  & $65\pm16.1$  & $28\pm25.3$  & $25\pm33.1$  \\
& PN200  & $69\pm23.0$  & $82\pm17.9$  & $95\pm10.7$  & $81\pm22.3$  & $64\pm19.0$  & $81\pm9.1$  & $83\pm13.4$  & $64\pm15.1$  & $31\pm25.2$  & $25\pm32.9$  \\
\cline{1-12}
{\bf \multirow{2}{*}{R-D100-Rand}}  
& PN100  & $70\pm24.7$  & $71\pm21.5$  & $94\pm11.2$  & $\bf 85\pm24.6$  & $64\pm19.3$  & $72\pm12.5$  & $93\pm5.1$  & $61\pm16.1$  & $28\pm27.2$  & $26\pm32.0$  \\
& PN200  & $72\pm22.3$  & $77\pm17.6$  & $96\pm8.8$  & $78\pm23.1$  & $59\pm20.6$  & $81\pm8.7$  & $91\pm6.9$  & $68\pm13.6$  & $32\pm25.7$  & $24\pm33.2$  \\
\cline{1-12}
{\bf \multirow{2}{*}{R-D25-Hash}}  
& PN100  & $73\pm21.3$  & $76\pm12.0$  & $90\pm12.1$  & $65\pm24.4$  & $64\pm19.9$  & $83\pm6.4$  & $77\pm20.3$  & $67\pm15.0$  & $38\pm25.0$  & $32\pm29.6$  \\
& PN200  & $77\pm18.3$  & $81\pm14.6$  & $89\pm14.8$  & $59\pm23.9$  & $63\pm20.2$  & $96\pm2.7$  & $70\pm23.5$  & $55\pm19.6$  & $35\pm25.8$  & $33\pm29.8$  \\
\cline{1-12}
{\bf \multirow{2}{*}{R-D50-Hash}}  
& PN100  & $73\pm21.8$  & $79\pm16.5$  & $98\pm3.2$  & $75\pm24.1$  & $66\pm21.3$  & $73\pm11.8$  & $\bf 100\pm0.1$  & $64\pm16.0$  & $34\pm25.3$  & $22\pm32.3$  \\
& PN200  & $75\pm21.7$  & $83\pm14.2$  & $\bf 99\pm1.2$  & $74\pm22.5$  & $68\pm21.6$  & $79\pm10.5$  & $\bf 100\pm0.1$  & $63\pm15.7$  & $39\pm24.4$  & $21\pm33.5$  \\
\cline{1-12}
{\bf \multirow{2}{*}{R-D100-Hash}}  
& PN100  & $82\pm15.0$  & $88\pm10.1$  & $95\pm9.1$  & $80\pm32.9$  & $56\pm22.0$  & $91\pm4.6$  & $\bf 100\pm0.1$  & $64\pm14.5$  & $30\pm25.9$  & $27\pm31.6$  \\
& PN200  & $90\pm8.3$  & $92\pm8.5$  & $95\pm9.3$  & $80\pm33.6$  & $52\pm23.8$  & $92\pm5.3$  & $\bf 100\pm0.1$  & $67\pm13.3$  & $30\pm26.3$  & $28\pm31.8$  \\
\cline{1-12}
{\bf \multirow{2}{*}{RB-D25-Rand}}  
& PN100  & $-$  & $-$  & $-$  & $-$  & $-$  & $76\pm7.0$  & $86\pm9.5$  & $80\pm8.8$  & $\bf 68\pm18.2$  & $49\pm23.2$  \\
& PN200  & $-$  & $-$  & $-$  & $-$  & $-$  & $92\pm3.4$  & $84\pm12.1$  & $78\pm10.2$  & $62\pm17.4$  & $59\pm19.2$  \\
\cline{1-12}
{\bf \multirow{2}{*}{RB-D50-Rand}}  
& PN100  & $-$  & $-$  & $-$  & $-$  & $-$  & $86\pm5.8$  & $98\pm2.0$  & $73\pm11.8$  & $58\pm18.4$  & $44\pm26.5$  \\
& PN200  & $-$  & $-$  & $-$  & $-$  & $-$  & $93\pm3.6$  & $97\pm2.7$  & $77\pm10.8$  & $58\pm18.3$  & $62\pm17.8$  \\
\cline{1-12}
{\bf \multirow{2}{*}{RB-D100-Rand}}  
& PN100  & $-$  & $-$  & $-$  & $-$  & $-$  & $96\pm4.2$  & $\bf 100\pm0.6$  & $74\pm11.6$  & $46\pm24.0$  & $54\pm20.8$  \\
& PN200  & $-$  & $-$  & $-$  & $-$  & $-$  & $95\pm5.0$  & $\bf 100\pm0.1$  & $72\pm11.8$  & $51\pm21.7$  & $53\pm22.0$  \\
\cline{1-12}
{\bf \multirow{2}{*}{RB-D25-Hash}}  
& PN100  & $-$  & $-$  & $-$  & $-$  & $-$  & $89\pm2.9$  & $94\pm6.1$  & $79\pm10.3$  & $64\pm20.7$  & $71\pm14.4$  \\
& PN200  & $-$  & $-$  & $-$  & $-$  & $-$  & $89\pm3.6$  & $89\pm8.9$  & $77\pm10.5$  & $61\pm16.1$  & $\bf 77\pm12.0$  \\
\cline{1-12}
{\bf \multirow{2}{*}{RB-D50-Hash}}  
& PN100  & $-$  & $-$  & $-$  & $-$  & $-$  & $75\pm12.0$  & $98\pm1.7$  & $\bf 82\pm9.0$  & $42\pm25.3$  & $52\pm22.7$ \\
& PN200  & $-$  & $-$  & $-$  & $-$  & $-$  & $98\pm1.9$  & $98\pm1.7$  & $\bf 82\pm8.7$  & $59\pm19.5$  & $71\pm14.0$ \\
\cline{1-12}
{\bf \multirow{2}{*}{RB-D100-Hash}}  
& PN100  & $-$  & $-$  & $-$  & $-$  & $-$  & $97\pm1.9$  & $99\pm1.3$  & $\bf 82\pm8.9$  & $51\pm20.8$  & $67\pm14.7$  \\
& PN200  & $-$  & $-$  & $-$  & $-$  & $-$  & $\bf 99\pm1.4$  & $98\pm1.7$  & $\bf82\pm9.8$  & $53\pm22.2$  & $71\pm13.1$ \\
\cline{1-12}
\multicolumn{2}{c|}{\bf L1T}  & $\bf 99\pm2.2$  & $92\pm8.8$  & $-$  & $77\pm37.4$  & $-$  & $-$  & $\bf 100\pm0.0$  & $-$  & $34\pm26.4$  & $-$ \\
    \hline\hline
    \end{tabular}
    }
    \label{tab:accuracy}
    \end{table*}

    As illustrated in Table \ref{tab:accuracy}, all the tracking approaches achieve
    similar performances on the sequence with simple background and stable illumination
    (\emph{dp} and \emph{cubicle}). For the video sequence \emph{fish}, traditional
    methods show higher capacity for accommodating extreme illumination variation. On the other
    hand, for the outdoor scene and complex background tasks, \ie, the other $7$
    sequences, CS-based trackers consistently outperform PF tracker and KMS tracker.
    All the best performances are observed with RTCST and RTCST-B for these video
    sequences. Considering that the target could be viewed as missed when the TSP is below
    $30\%$, the traditional trackers are failure for the majority of these video datasets,
    \ie, KMS tracker for \emph{car4}, \emph{pets2001\_c1}, \emph{pets2002\_p1} and
    \emph{pets2004-2\_p1}; PF tracker for \emph{pets2002\_p1} and
    \emph{pets2004\_p1}. Moreover, $\ell_1$ tracker also fails on \emph{pets2004\_p1} and
    \emph{pets2004-2\_p1} due to the unstable target appearances. Our methods, on the
    contrary, do much better than the competitors and handle some intractable sequences
    (\eg,  \emph{pets2004\_p1} and \emph{pets2004-2\_p1}) very smoothly (with the TSP $>
    65\%$). Particularly, for the camera-fixed scenes, RTCST-B is applied and always
    achieves the highest accuracy. The superiority of RTCST-B over all the other trackers
    confirms our assumption that higher accuracy would be achieved when the tracking is
    considered as binary classification problem. 
    
\begin{figure*}[h]
      \centering
      \renewcommand{\thesubfigure}{(\arabic{subfigure})} 
      \subfigure[\#1]{\includegraphics[width=0.19\textwidth]{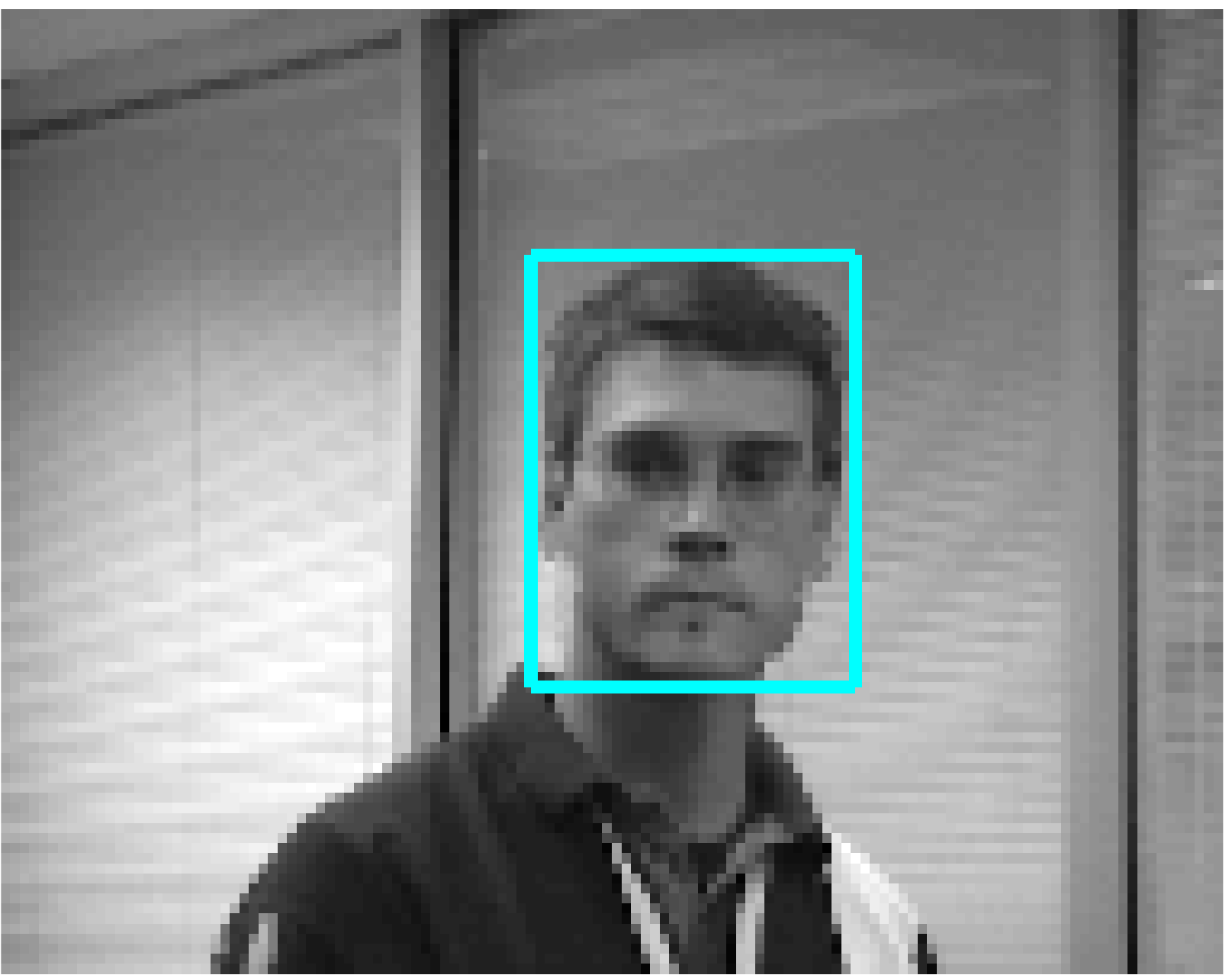}}
      \subfigure[\#16]{\includegraphics[width=0.19\textwidth]{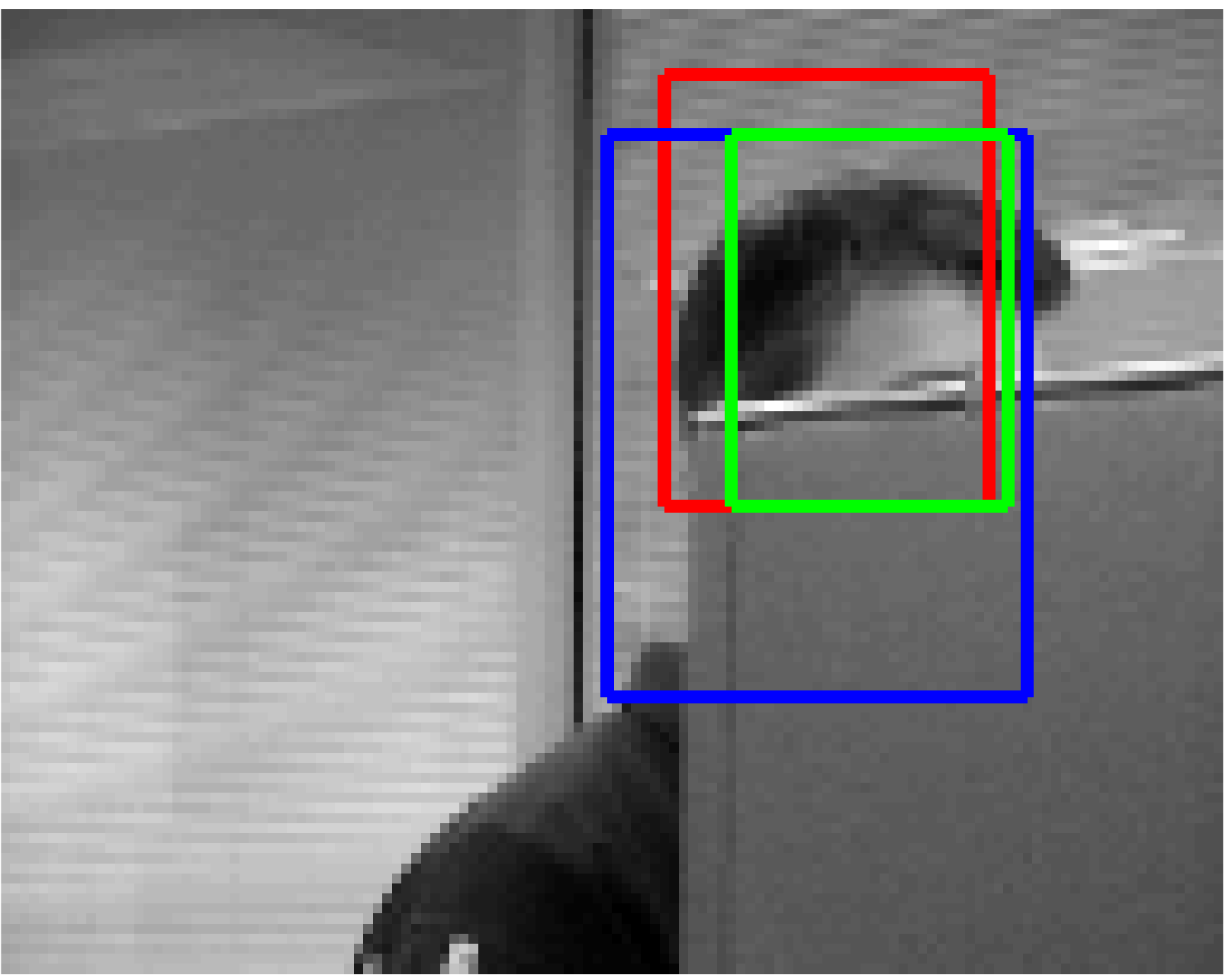}}
      \subfigure[\#31]{\includegraphics[width=0.19\textwidth]{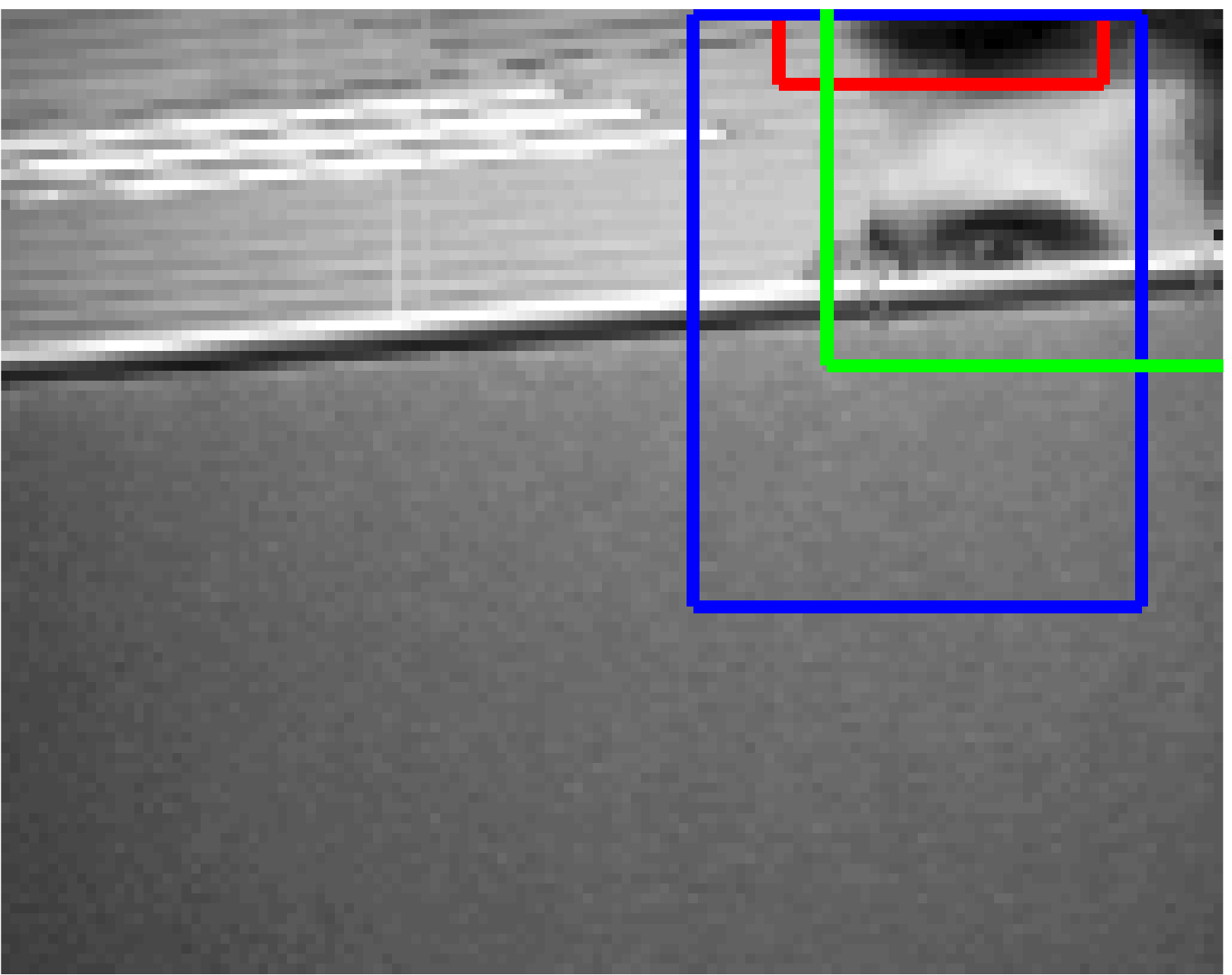}}
      \subfigure[\#41]{\includegraphics[width=0.19\textwidth]{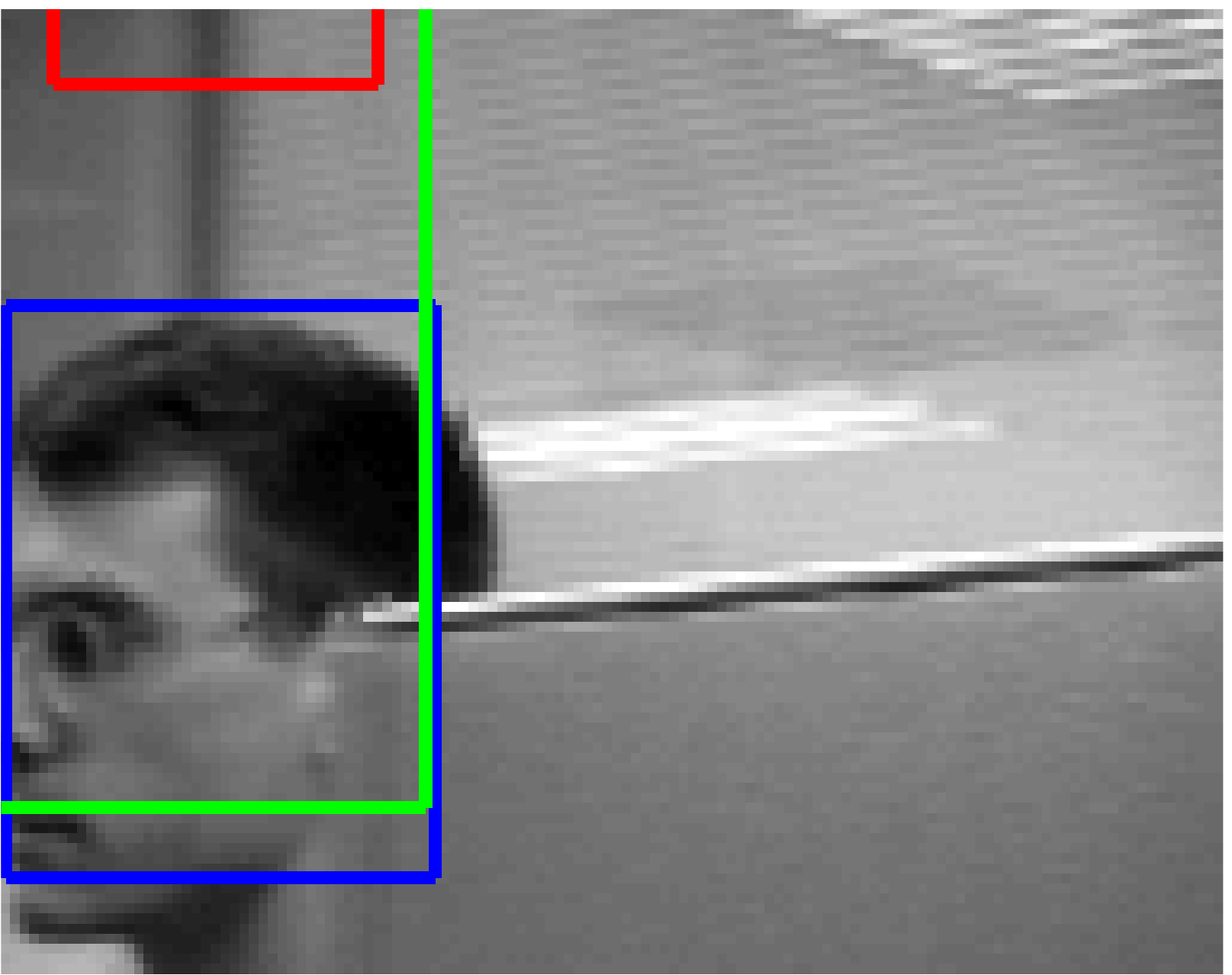}}
      \subfigure[\#51]{\includegraphics[width=0.19\textwidth]{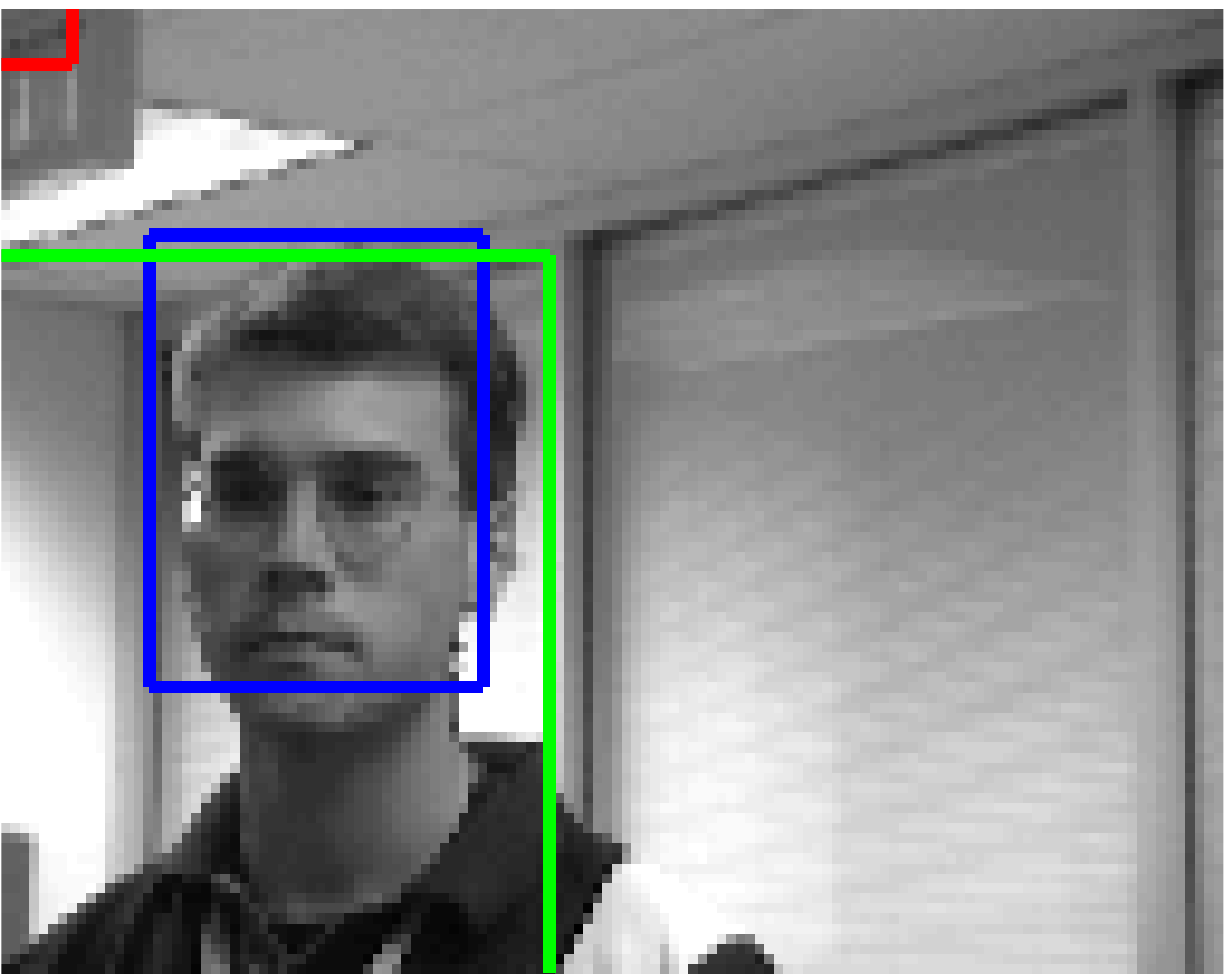}}
      \subfigure[\#1]{\includegraphics[width=0.19\textwidth]{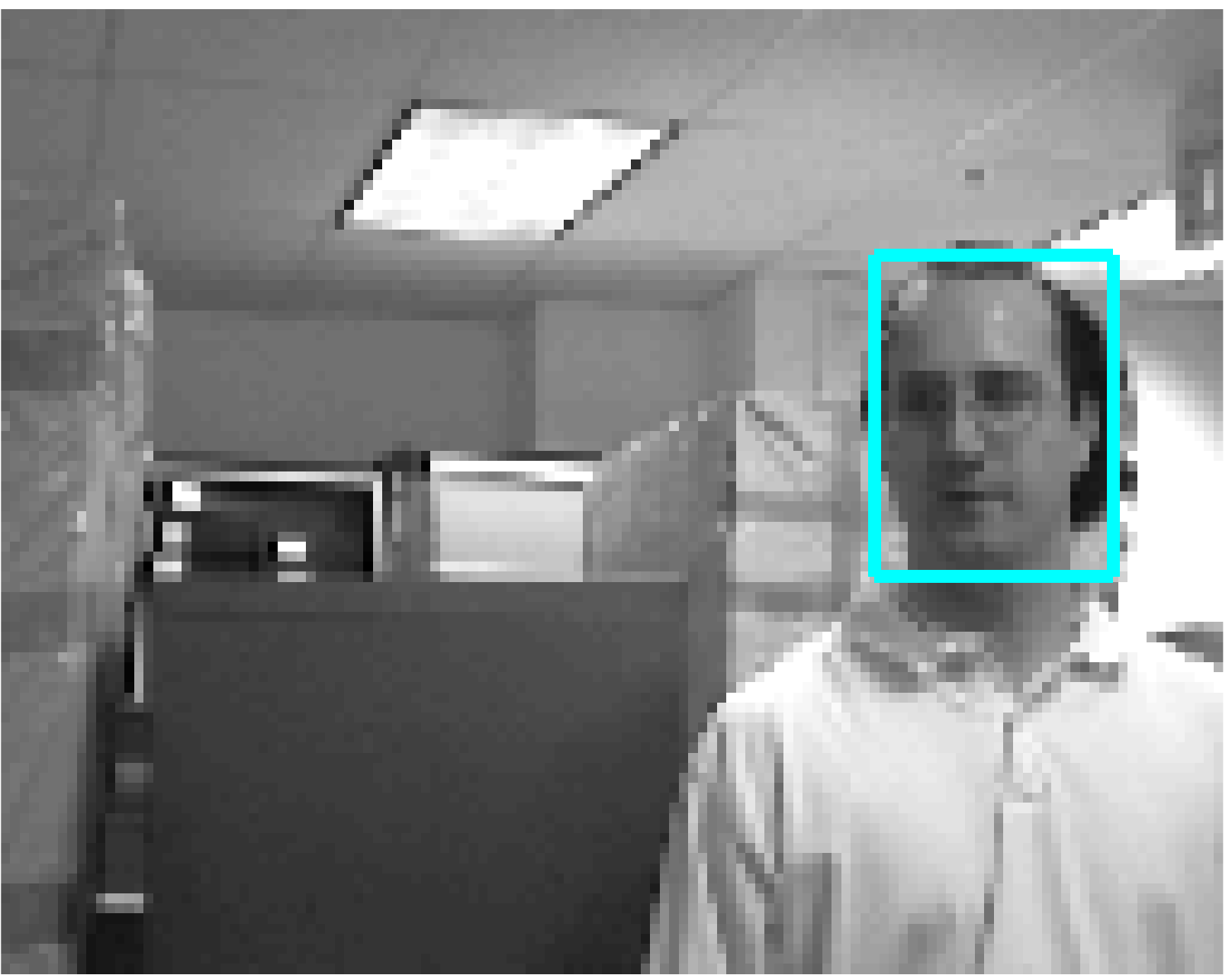}}
      \subfigure[\#21]{\includegraphics[width=0.19\textwidth]{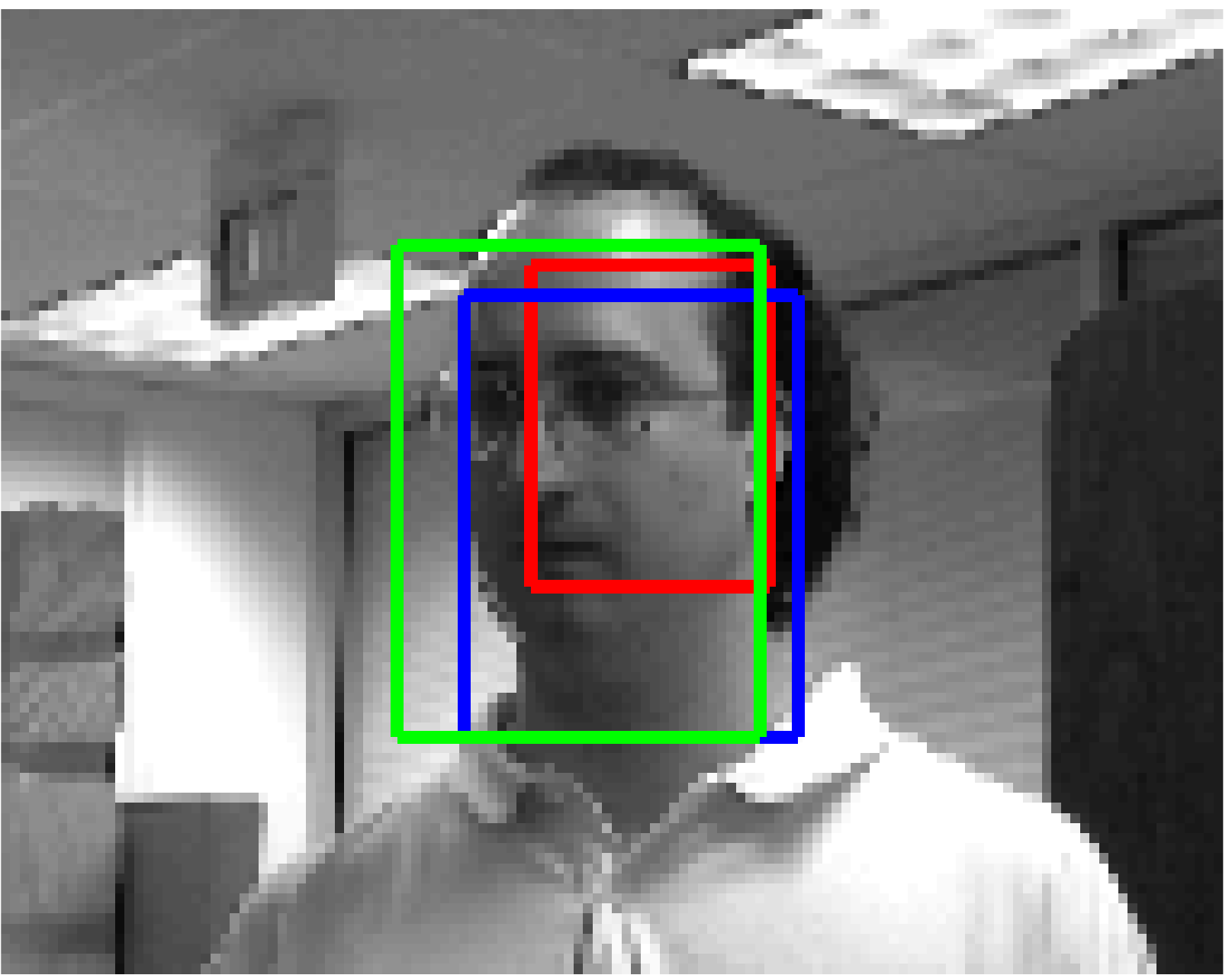}}
      \subfigure[\#36]{\includegraphics[width=0.19\textwidth]{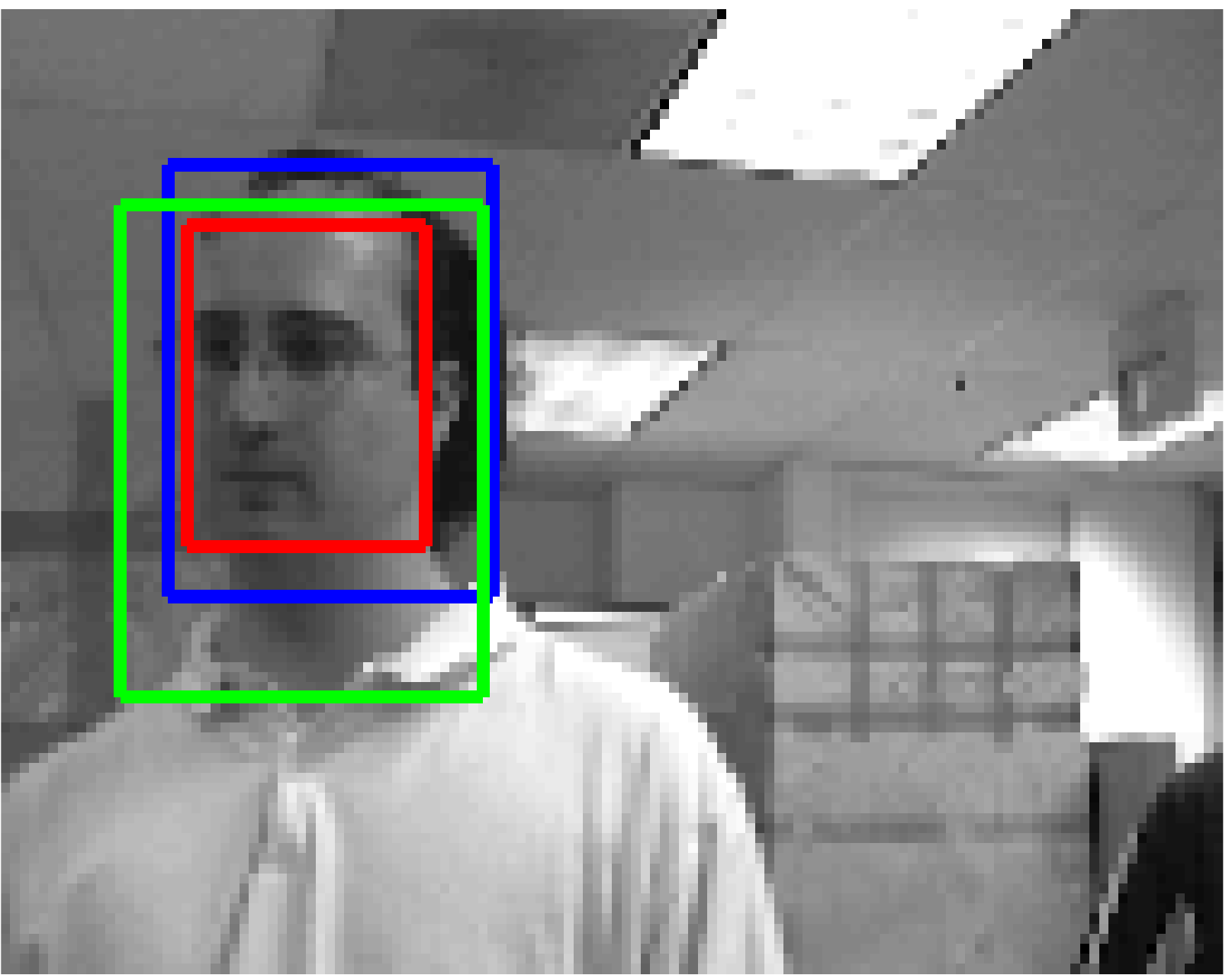}}
      \subfigure[\#46]{\includegraphics[width=0.19\textwidth]{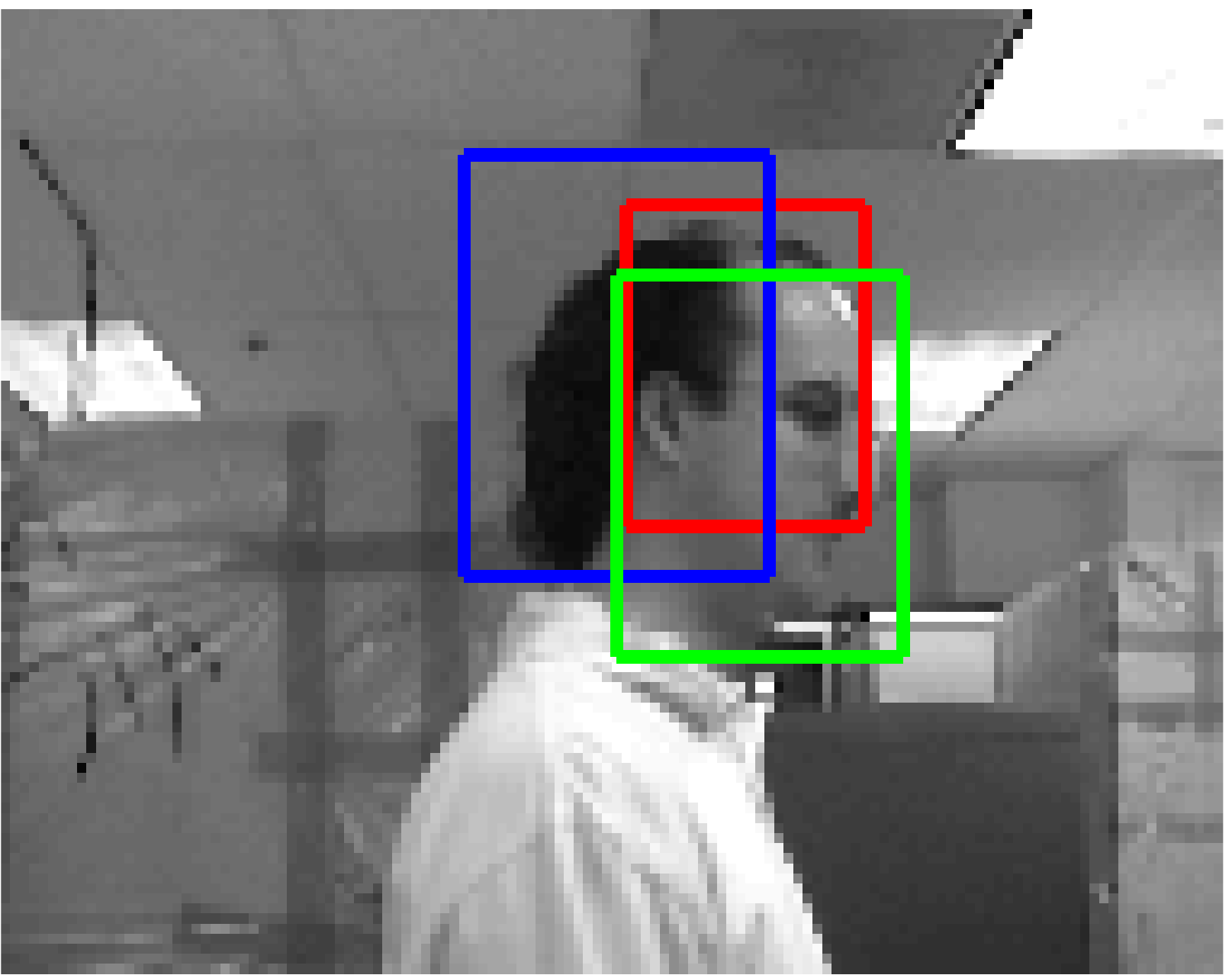}}
      \subfigure[\#66]{\includegraphics[width=0.19\textwidth]{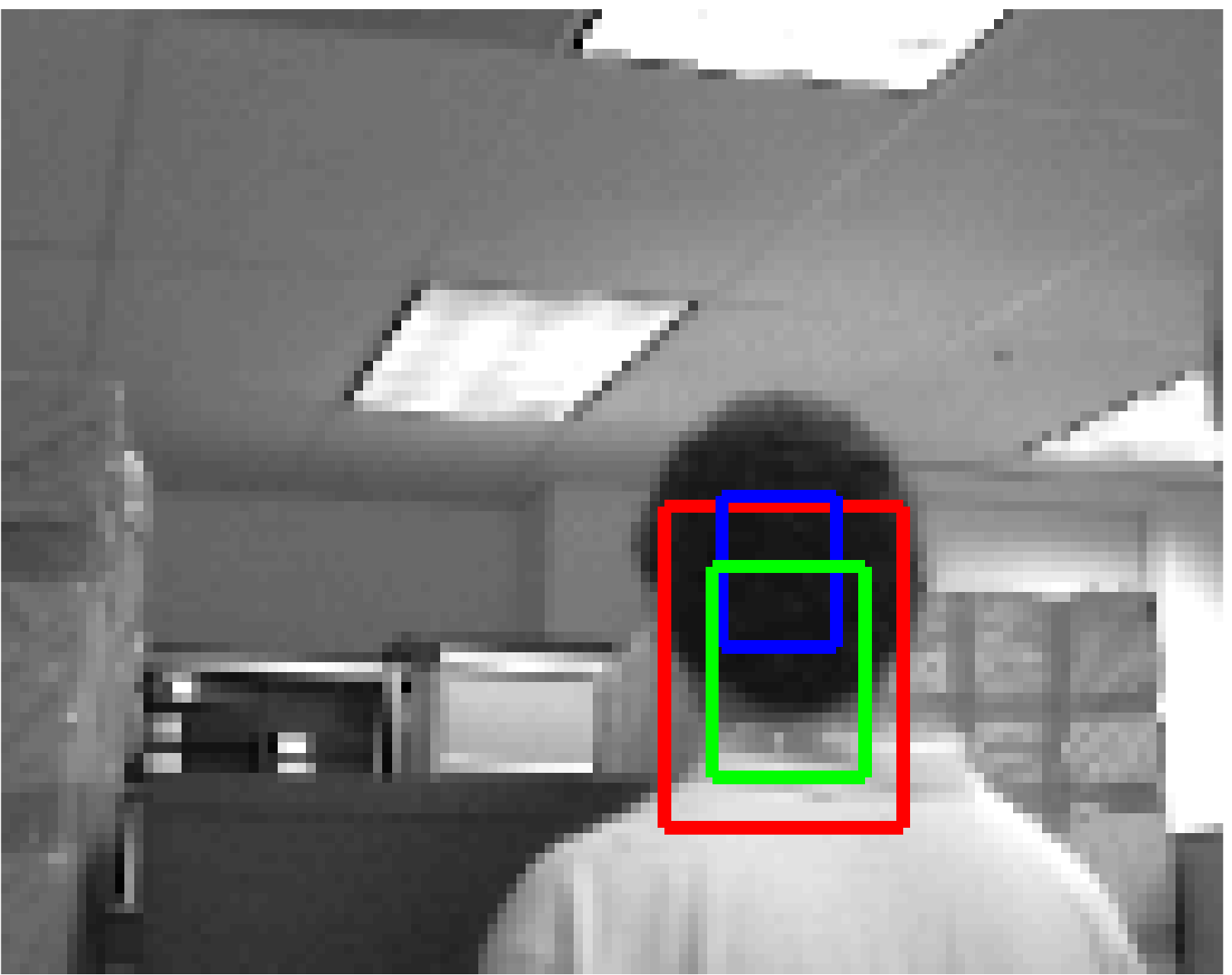}}
      \subfigure[\#1]{\includegraphics[width=0.19\textwidth]{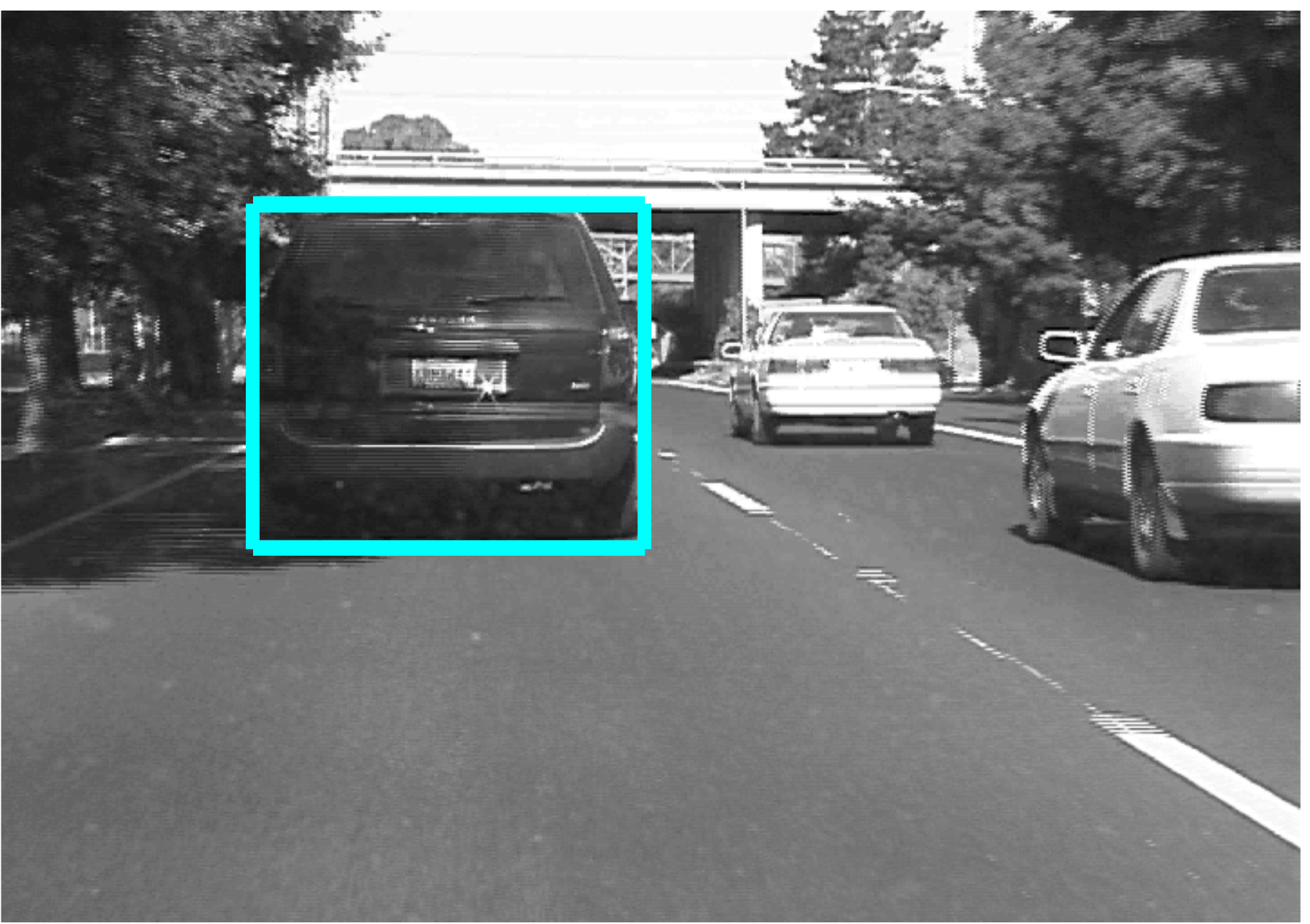}}
      \subfigure[\#21]{\includegraphics[width=0.19\textwidth]{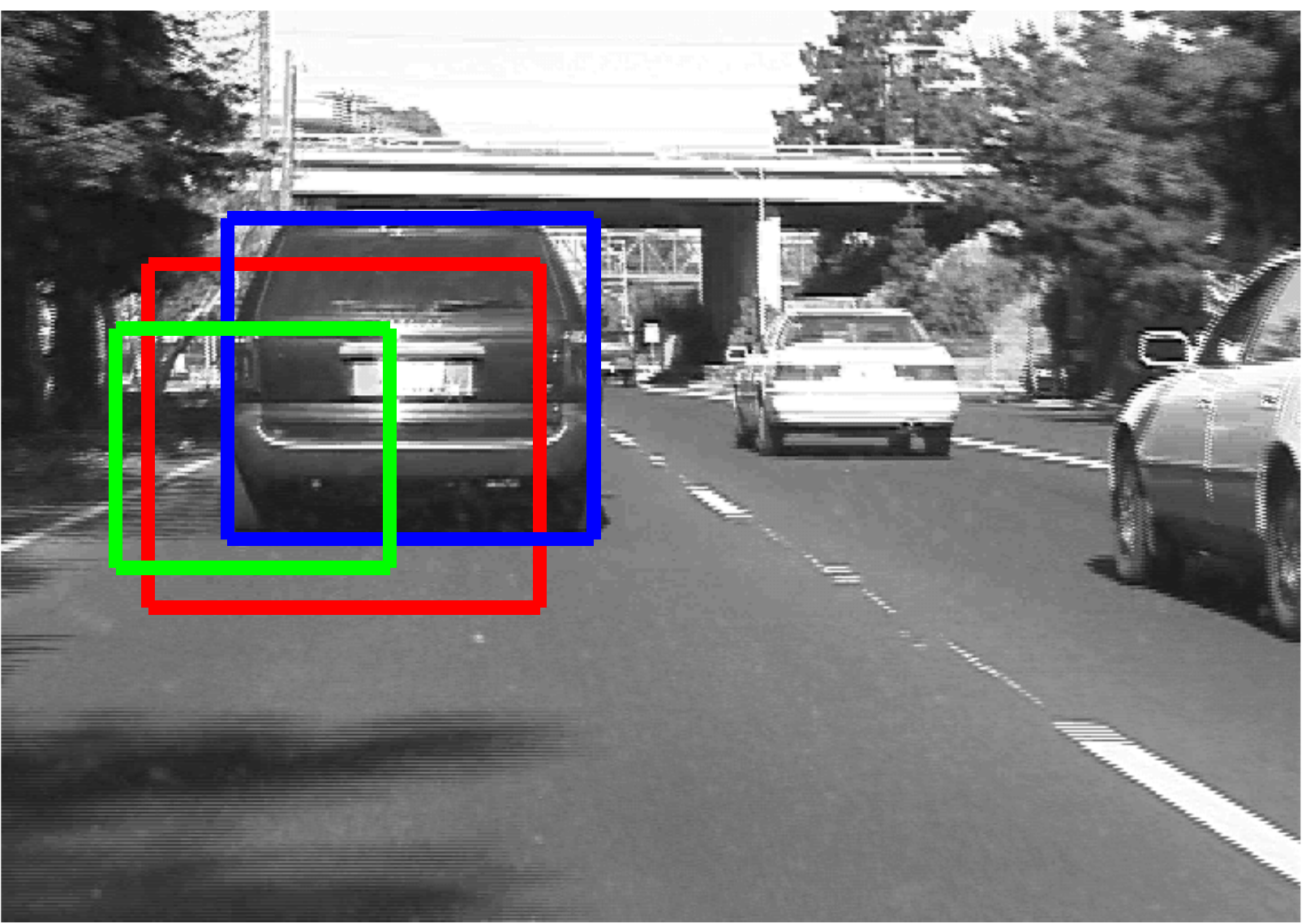}}
      \subfigure[\#51]{\includegraphics[width=0.19\textwidth]{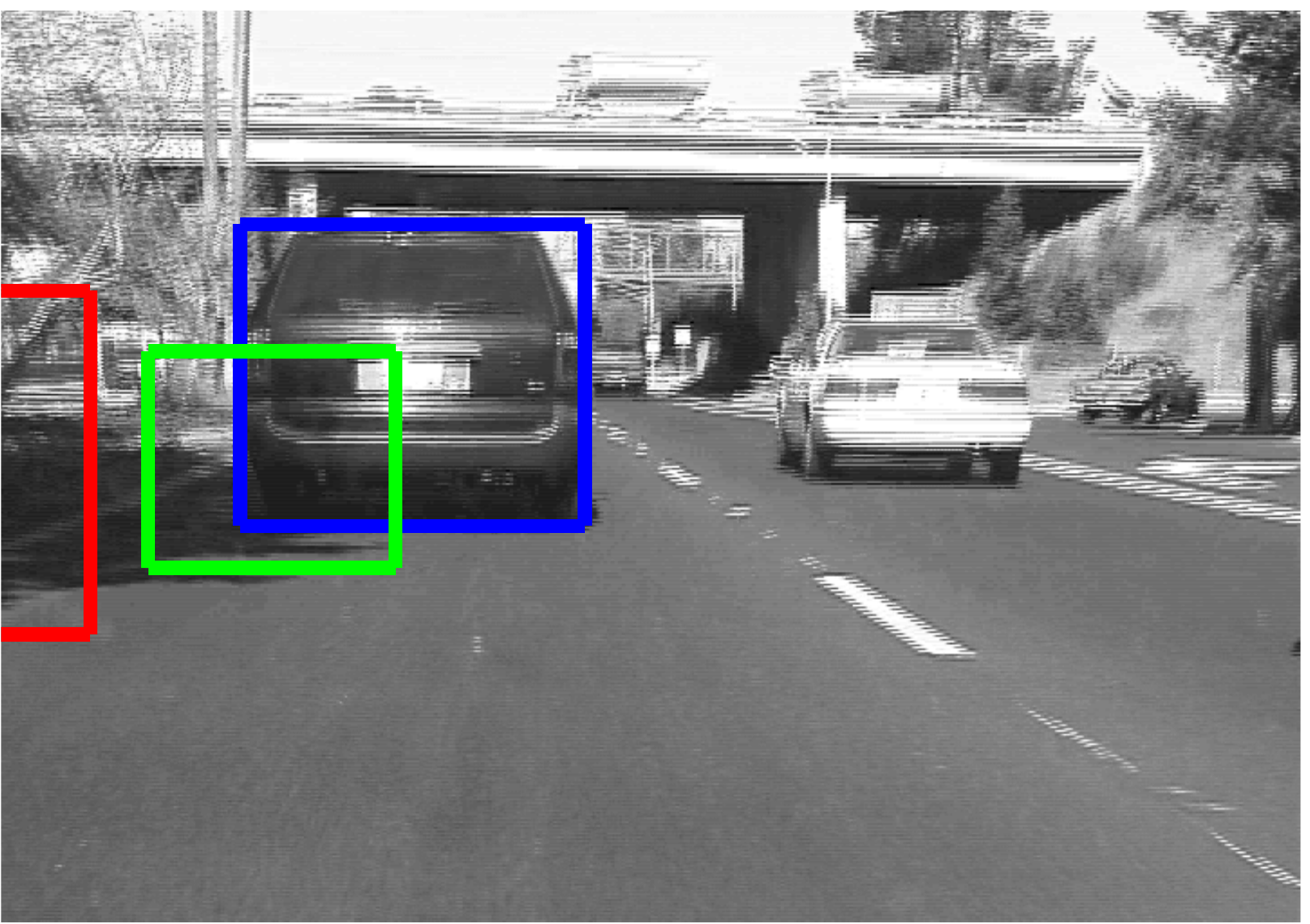}}
      \subfigure[\#101]{\includegraphics[width=0.19\textwidth]{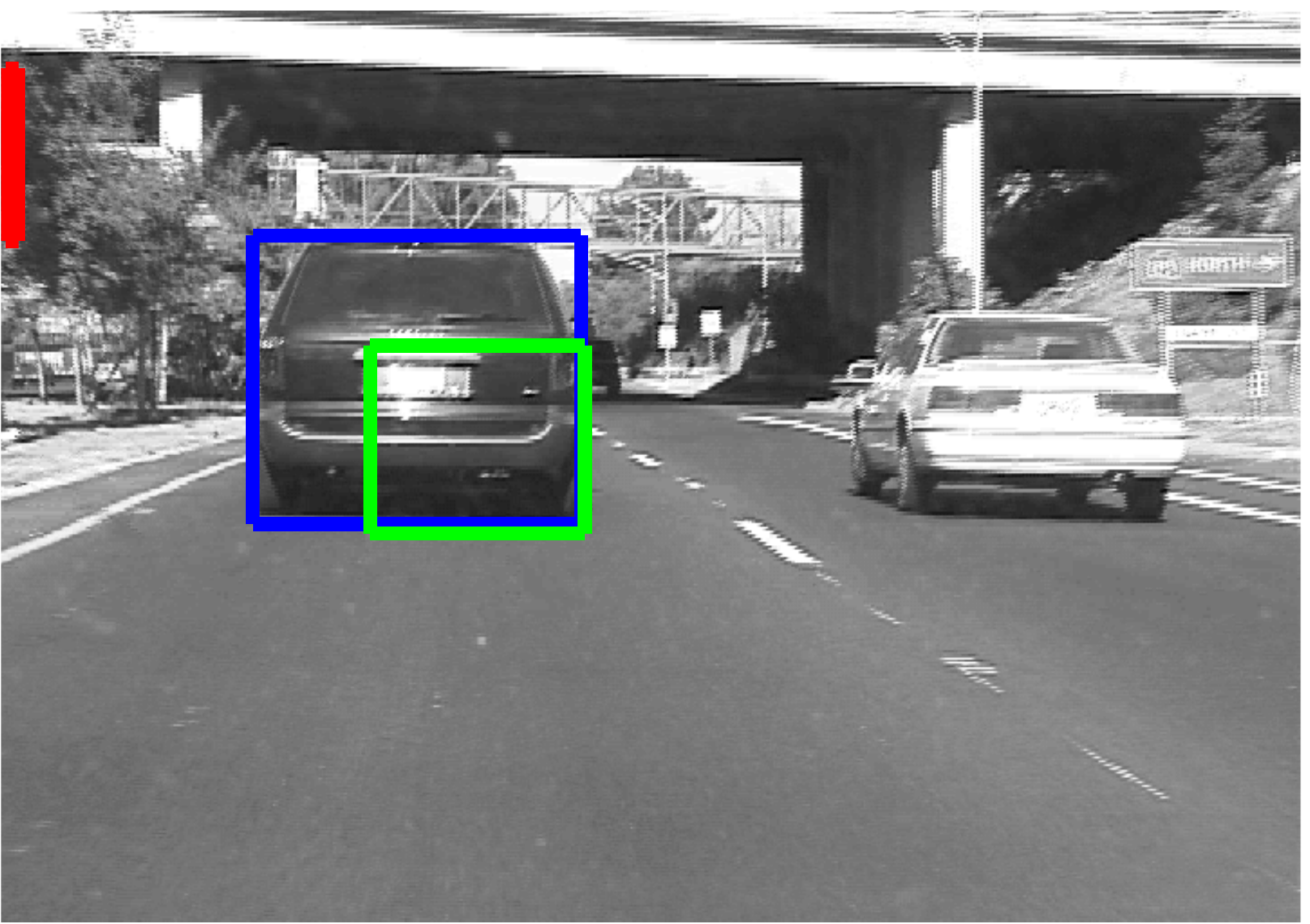}}
      \subfigure[\#161]{\includegraphics[width=0.19\textwidth]{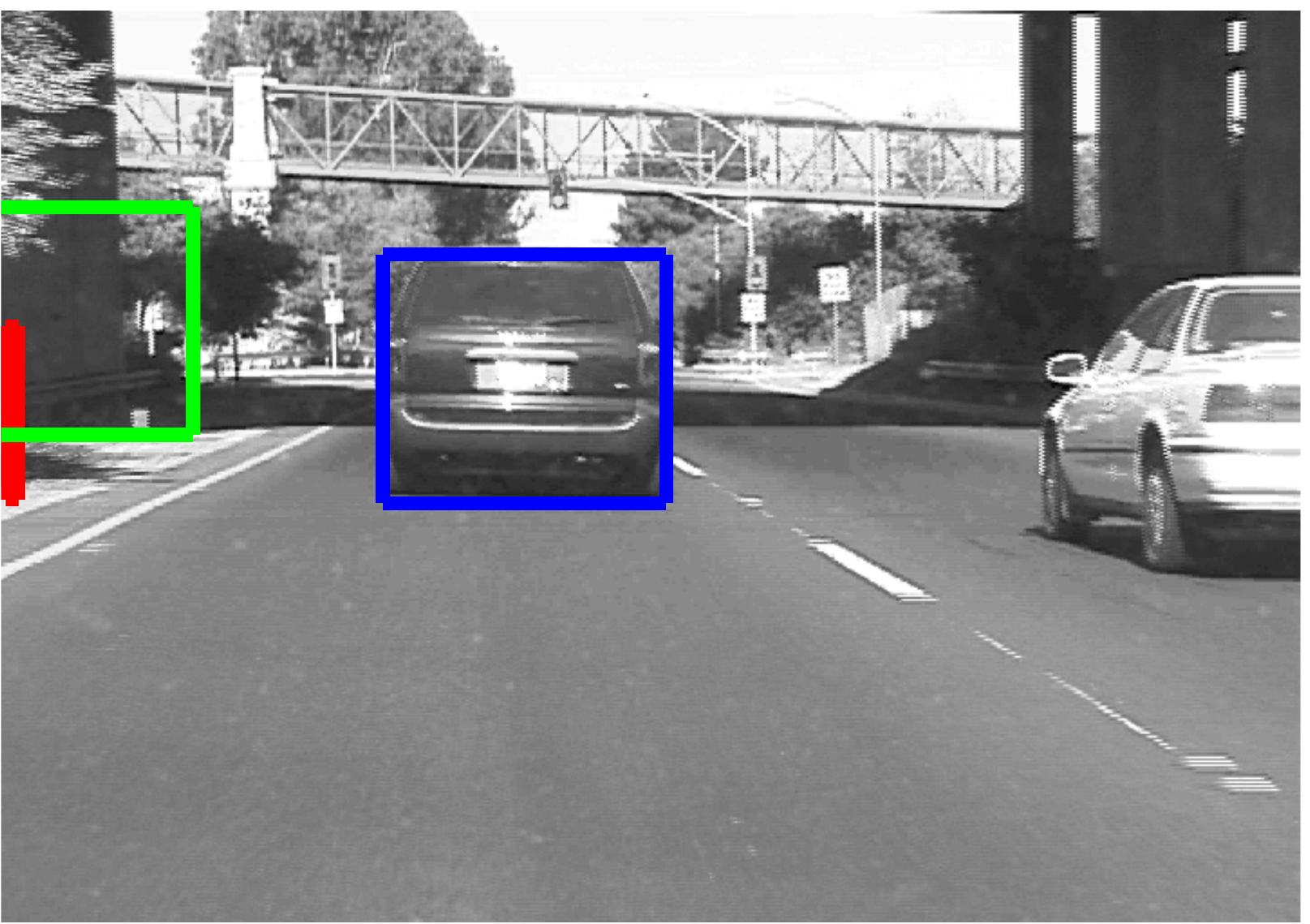}}
      \subfigure[\#122]{\includegraphics[width=0.19\textwidth]{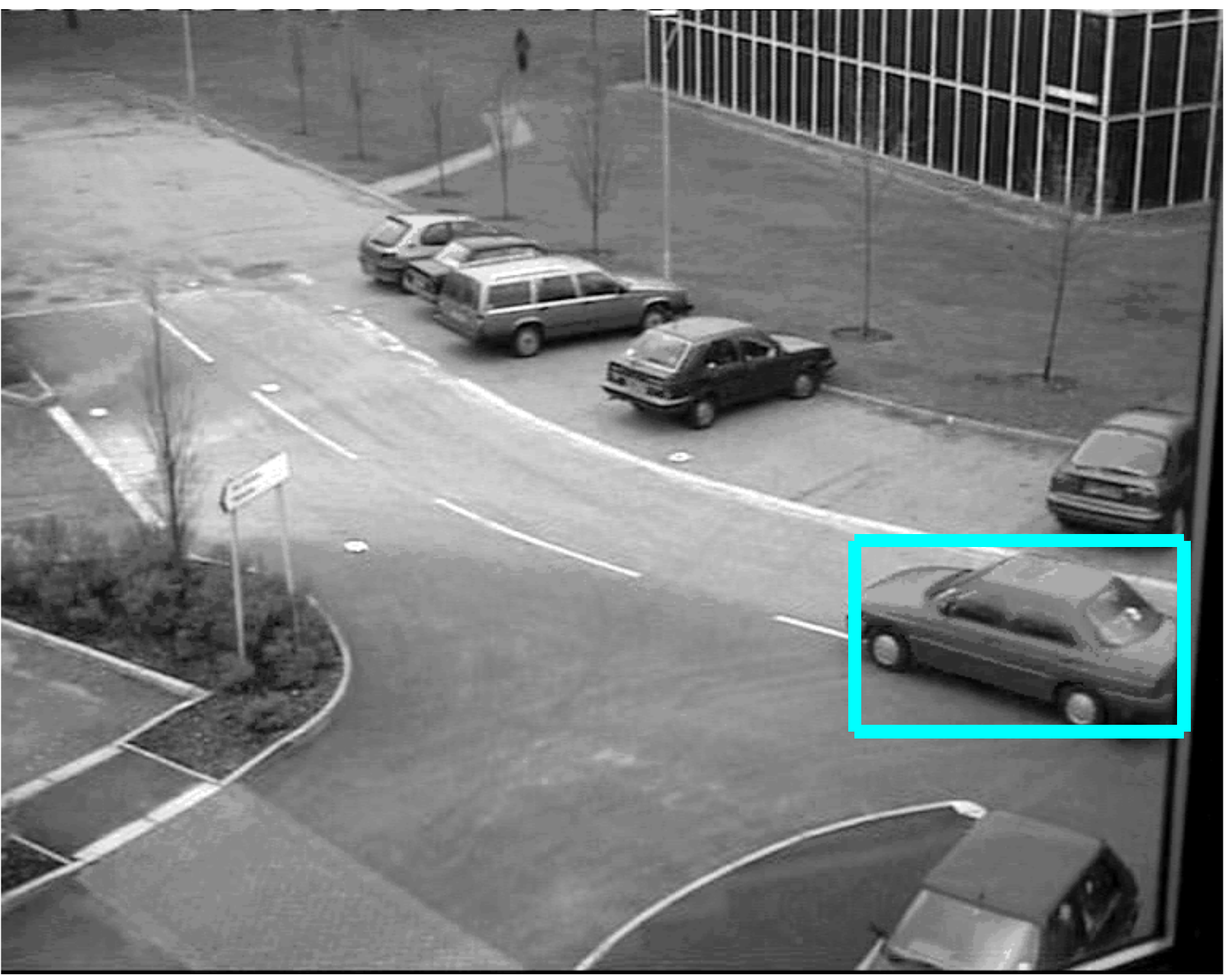}}
      \subfigure[\#162]{\includegraphics[width=0.19\textwidth]{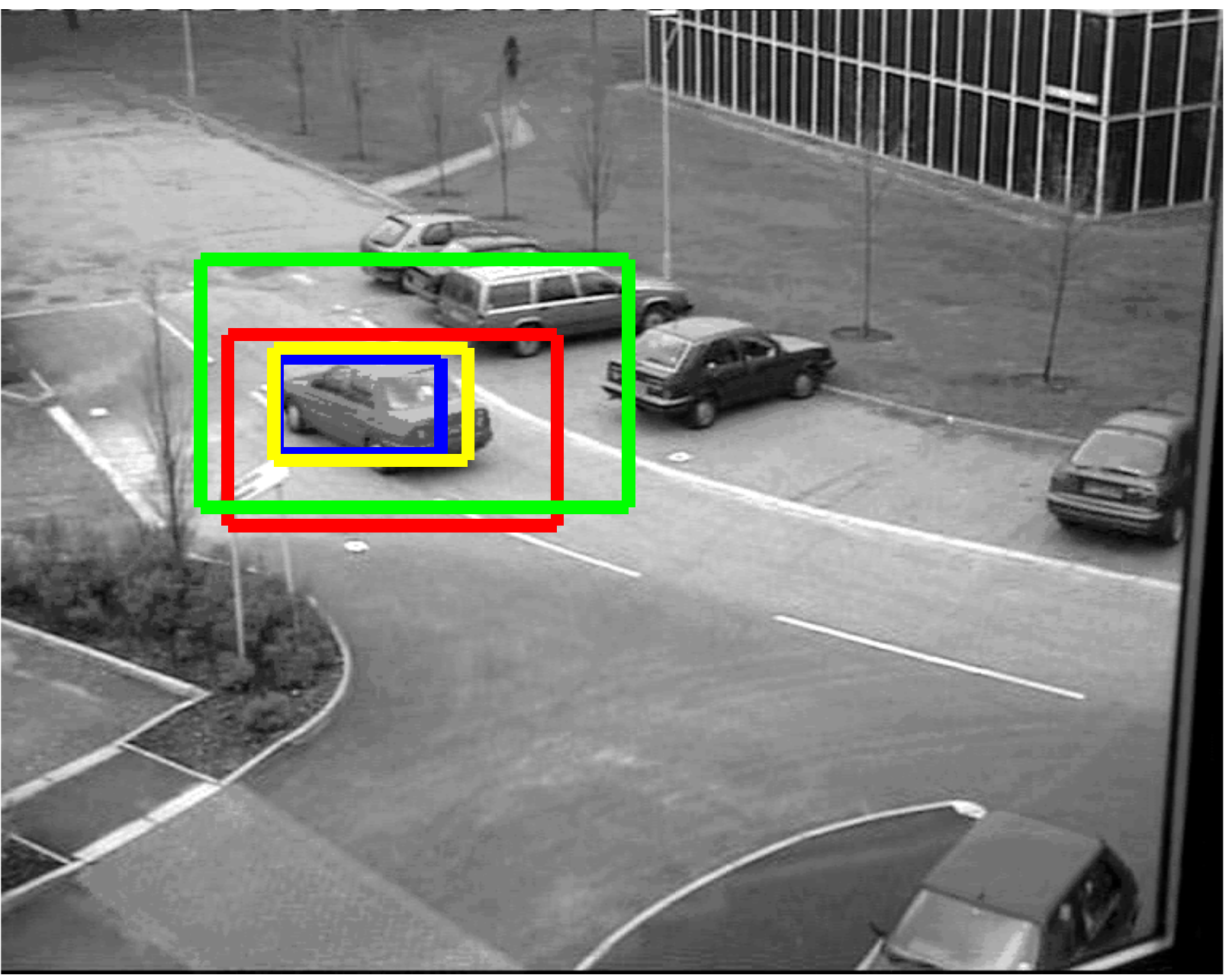}}
      \subfigure[\#192]{\includegraphics[width=0.19\textwidth]{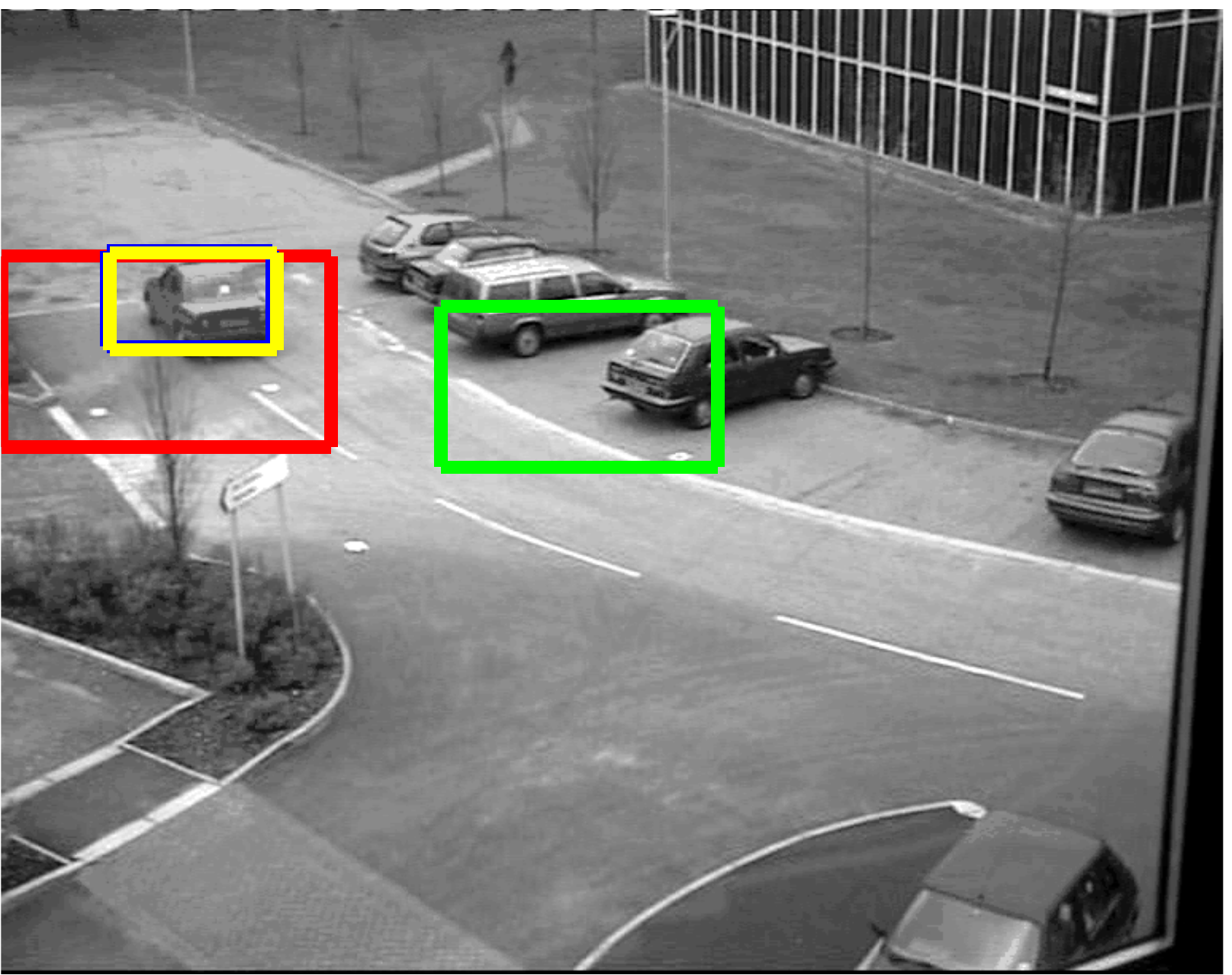}}
      \subfigure[\#222]{\includegraphics[width=0.19\textwidth]{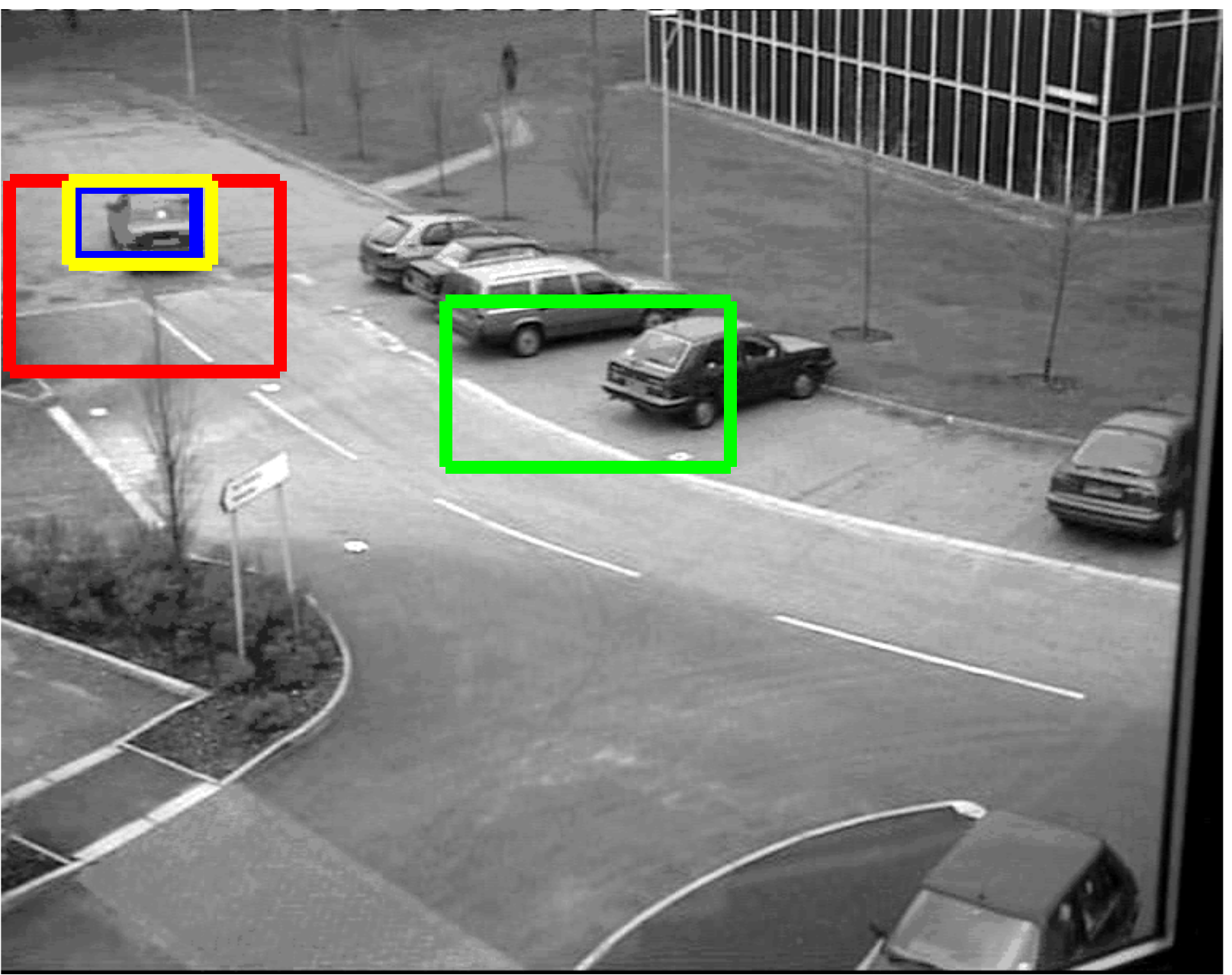}}
      \subfigure[\#302]{\includegraphics[width=0.19\textwidth]{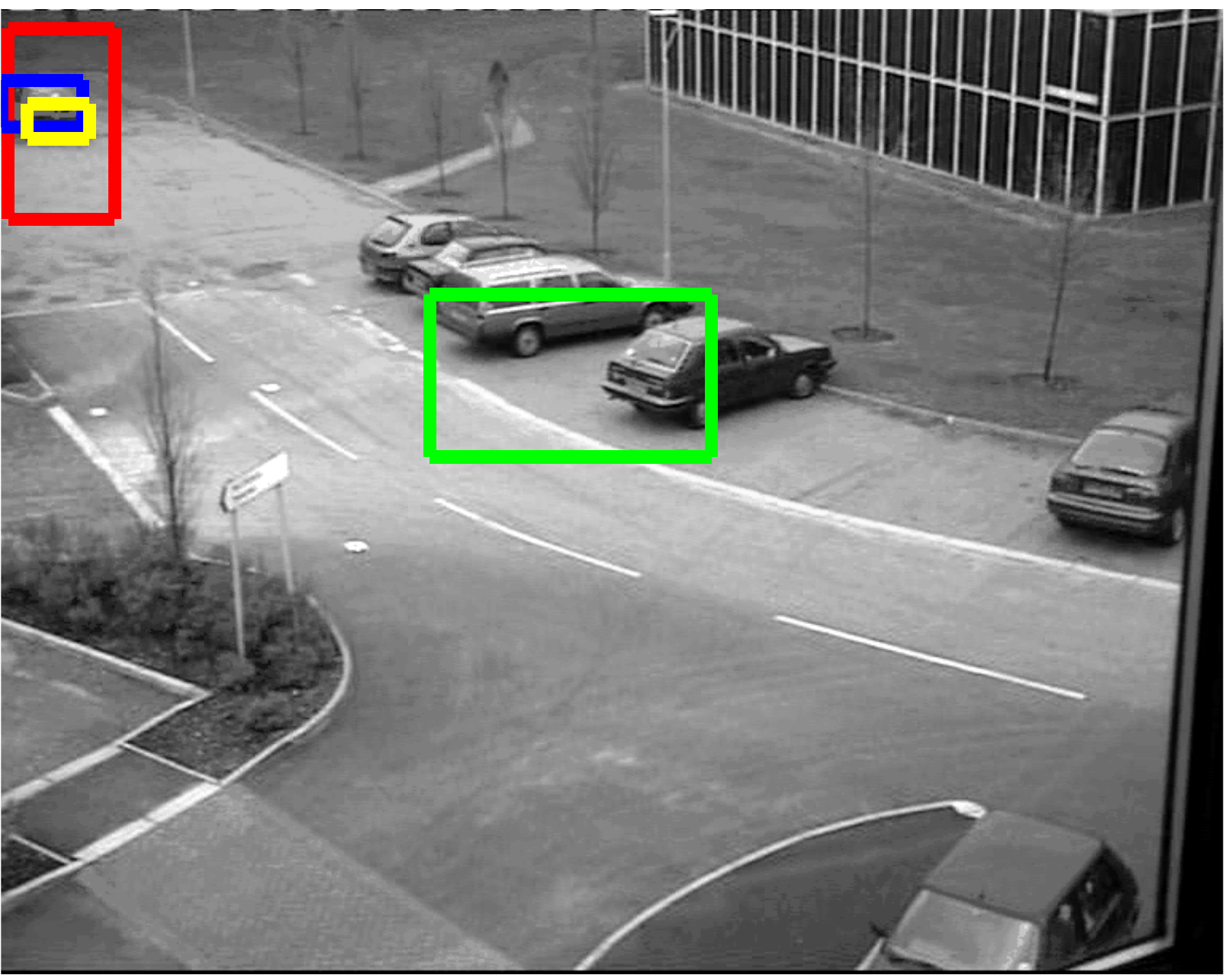}}
      \subfigure[\#275]{\includegraphics[width=0.19\textwidth, height=0.13\textwidth]{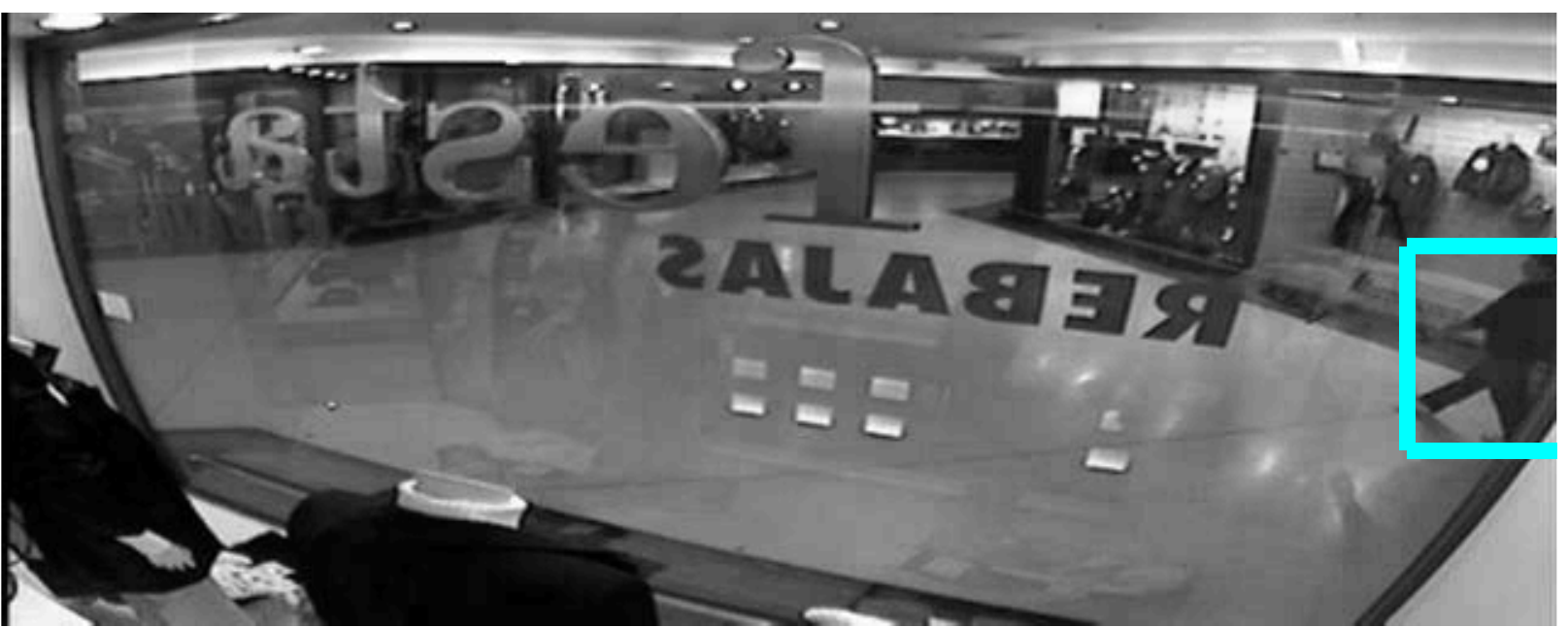}}
      \subfigure[\#305]{\includegraphics[width=0.19\textwidth, height=0.13\textwidth]{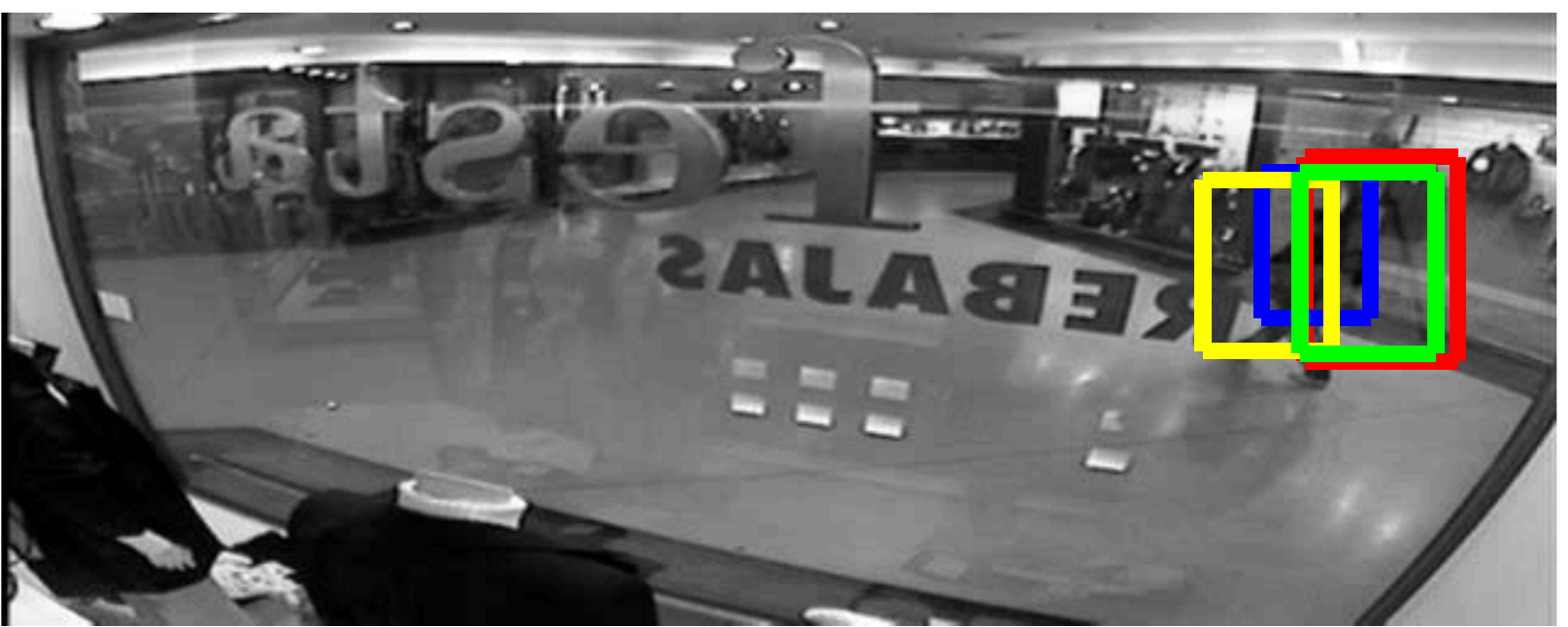}}
      \subfigure[\#335]{\includegraphics[width=0.19\textwidth, height=0.13\textwidth]{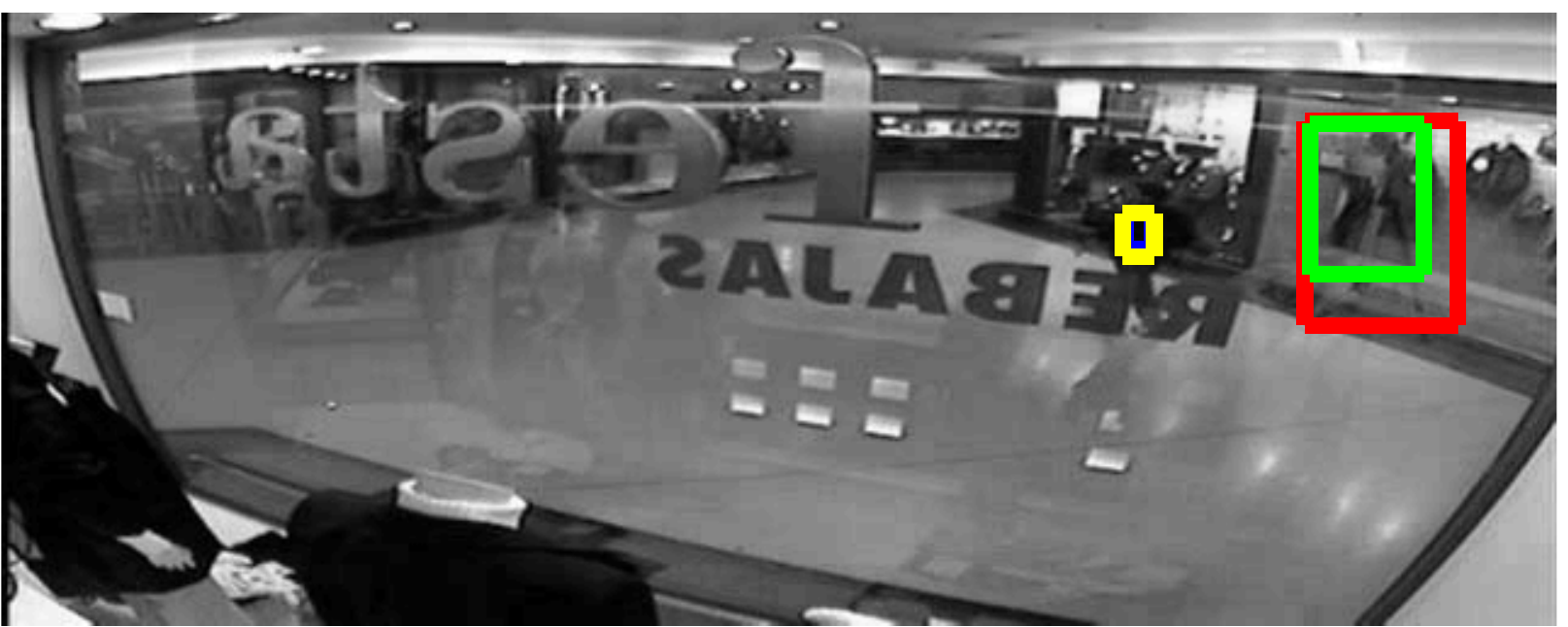}}
      \subfigure[\#365]{\includegraphics[width=0.19\textwidth, height=0.13\textwidth]{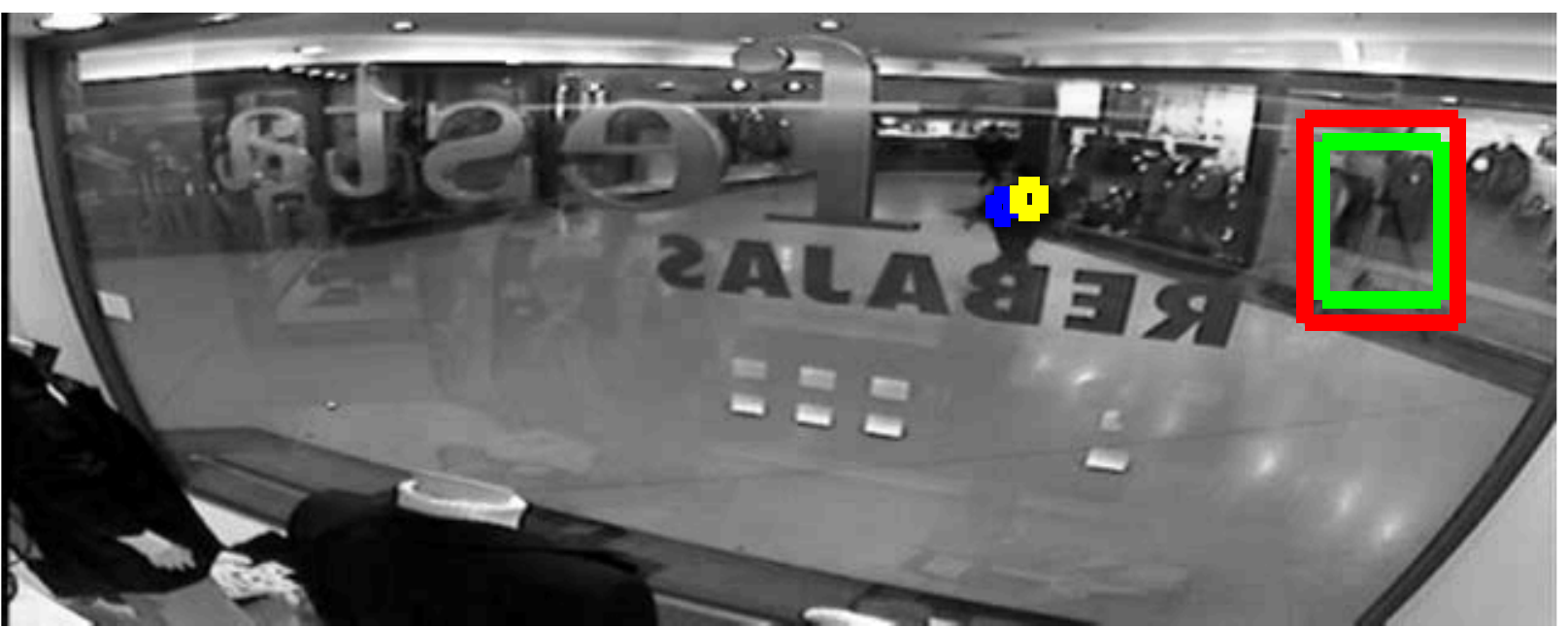}}
      \subfigure[\#385]{\includegraphics[width=0.19\textwidth, height=0.13\textwidth]{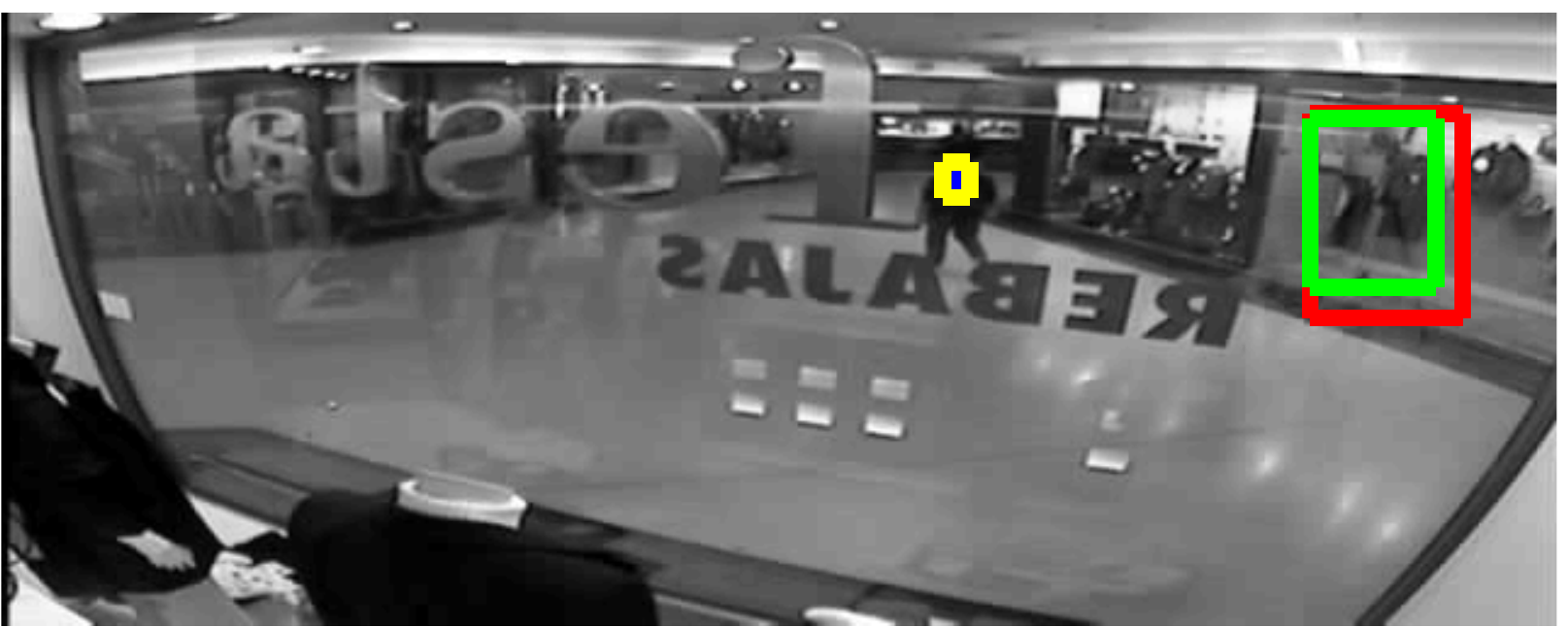}}
      \subfigure[\#1]{\includegraphics[width=0.19\textwidth]{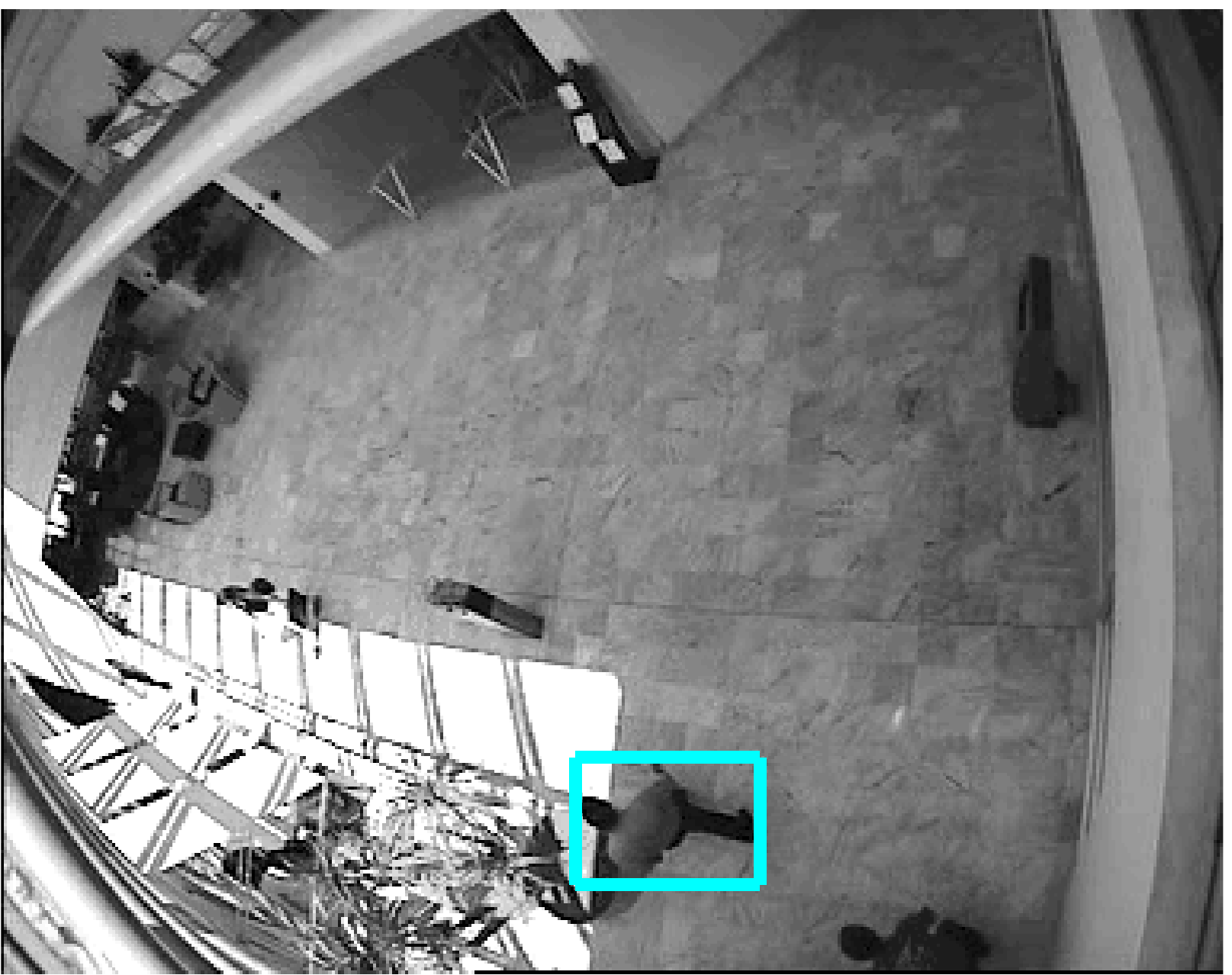}}
      \subfigure[\#46]{\includegraphics[width=0.19\textwidth]{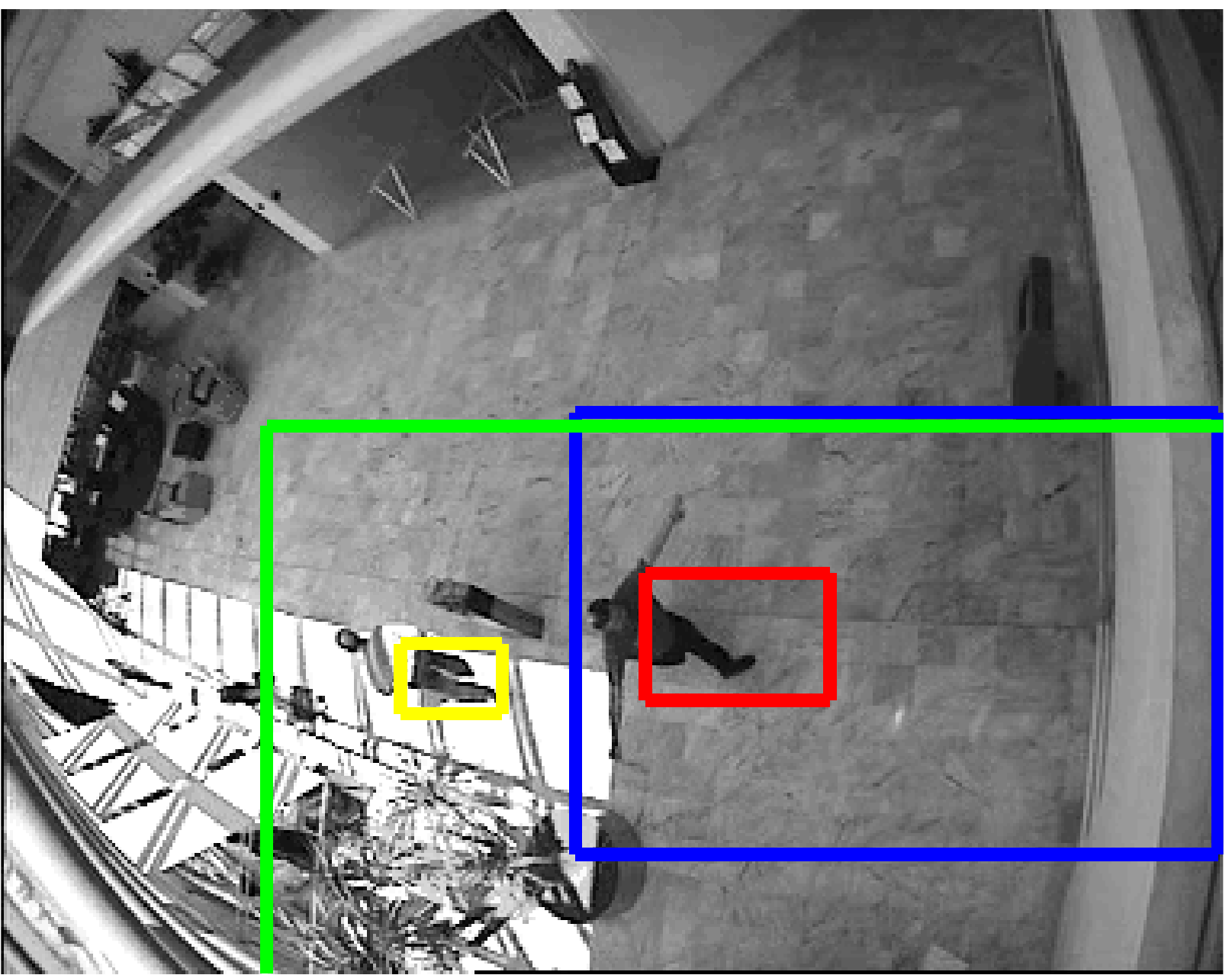}}
      \subfigure[\#91]{\includegraphics[width=0.19\textwidth]{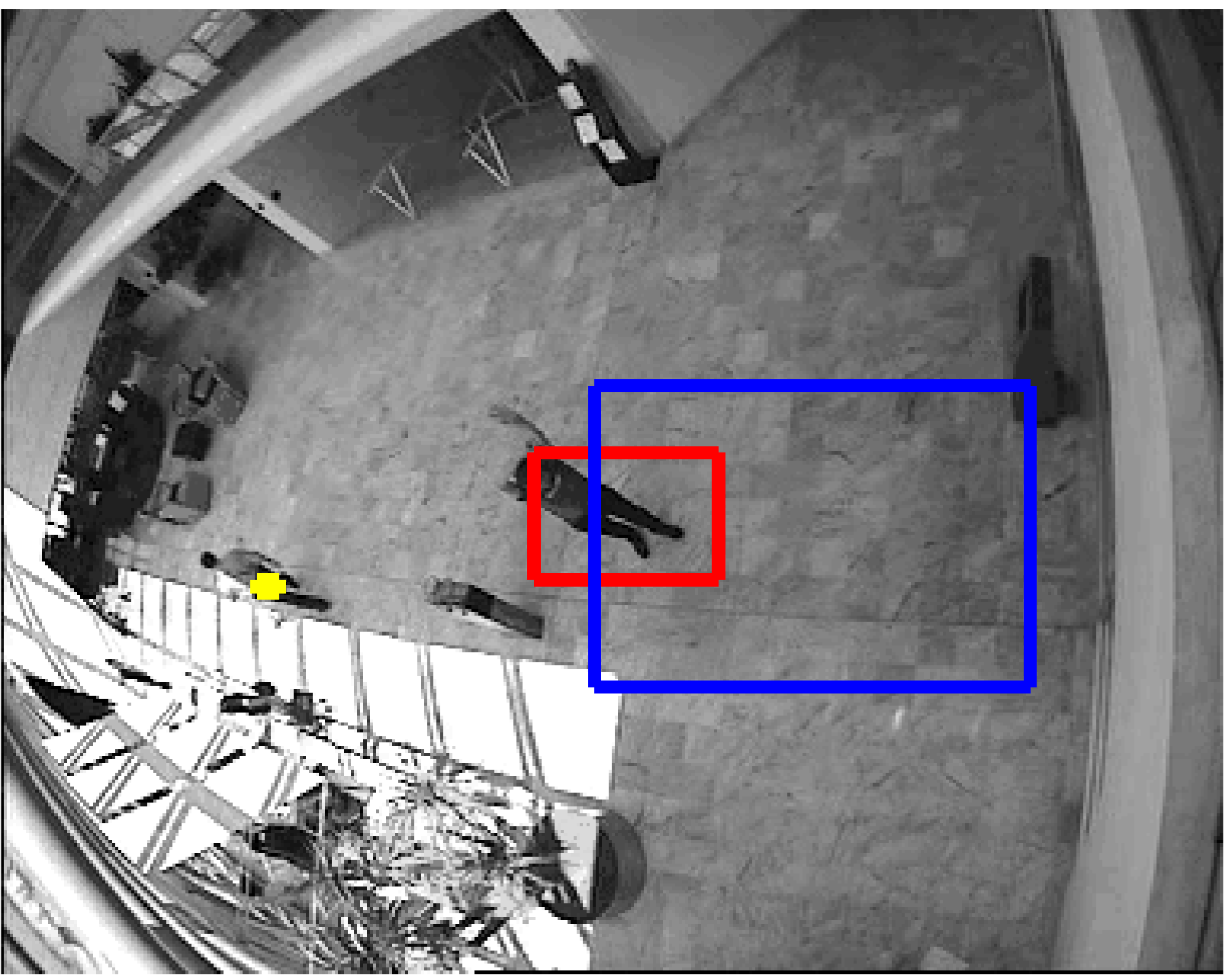}}
      \subfigure[\#136]{\includegraphics[width=0.19\textwidth]{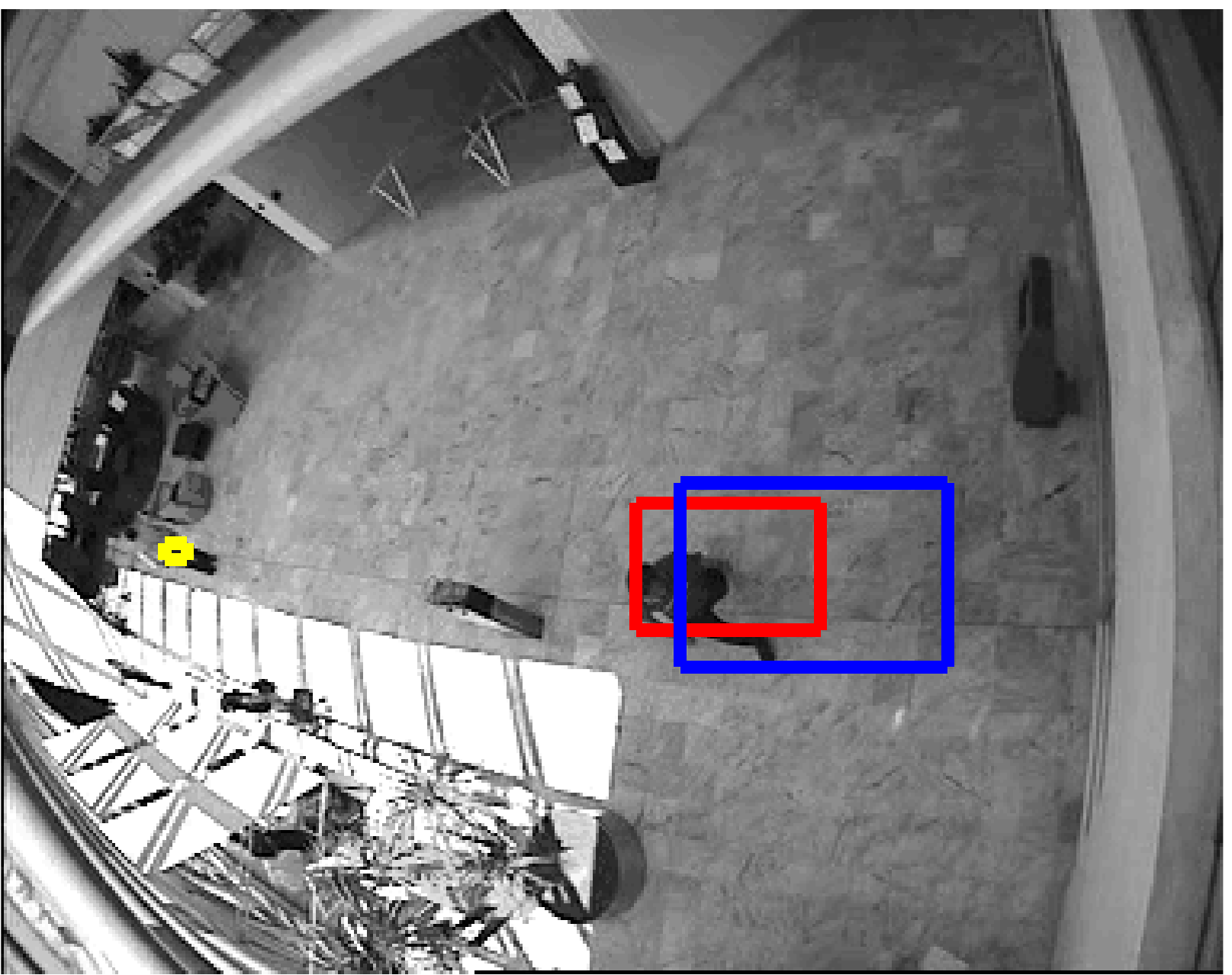}}
      \subfigure[\#166]{\includegraphics[width=0.19\textwidth]{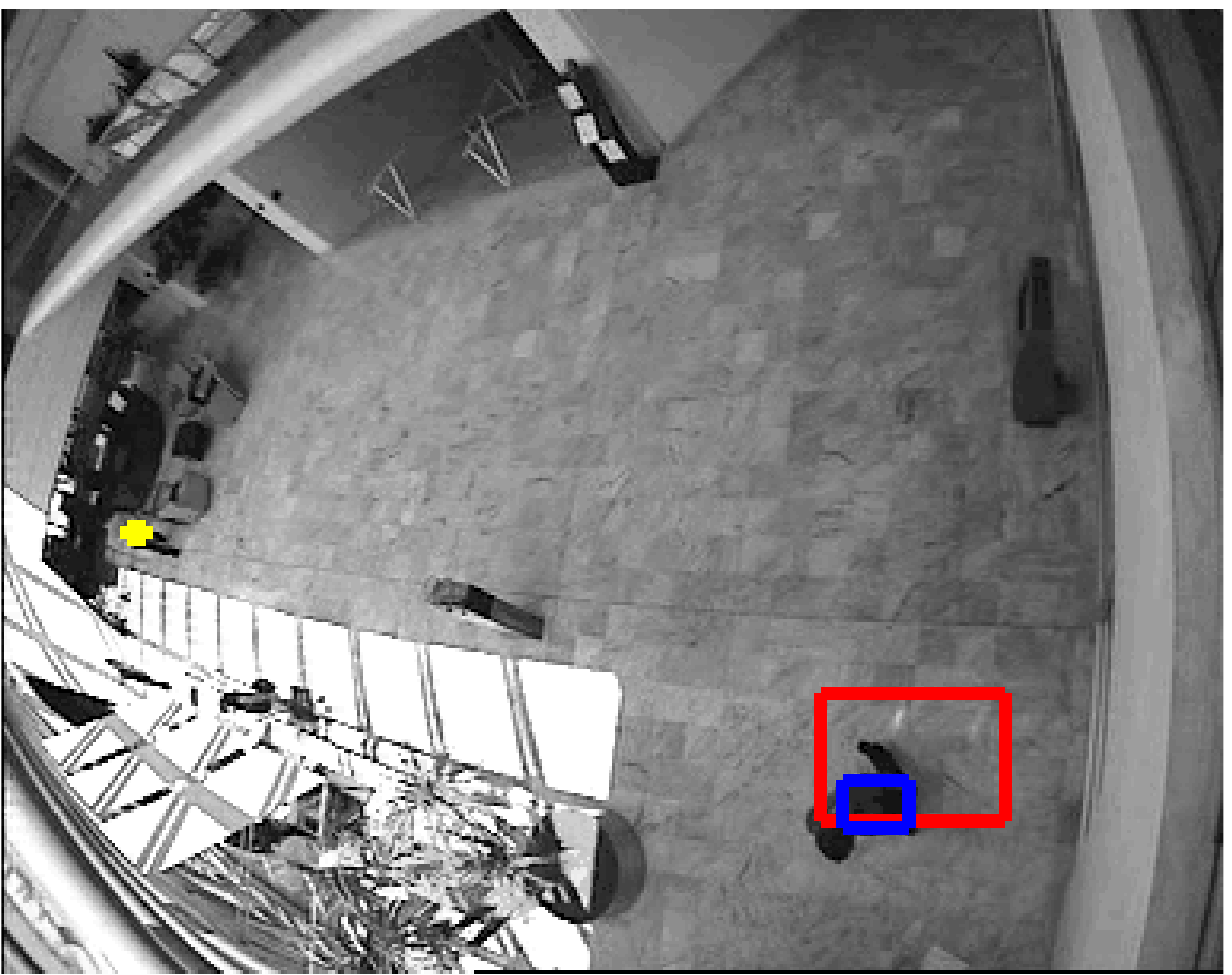}}
    \caption{
            Tracking results shown as rectangles for $6$ video sequences, namely,
            \emph{cubicle}, \emph{dp}, \emph{car4}, \emph{pets2000\_c1},
            \emph{pets2002\_p1} and \emph{pets2004\_p1}. Symbol \#$x$ stands for the $x$th
            frame.  The initial target position is shown in light blue while the red,
            green, dark blue and yellow rectangle denote the tracked area by KMS tracker,
            PF tracker (PN500), RTCST (D100-Rand-PN200) and RTCST-B (D100-Rand-PN200)
            respectively. For a certain tracker, the illustrated result is the one with
            the highest TSP value among all the associated results. PF tracker extends the
            tracking region to the whole scene in the latter frames on
            \emph{pets2004\_p1}, this is why we can not see the green rectangle in
            these frames. RTCST and RTCST-B tracking the similar regions for the last
            frame on \emph{pets2002\_p1} and the yellow rectangle covers the blue one.
         }
      \label{fig:track_frames}
   \end{figure*}

\begin{figure*}[h]
      \centering 
      \subfigure[]{\includegraphics[width=0.328\textwidth]{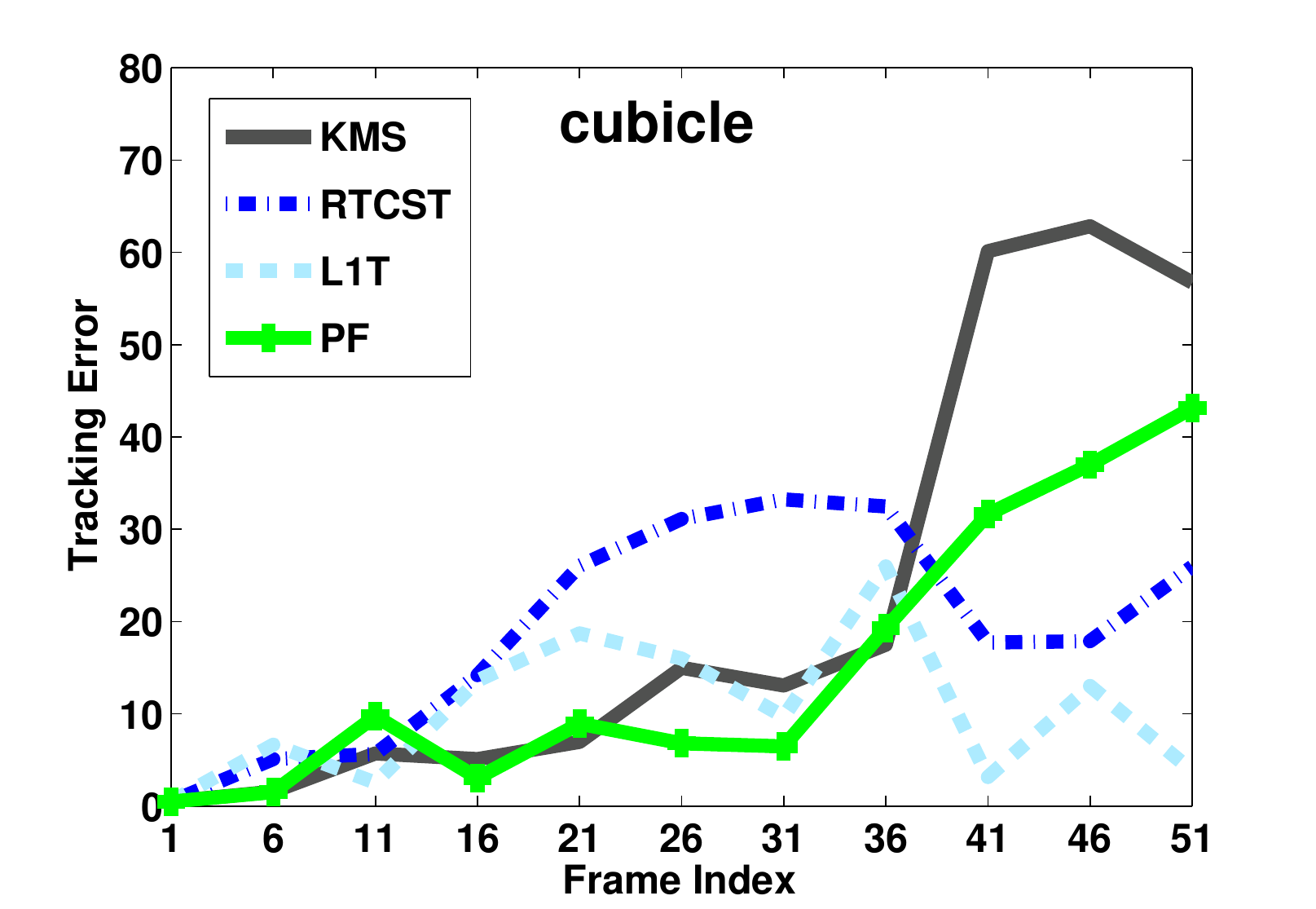}}
      \subfigure[]{\includegraphics[width=0.328\textwidth]{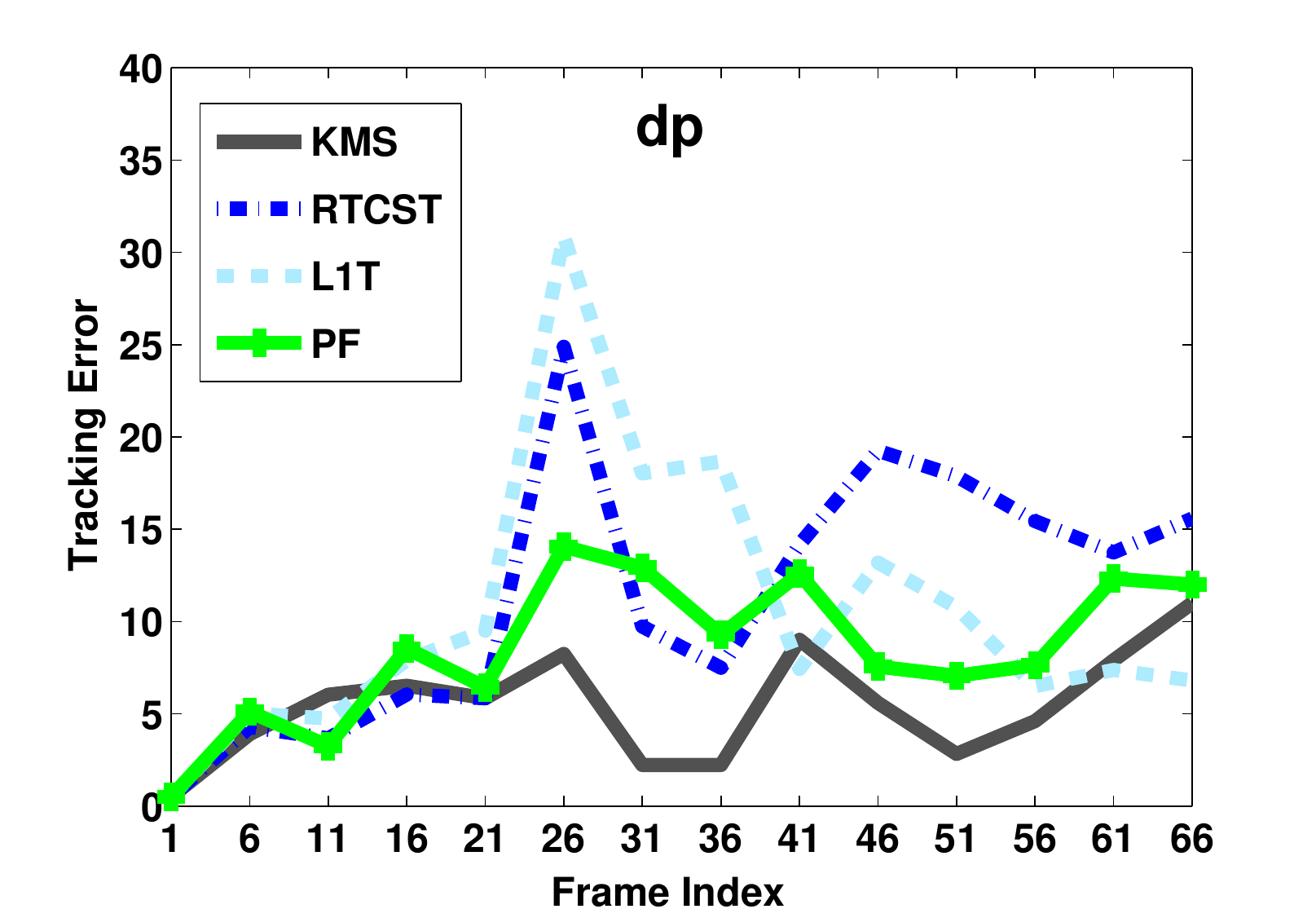}}
      \subfigure[]{\includegraphics[width=0.328\textwidth]{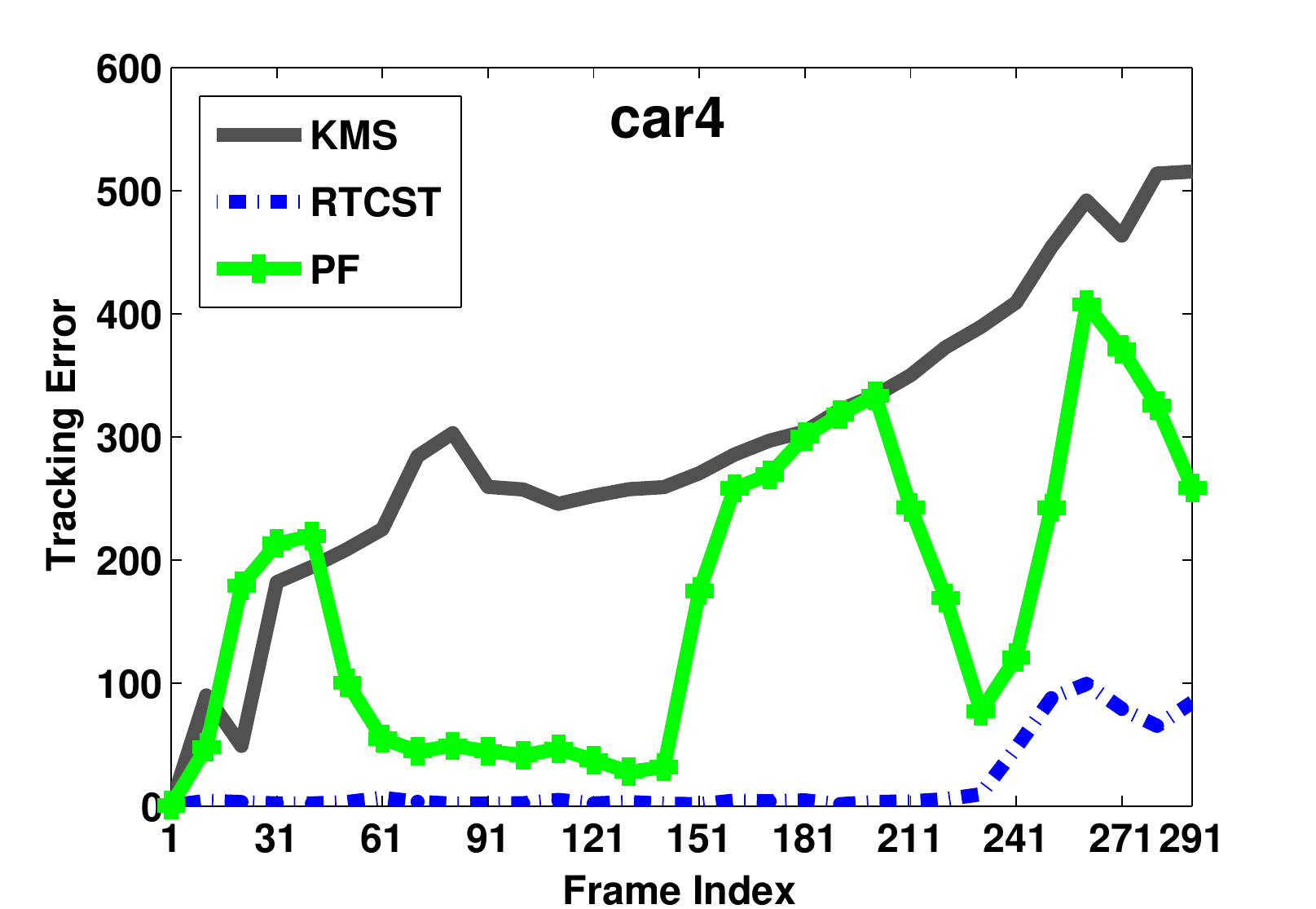}}
      \subfigure[]{\includegraphics[width=0.328\textwidth]{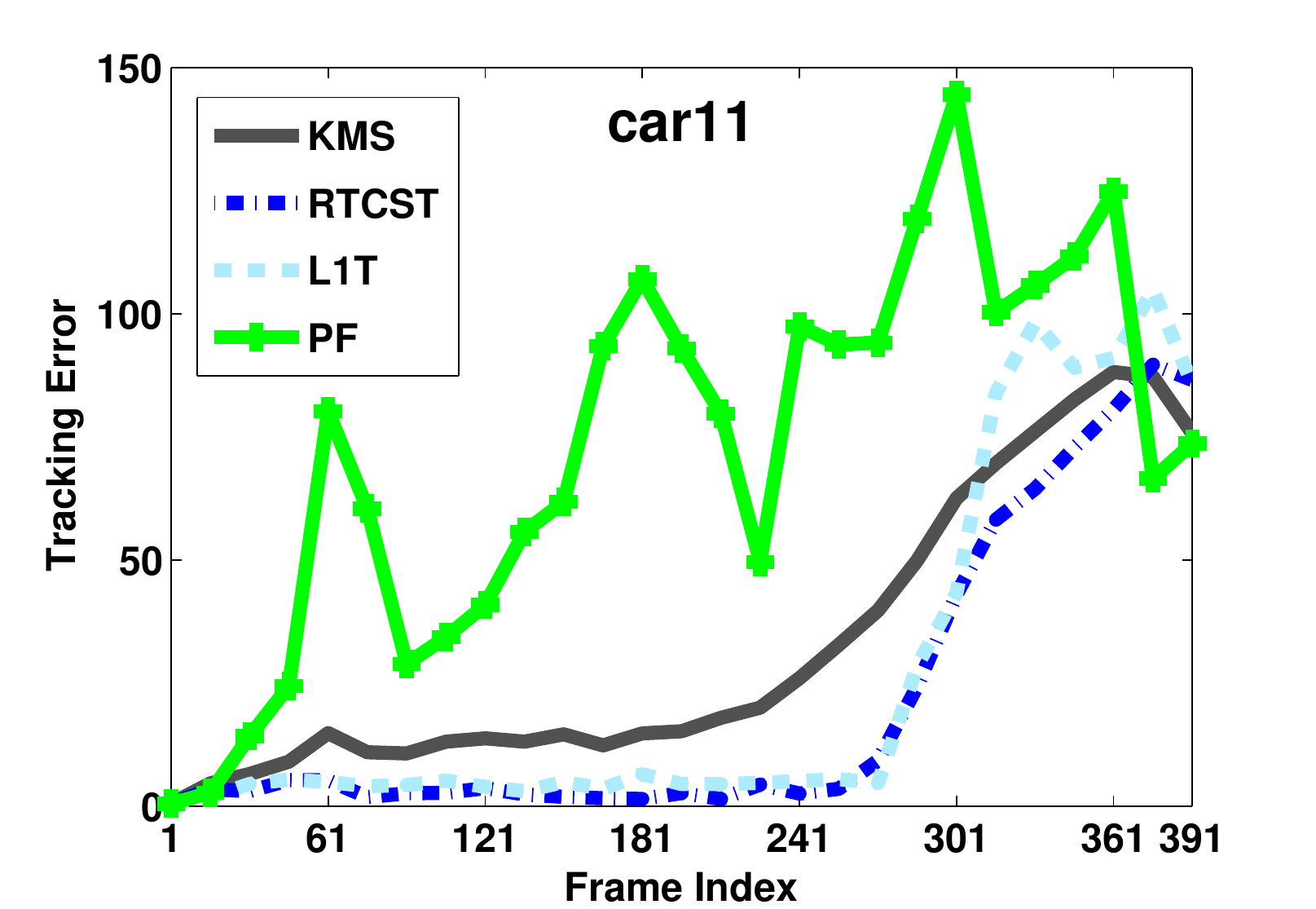}}
      \subfigure[]{\includegraphics[width=0.328\textwidth]{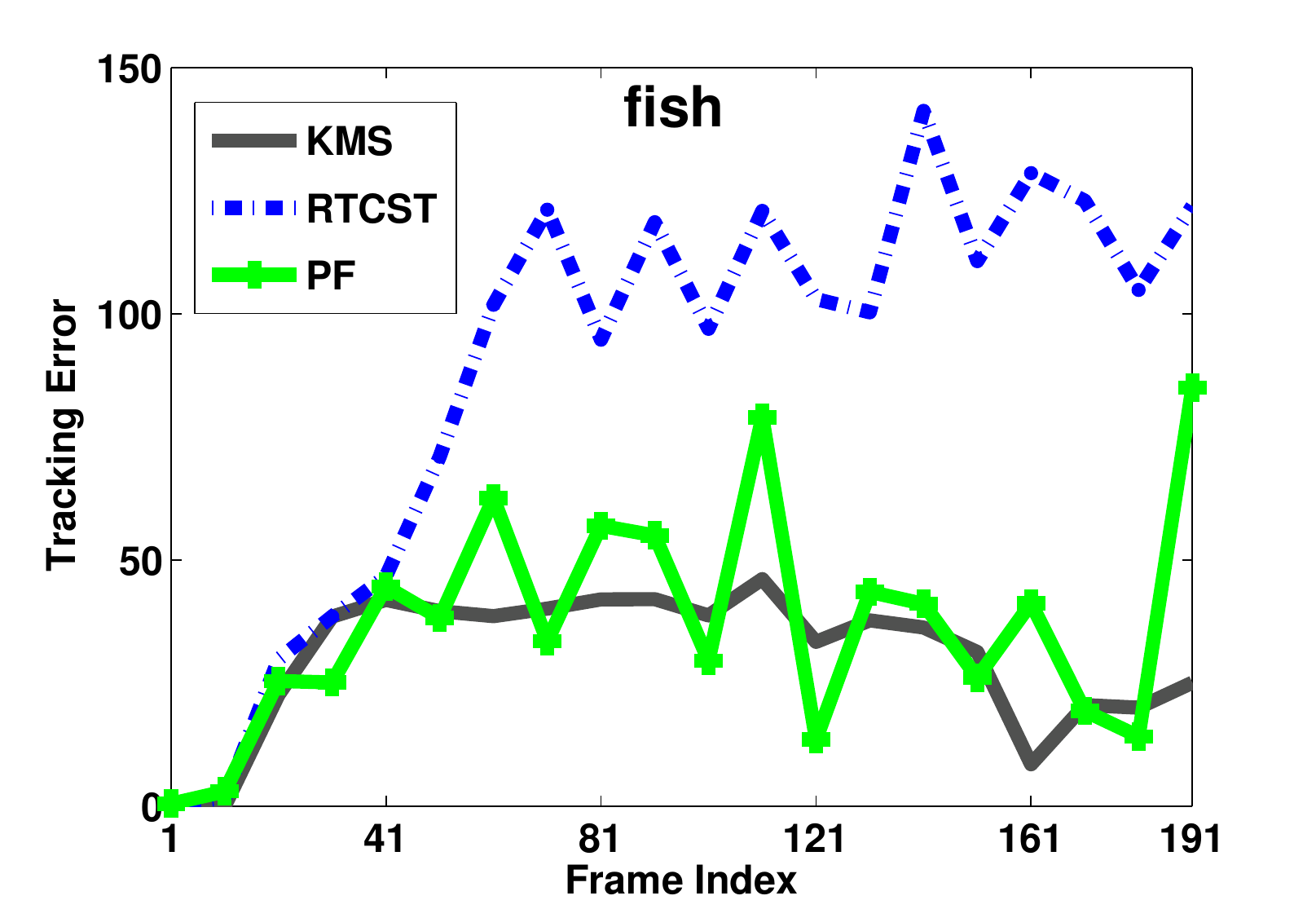}}
      \subfigure[]{\includegraphics[width=0.328\textwidth]{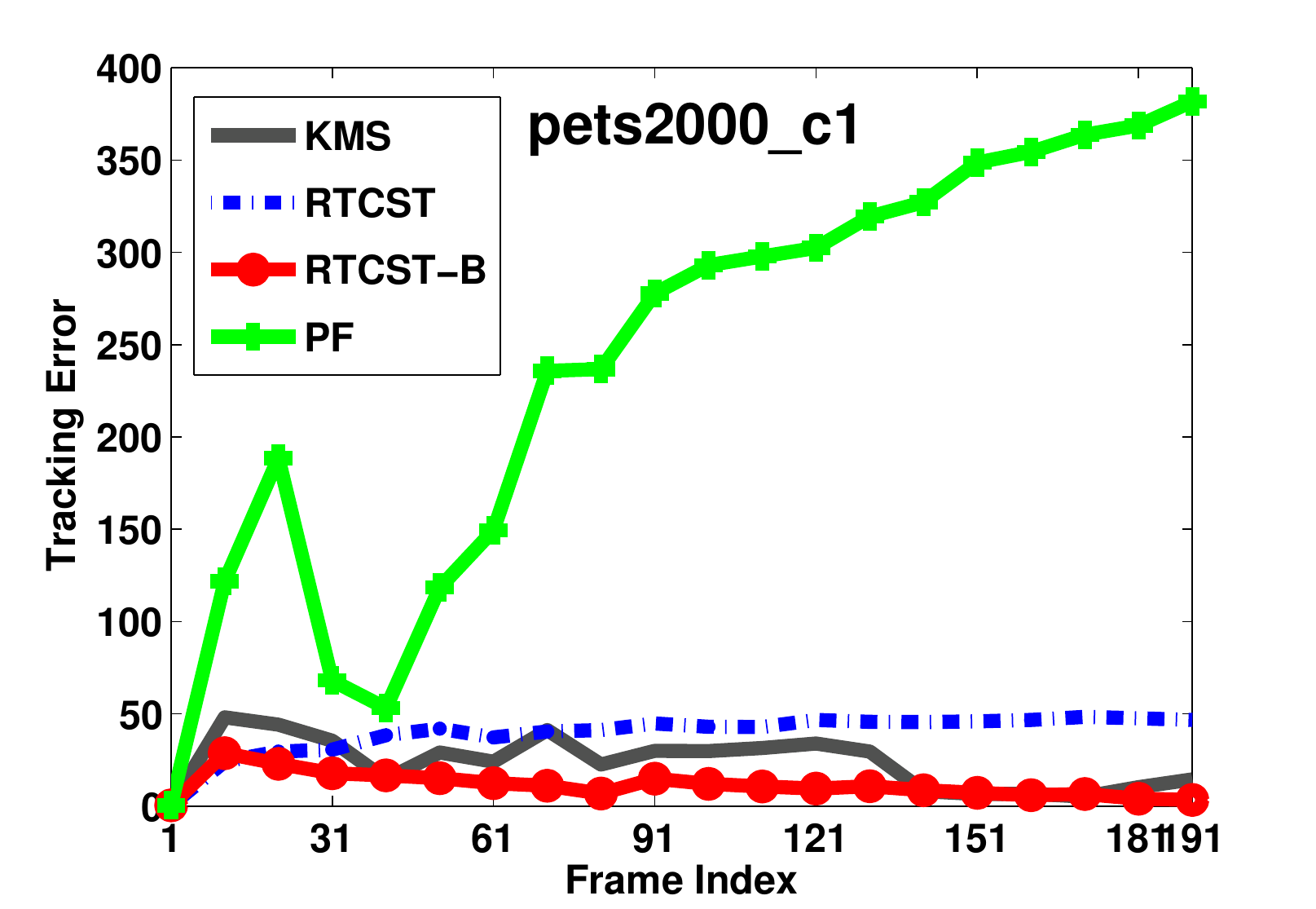}}
      \subfigure[]{\includegraphics[width=0.328\textwidth]{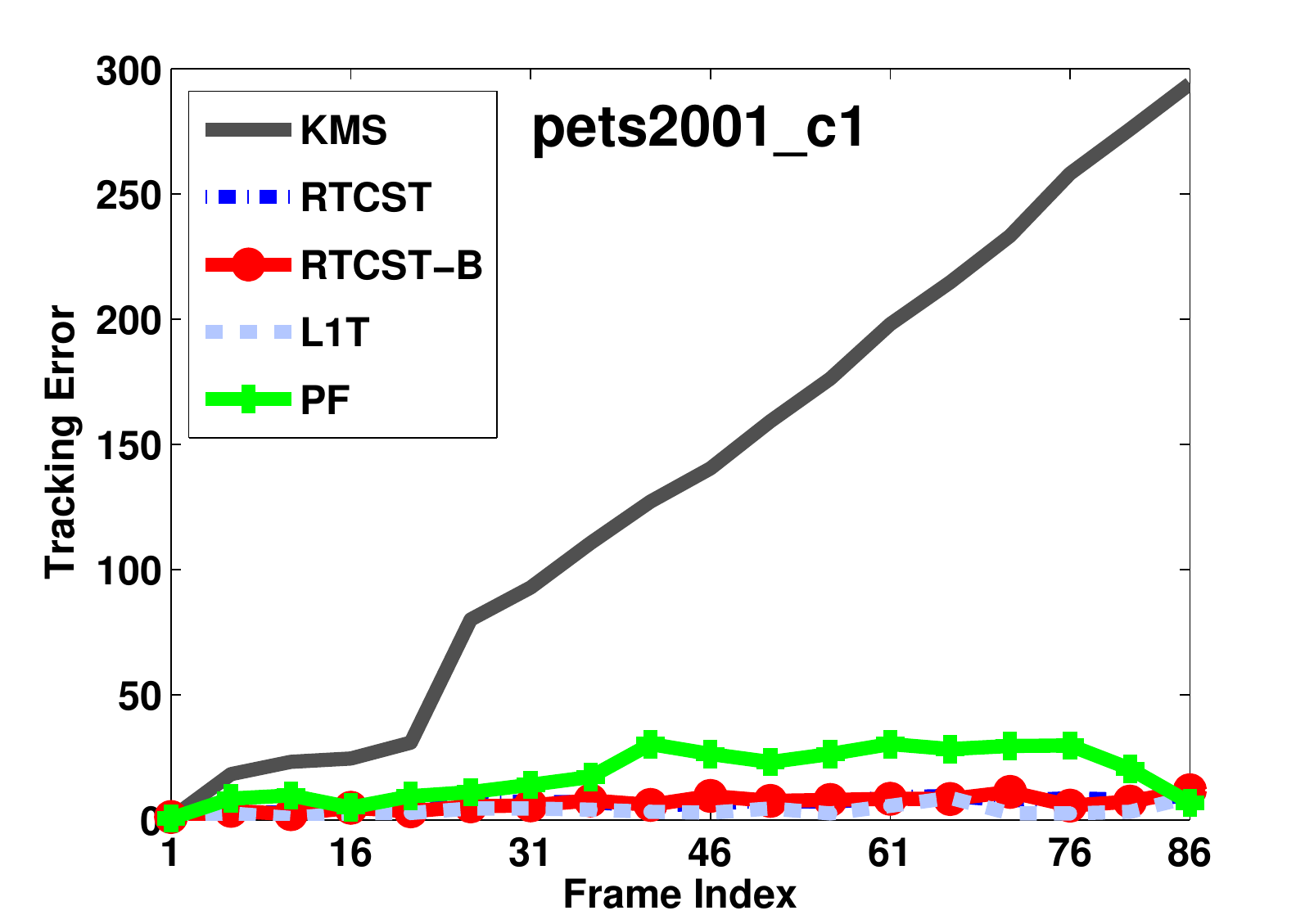}}
      \subfigure[]{\includegraphics[width=0.328\textwidth]{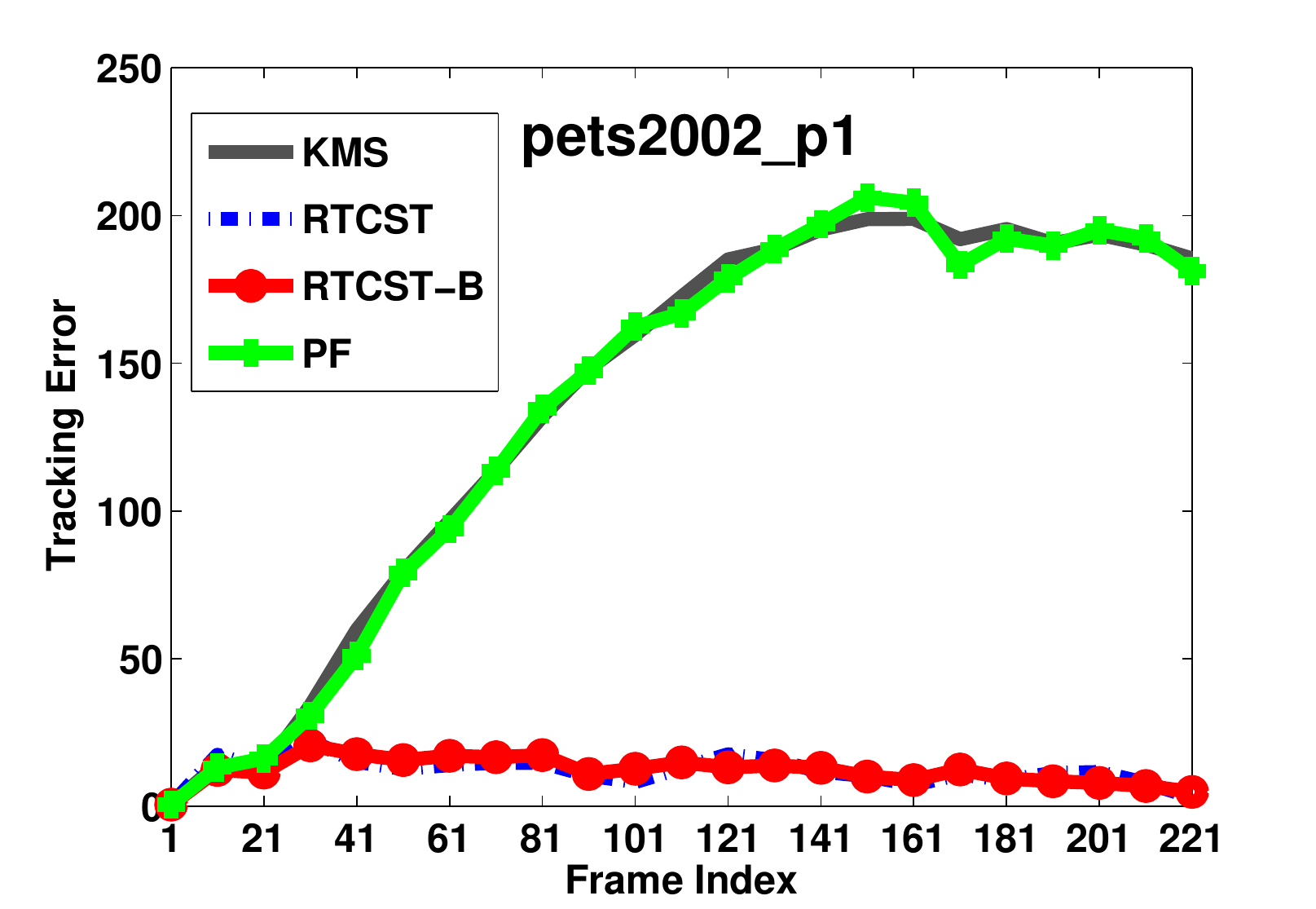}}
      \subfigure[]{\includegraphics[width=0.328\textwidth]{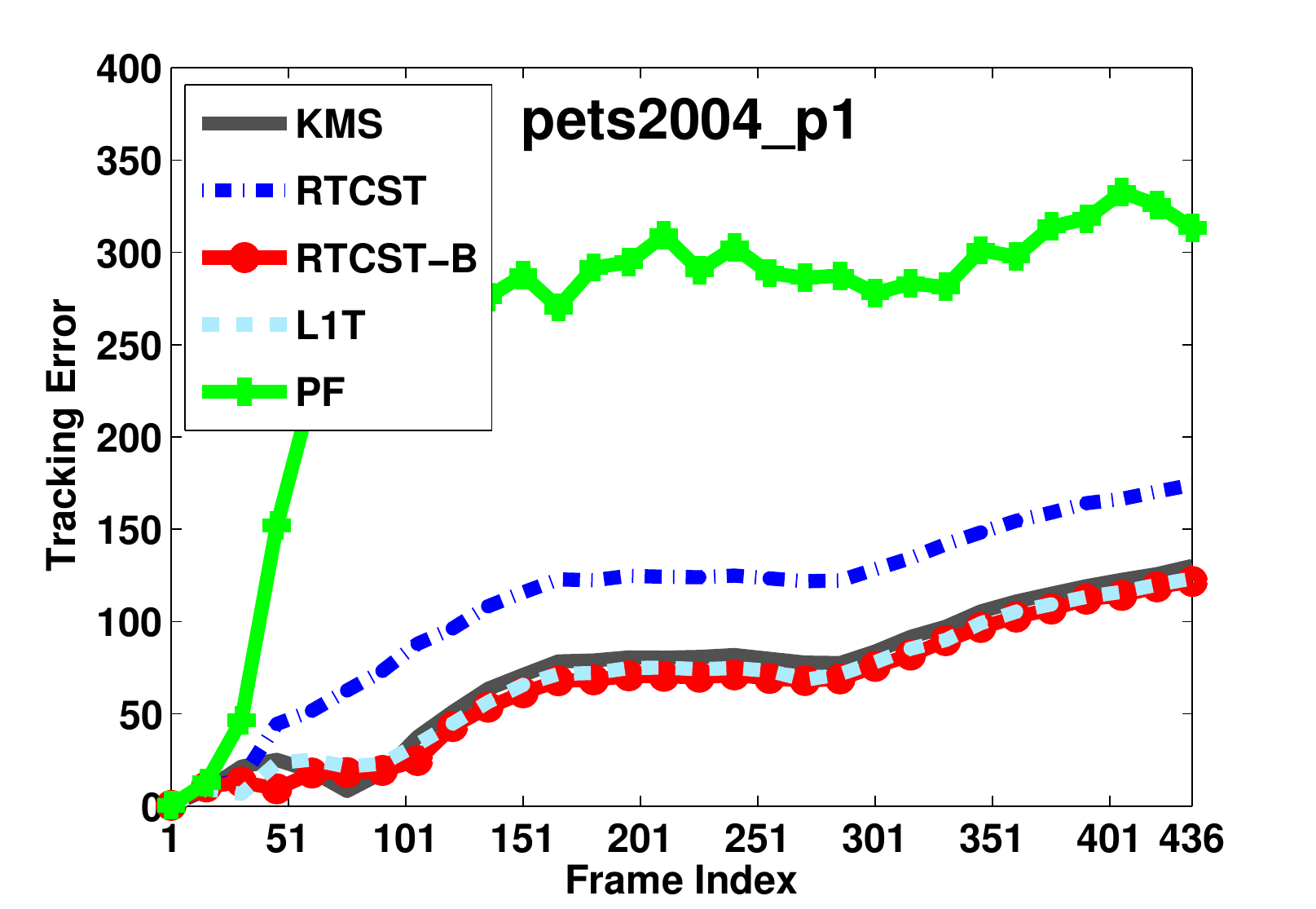}}
      \subfigure[]{\includegraphics[width=0.328\textwidth]{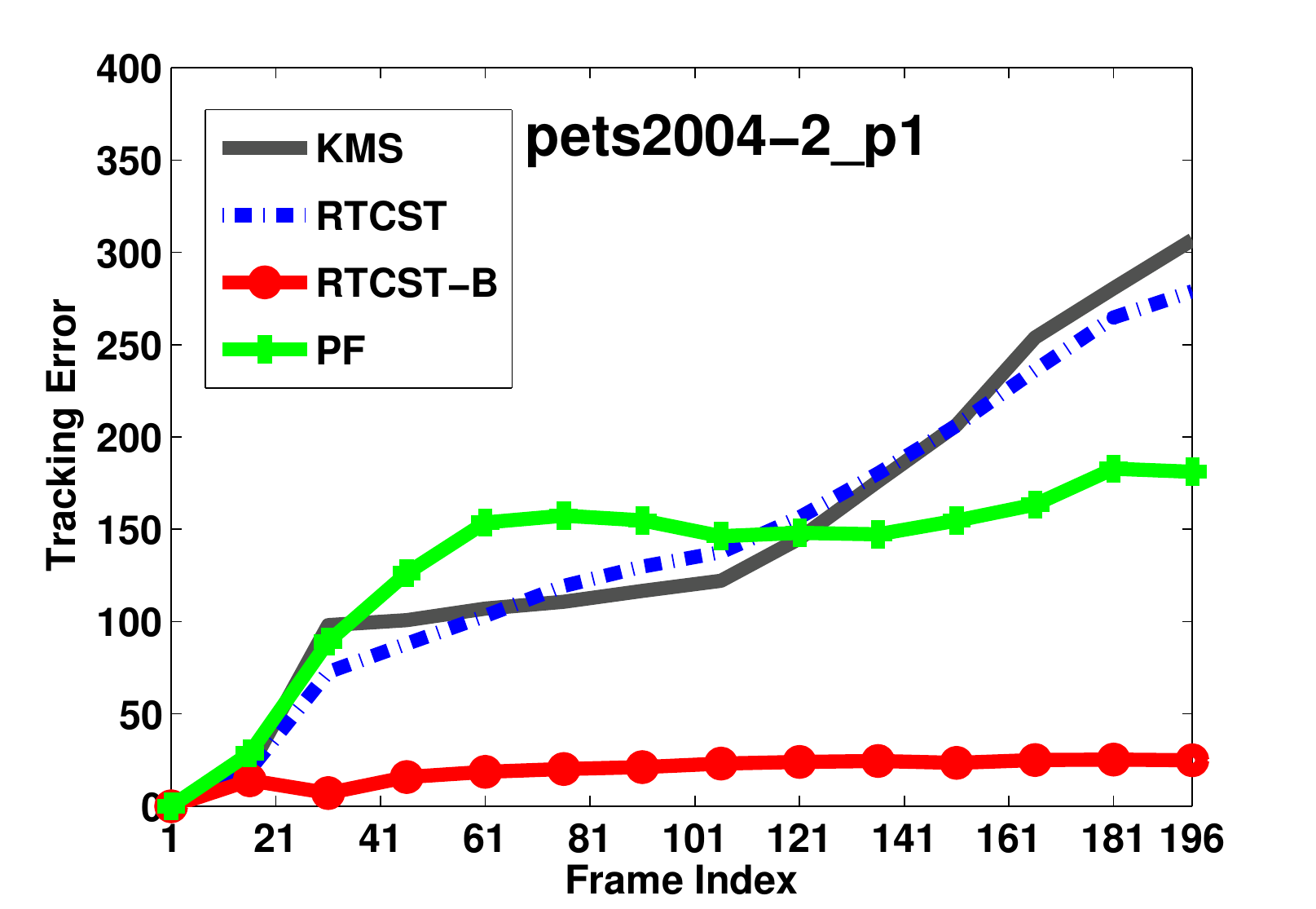}}
      \subfigure[]{\label{subfig:tsl_cubicle}\includegraphics[width=0.328\textwidth]{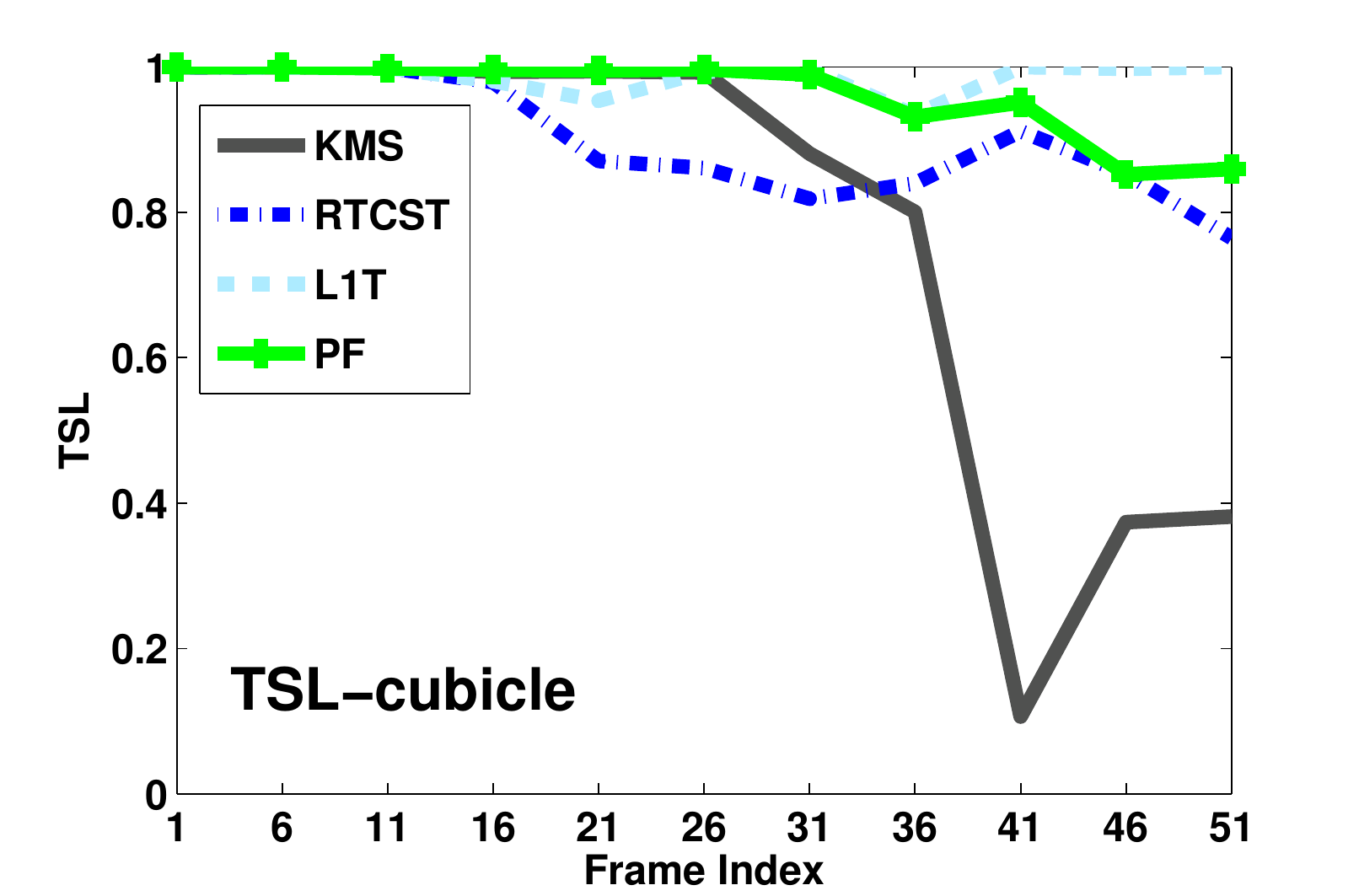}}
      \subfigure[]{\label{subfig:tsl_pets2002}\includegraphics[width=0.328\textwidth]{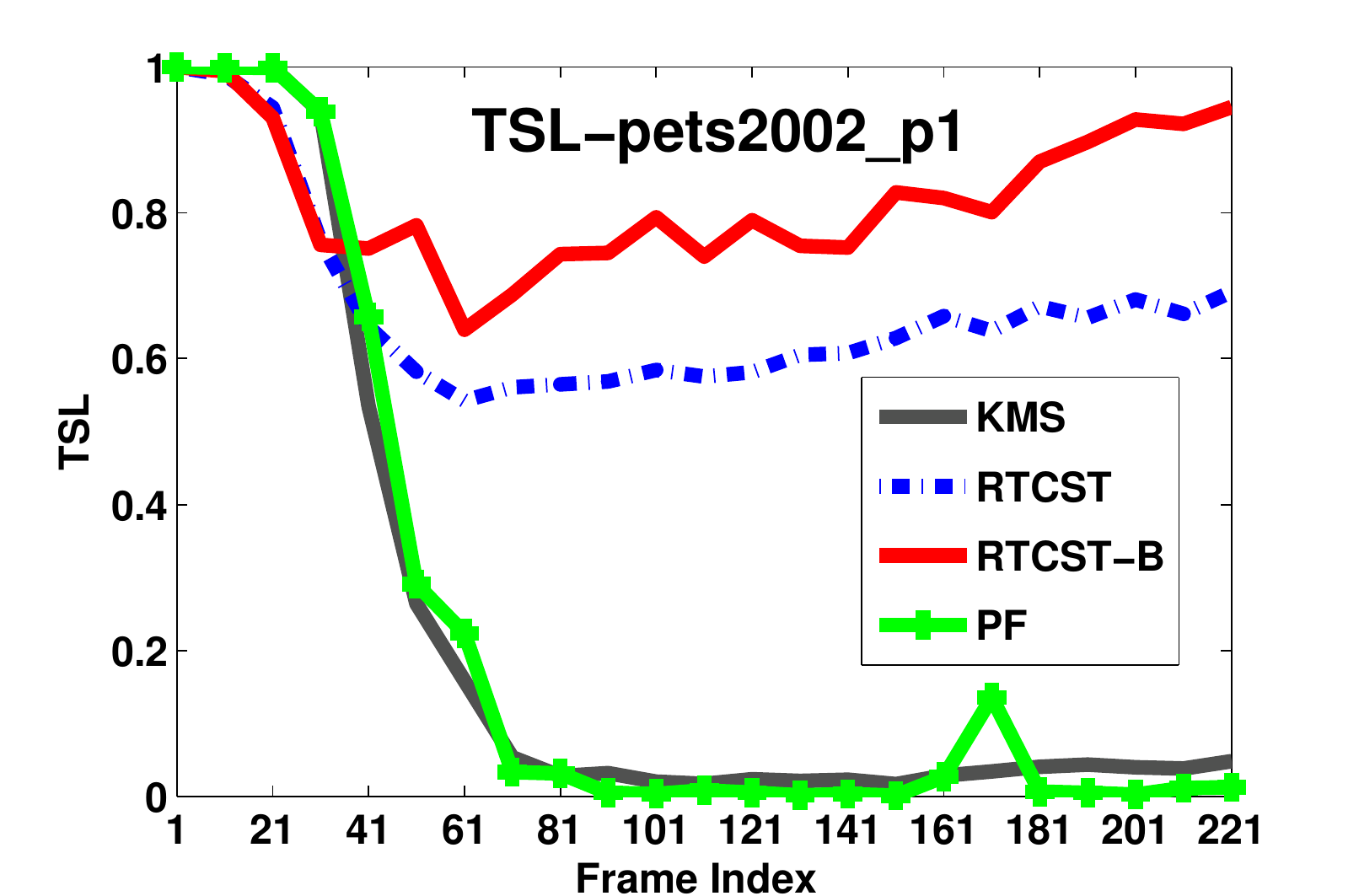}}
      \caption{The tracking errors and TSP values changing along with the frame index. All
              the visual trackers employ the optimal parameters, \ie, $500$ particles for
              PF traker; $200$ particles and Dimension-$100$ for both RTCST and RTCST-B.
               }
         \label{fig:track_error}
    \end{figure*}

    Besides the TSP values, video frames with the tracked regions are listed in
    Figure~\ref{fig:track_frames} while tracking errors changing along with the frame index
    are also plotted in Figure~\ref{fig:track_error}. 
    
    In Figure~\ref{fig:track_frames}, only the best (with the highest average TSP value)
    result is employed to be shown for each tracker. The explicit tracking results support
    the statistics in Table~\ref{tab:accuracy}. RTCST beats KMS tracker and PF tracker on
    \emph{cubicle}, \emph{car4}, \emph{pets2000\_c1} and \emph{pets2002\_p1} and obtain
    the similar performance as its competitors on \emph{dp}. Being facilitated with CSBM,
    RTCS-B always achieves the highest accuracy if it is present. Quite the contrary, the
    traditional trackers fail in some complex scenarios, \eg PF tracker on \emph{car4} and
    \emph{pets2002\_p1}; KMS tracker on \emph{car4} and \emph{pets2002\_p1}. 
    
    From the error curves shown in Figure~\ref{fig:track_error}, we can find that our
    methods beat other visual tracking algorithms on most video sequences except \emph{dp}
    and \emph{fish}. Given that all the trackers perform similarly for \emph{dp} and video
    \emph{fish} is generated with extreme illumination variation which is added
    deliberately, RTCST and RTCST-B could be considered better than their competitors in
    terms of accuracy. 

    To evaluate the new measurement, the TSP curves for \emph{cubicle} and
    \emph{pets2002\_p1} are also available in the Figure~\ref{subfig:tsl_cubicle} and
    Figure~\ref{subfig:tsl_pets2002}. We can see that the TSP value and tracking error
    change oppositely, which is as expected. However, based on TSP, we can verify the
    capacity of single tracker without any ``reference tracker''. This is hard to achieve
    based on tracking error.

    \subsection{Tracking Efficiency}
    \label{subsec:efficiency}

    Efficiency plays a fatal role in real-time visual tracking applications. We record the
    elapsed time of each tracker in our experiment. The time consumptions (in $ms$) for
    processing one frame by the tracking algorithms are reported in
    Table~\ref{tab:run_time}. In the table, huge differences in tracking speed are
    observed. KMS tracker illustrates the highest efficiency with the lowest running speed
    of $83$ \emph{ms per frame} ($83~mspf$). On the contrary, $\ell_1$ tracker (both for
    C-based version and Matlab-based version) is consistently slower than $14000~mspf$ due
    to the high computational complexity. Being equipped with OMP and dimension reduction
    manners, RTCST and RTCST-B are able to accelerate the original CS-based tracker by
    $117.3$ (\emph{dp}) to $6271.2$ (\emph{pets2004\_p1}) times. The speed range for RTCST
    is $54\sim 968~mspf$ while that for RTCST-B is $85\sim 534~mspf$. PF tracker shows
    unstable efficiency among all the tests. Its running speed varies from $37$ to $1727
    ~mspf$ for the experiment with $500$ particles. Supposed that the speed threshold for
    real-time application is $100~mspf$, most of the traditional methods and a part of our
    methods are qualified. $\ell_1$ tracker could not be viewed as ``real-time'' from any
    perspective. 
    
    Moreover, since RTCST and RTCST-B are implemented in Matlab with single core, their
    running speeds could be increased remarkably by employing C/C++ language and multiple
    cores. Actually, the speed of Matlab-based $\ell_1$ is already raised by $3.7$
    (\emph{pets2004\_p1}) to $8.4$ (\emph{cubicle}) times in its C/C++ counterpart even
    though only one core is used. If we conservatively predict $10$-time speed growth ,
    both RTCST and RTCST-B will be qualified for real-time application in all the
    circumstances. 

    \begin{table*}[]
    \centering
    \caption
    {
      Running time of visual trackers for one frame (ms). Note that every time consumption
      based on Matlab implementation is labeled by signal ``$\star$''. The notations of
      algorithm names are the same to those used in Table~\ref{tab:accuracy}. 
    }
    \resizebox{1.000\textwidth}{!}
    {
    \begin{tabular}{ c | c | l | l | l | l | l | l | l | l | l | l }

    \hline\hline
\multicolumn{2}{c|}{}   & \bf cubicle    & \bf dp    & \bf car4    & \bf car11    & \bf
fish    & \bf pets2000\_c1    & \bf pets2001\_c1    & \bf pets2002\_p1    & \bf
pets2004\_p1    & \bf pets2004-2\_p1   \\
\cline{1-12}
\multicolumn{2}{c|}{\bf KMS}  & $22\pm0$  & $17\pm0$  & $60\pm0$  & $15\pm0$  & $36\pm0$  & $83\pm0$  & $40\pm0$  & $22\pm0$  & $21\pm0$  & $31\pm0$ \\
\cline{1-12}
{\bf \multirow{3}{*}{PF}}  
& PN100  & $18\pm0$  & $17\pm0$  & $173\pm0$  & $22\pm0$  & $39\pm0$  & $199\pm0$  & $35\pm0$  & $28\pm0$  & $53\pm0$  & $381\pm0$  \\
& PN200  & $27\pm0$  & $20\pm0$  & $321\pm0$  & $32\pm0$  & $56\pm0$  & $279\pm0$  & $37\pm0$  & $44\pm0$  & $82\pm0$  & $734\pm0$  \\
& PN500  & $40\pm0$  & $37\pm0$  & $770\pm0$  & $65\pm0$  & $139\pm0$  & $631\pm0$  & $45\pm0$  & $83\pm0$  & $184\pm0$  & $1727\pm0$ \\
\cline{1-12}
{\bf \multirow{2}{*}{R-D25-Rand}}  
& PN100  & ${84\pm3}^{\star}$  & ${100\pm4}^{\star}$  & ${115\pm2}^{\star}$  & ${103\pm4}^{\star}$  & ${114\pm4}^{\star}$  & ${105\pm3}^{\star}$  & ${103\pm4}^{\star}$  & ${131\pm3}^{\star}$  & ${109\pm2}^{\star}$  & ${117\pm3}^{\star}$  \\
& PN200  & ${148\pm9}^{\star}$  & ${152\pm11}^{\star}$  & ${193\pm5}^{\star}$  & ${186\pm11}^{\star}$  & ${198\pm15}^{\star}$  & ${186\pm9}^{\star}$  & ${177\pm11}^{\star}$  & ${223\pm8}^{\star}$  & ${168\pm7}^{\star}$  & ${198\pm5}^{\star}$  \\
\cline{1-12}
{\bf \multirow{2}{*}{R-D50-Rand}}  
& PN100  & ${155\pm4}^{\star}$  & ${171\pm4}^{\star}$  & ${189\pm4}^{\star}$  & ${197\pm5}^{\star}$  & ${168\pm10}^{\star}$  & ${192\pm7}^{\star}$  & ${187\pm5}^{\star}$  & ${188\pm5}^{\star}$  & ${169\pm5}^{\star}$  & ${167\pm4}^{\star}$  \\
& PN200  & ${276\pm18}^{\star}$  & ${301\pm19}^{\star}$  & ${337\pm13}^{\star}$  & ${358\pm20}^{\star}$  & ${333\pm33}^{\star}$  & ${334\pm23}^{\star}$  & ${334\pm24}^{\star}$  & ${347\pm15}^{\star}$  & ${286\pm13}^{\star}$  & ${336\pm13}^{\star}$  \\
\cline{1-12}
{\bf \multirow{2}{*}{R-D100-Rand}}  
& PN100  & ${477\pm21}^{\star}$  & ${474\pm17}^{\star}$  & ${480\pm23}^{\star}$  & ${535\pm10}^{\star}$  & ${435\pm32}^{\star}$  & ${473\pm11}^{\star}$  & ${496\pm26}^{\star}$  & ${478\pm18}^{\star}$  & ${439\pm17}^{\star}$  & ${481\pm13}^{\star}$  \\
& PN200  & ${825\pm97}^{\star}$  & ${742\pm101}^{\star}$  & ${939\pm21}^{\star}$  & ${968\pm71}^{\star}$  & ${870\pm101}^{\star}$  & ${798\pm86}^{\star}$  & ${863\pm94}^{\star}$  & ${863\pm42}^{\star}$  & ${681\pm58}^{\star}$  & ${872\pm29}^{\star}$  \\
\cline{1-12}
{\bf \multirow{2}{*}{R-D25-Hash}}  
& PN100  & ${91\pm3}^{\star}$  & ${92\pm4}^{\star}$  & ${109\pm3}^{\star}$  & ${109\pm3}^{\star}$  & ${103\pm5}^{\star}$  & ${110\pm4}^{\star}$  & ${108\pm4}^{\star}$  & ${131\pm3}^{\star}$  & ${102\pm3}^{\star}$  & ${108\pm2}^{\star}$  \\
& PN200  & ${166\pm9}^{\star}$  & ${161\pm7}^{\star}$  & ${172\pm7}^{\star}$  & ${191\pm13}^{\star}$  & ${193\pm14}^{\star}$  & ${204\pm9}^{\star}$  & ${195\pm12}^{\star}$  & ${217\pm10}^{\star}$  & ${161\pm9}^{\star}$  & ${194\pm6}^{\star}$  \\
\cline{1-12}
{\bf \multirow{2}{*}{R-D50-Hash}}  
& PN100  & ${57\pm1}^{\star}$  & ${54\pm1}^{\star}$  & ${67\pm1}^{\star}$  & ${62\pm2}^{\star}$  & ${56\pm1}^{\star}$  & ${70\pm1}^{\star}$  & ${65\pm1}^{\star}$  & ${70\pm1}^{\star}$  & ${59\pm2}^{\star}$  & ${59\pm1}^{\star}$  \\
& PN200  & ${96\pm2}^{\star}$  & ${96\pm2}^{\star}$  & ${118\pm1}^{\star}$  & ${114\pm2}^{\star}$  & ${101\pm2}^{\star}$  & ${118\pm2}^{\star}$  & ${114\pm3}^{\star}$  & ${116\pm2}^{\star}$  & ${109\pm3}^{\star}$  & ${108\pm1}^{\star}$  \\
\cline{1-12}
{\bf \multirow{2}{*}{R-D100-Hash}}  
& PN100  & ${73\pm1}^{\star}$  & ${85\pm2}^{\star}$  & ${87\pm2}^{\star}$  & ${86\pm1}^{\star}$  & ${84\pm3}^{\star}$  & ${96\pm3}^{\star}$  & ${83\pm2}^{\star}$  & ${96\pm5}^{\star}$  & ${78\pm2}^{\star}$  & ${82\pm1}^{\star}$  \\
& PN200  & ${138\pm4}^{\star}$  & ${154\pm4}^{\star}$  & ${148\pm3}^{\star}$  & ${156\pm2}^{\star}$  & ${157\pm2}^{\star}$  & ${162\pm4}^{\star}$  & ${146\pm4}^{\star}$  & ${169\pm3}^{\star}$  & ${134\pm2}^{\star}$  & ${159\pm1}^{\star}$  \\
\cline{1-12}
{\bf \multirow{2}{*}{RB-D25-Rand}}  
& PN100  & $-$  & $-$  & $-$  & $-$  & $-$  & ${167\pm5}^{\star}$  & ${175\pm6}^{\star}$  & ${204\pm6}^{\star}$  & ${142\pm23}^{\star}$  & ${184\pm4}^{\star}$  \\
& PN200  & $-$  & $-$  & $-$  & $-$  & $-$  & ${305\pm11}^{\star}$  & ${330\pm19}^{\star}$  & ${316\pm26}^{\star}$  & ${237\pm32}^{\star}$  & ${331\pm18}^{\star}$  \\
\cline{1-12}
{\bf \multirow{2}{*}{RB-D50-Rand}}  
& PN100  & $-$  & $-$  & $-$  & $-$  & $-$  & ${187\pm7}^{\star}$  & ${228\pm4}^{\star}$  & ${222\pm7}^{\star}$  & ${157\pm33}^{\star}$  & ${211\pm4}^{\star}$  \\
& PN200  & $-$  & $-$  & $-$  & $-$  & $-$  & ${389\pm27}^{\star}$  & ${500\pm29}^{\star}$  & ${397\pm28}^{\star}$  & ${295\pm74}^{\star}$  & ${427\pm21}^{\star}$  \\
\cline{1-12}
{\bf \multirow{2}{*}{RB-D100-Rand}}  
& PN100  & $-$  & $-$  & $-$  & $-$  & $-$  & ${215\pm4}^{\star}$  & ${246\pm3}^{\star}$  & ${248\pm7}^{\star}$  & ${148\pm36}^{\star}$  & ${253\pm8}^{\star}$  \\
& PN200  & $-$  & $-$  & $-$  & $-$  & $-$  & ${456\pm17}^{\star}$  & ${534\pm23}^{\star}$  & ${438\pm38}^{\star}$  & ${318\pm75}^{\star}$  & ${461\pm45}^{\star}$  \\
\cline{1-12}
{\bf \multirow{2}{*}{RB-D25-Hash}}  
& PN100  & $-$  & $-$  & $-$  & $-$  & $-$  & ${162\pm7}^{\star}$  & ${177\pm8}^{\star}$  & ${180\pm11}^{\star}$  & ${131\pm27}^{\star}$  & ${178\pm8}^{\star}$  \\
& PN200  & $-$  & $-$  & $-$  & $-$  & $-$  & ${274\pm18}^{\star}$  & ${377\pm28}^{\star}$  & ${306\pm29}^{\star}$  & ${227\pm41}^{\star}$  & ${351\pm15}^{\star}$  \\
\cline{1-12}
{\bf \multirow{2}{*}{RB-D50-Hash}}  
& PN100  & $-$  & $-$  & $-$  & $-$  & $-$  & ${95\pm2}^{\star}$  & ${88\pm3}^{\star}$  & ${106\pm2}^{\star}$  & ${85\pm2}^{\star}$  & ${94\pm1}^{\star}$  \\
& PN200  & $-$  & $-$  & $-$  & $-$  & $-$  & ${174\pm5}^{\star}$  & ${166\pm2}^{\star}$  & ${176\pm3}^{\star}$  & ${154\pm8}^{\star}$  & ${174\pm3}^{\star}$ \\
\cline{1-12}
{\bf \multirow{2}{*}{RB-D100-Hash}}  
& PN100  & $-$  & $-$  & $-$  & $-$  & $-$  & ${121\pm2}^{\star}$  & ${106\pm2}^{\star}$  & ${127\pm3}^{\star}$  & ${111\pm3}^{\star}$  & ${114\pm2}^{\star}$  \\
& PN200  & $-$  & $-$  & $-$  & $-$  & $-$  & ${220\pm6}^{\star}$  & ${211\pm7}^{\star}$  & ${229\pm5}^{\star}$  & ${207\pm8}^{\star}$  & ${217\pm5}^{\star}$ \\
\cline{1-12}
\multicolumn{2}{c|}{\bf L1T-Matlab}  & ${2.7\e{5}\pm1255}^{\star}$  & ${8.7\e{4}\pm1660}^{\star}$  & $-$  & ${1.8\e{5}\pm2402}^{\star}$  & $-$  & $-$  & ${1.6\e{5}\pm1944}^{\star}$  & $-$  & ${3.7\e{5}\pm1857}^{\star}$  & $-$ \\
\cline{1-12}
\multicolumn{2}{c|}{\bf L1T-C++}  & ${3.2\e{4}\pm506}$  & ${1.4\e{4}\pm320}$  & $-$  & ${3.8\e{4}\pm1417}$  & $-$  & $-$  & ${3.4\e{4}\pm484}$  & $-$  & ${1.02\e{5}\pm607}$  & $-$ \\
    \hline\hline
    \end{tabular}
    }
    \label{tab:run_time}
    \end{table*}

    \subsection{Tracking Robustness}
    \label{subsec:robustness}

    As mentioned before, no trick is played to select the initial target region. The first
    region $R$ should always be the minimum bounding box covers the whole target. Nonetheless,
    the bounding box could merely obtained manually, and hence, approximately. In
    practice, the selection error is unavoidable. If the visual tracker is not robust
    enough, minor selection error would lead to massive deviation with respect to tracking
    performance. We design a new experiment to test the robustness of tracking algorithms.
    In every repetition of the experiment, a fluctuation vector $\boldsymbol{\delta} =
    [\delta_l, \delta_r, \delta_s]$, is generated randomly as
    \begin{equation*}
    \setlength{\abovedisplayskip}{0.1cm}
    \setlength{\belowdisplayskip}{0.1cm}
      \delta_l \sim \mathcal{N}(0,~\omega), ~ \delta_t \sim \mathcal{N}(0,~\omega), ~
      \delta_s \sim \mathcal{N}(0,~\frac{\omega}{25})
    \end{equation*}
    where $\omega$ is a preset standard deviation with small value. The original 
    bounding box $R = [l, r, t, b]$ is then imposed by $\boldsymbol{\delta}$ to
    obtain a fluctuated rectangle region $R^*$ as
    \begin{equation*}
    \setlength{\abovedisplayskip}{0.1cm}
    \setlength{\belowdisplayskip}{0.1cm}
        R^* = [l^*, r^*, t^*, b^*]
      \label{equ:new_box}
    \end{equation*}
    where $l^*$, $r^*$, $t^*$ and $b^*$ are the new coordinates which are defined as
    \begin{equation*}
    \setlength{\abovedisplayskip}{0.1cm}
    \setlength{\belowdisplayskip}{0.1cm}
      \begin{split}
        l^* & = l + \delta_l, \; \;  t^* = t + \delta_t, \\
        r^* & = (1 + \delta_s)\cdot  (r - l) + l + \delta_l, \\
        b^* & = (1 + \delta_s)\cdot  (b - t) + t + \delta_t.
      \end{split}
    \end{equation*}
    The tracking is then conduct based on $R^*$. This procedure is repeated for $100$
    times for each tracker. Afterwards, the mean $\overline{T}$ and standard deviation
    $T_{std}$ of TSP values are calculated for each frame. Finally, we plot the \emph{TSP
    band}, which is a band changing along with frame index and covers the range
    $[\overline{T} - T_{std}, \overline{T} + T_{std}]$, for every visual tracker. 

\begin{figure}[h]
      \includegraphics[width=0.4\textwidth]{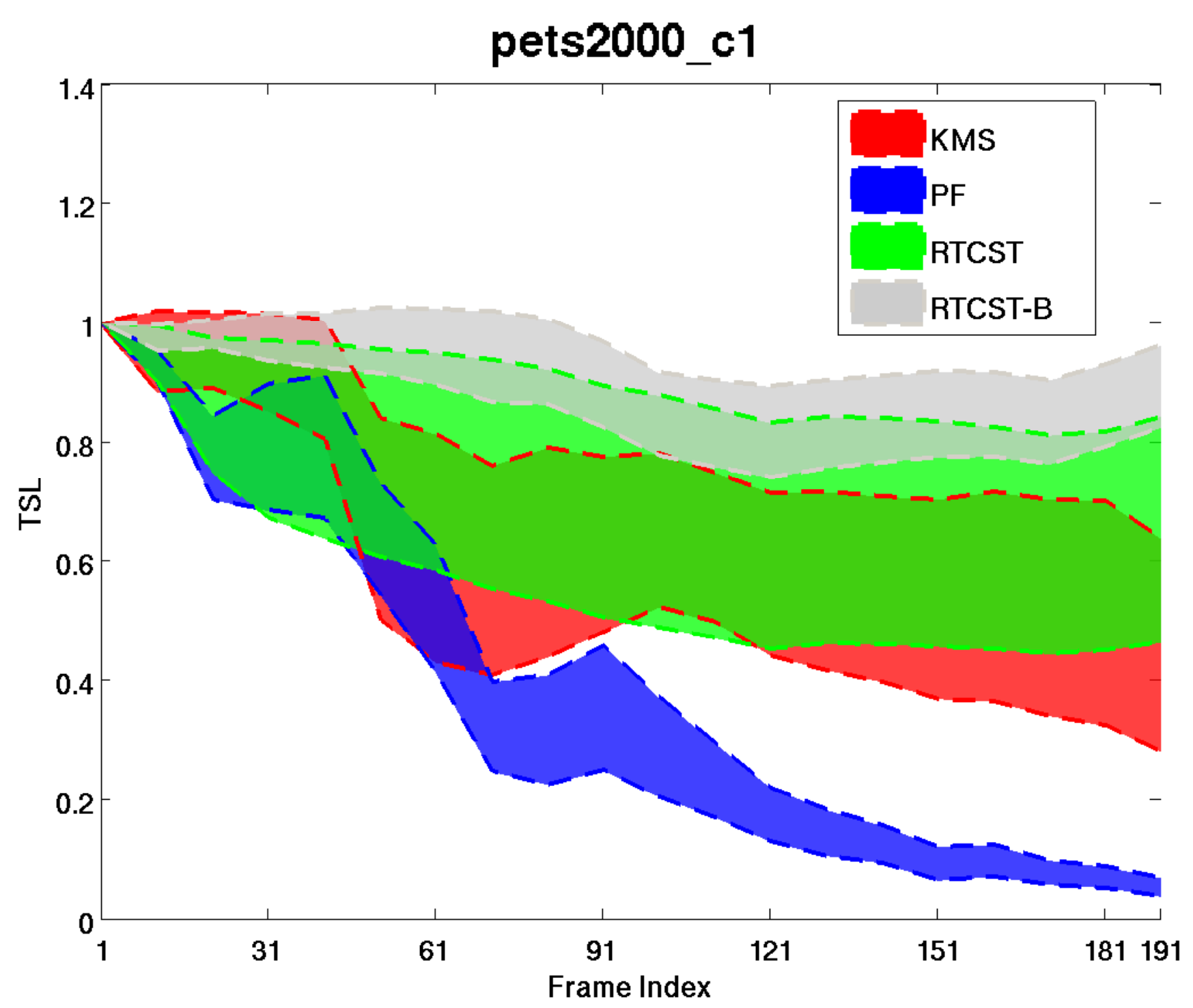}
      \caption{Robustness Verification for visual trackers. The semi-transparent patches
               stand for the TSP bands of trackers. Note that here RTCST and RTCST-B are
               performed with D-$100$ features which is generated via random projection
               and $200$ particles; PF tracker uses $500$ particles.
               }
         \label{fig:robustness}
    \end{figure}

    The new experiment is carried out on video sequence \emph{pets2000\_c1} and
    the \emph{TSP bands} are demonstrated in Figure~\ref{fig:robustness}. 
    An ideal \emph{TSP band} should be with small variance and centered around a
    relatively high mean. We can see that in Figure~\ref{fig:robustness}, RTCST and KMS
    tracker show similar variance but RTCST has a higher TSP mean. PF tracker illustrates
    smaller variance but suffers from very low accuracy. RTCST-B comes with the highest
    average TSP value while still achieves smallest standard deviation. The experiment
    result exhibits the unstable nature of KMS tracker with respect to original target
    position. Meanwhile, it also confirms our conjecture about the presence of high
    robustness when background information is taken into consideration. 

\section{Conclusion and Future Directions}
\label{sec:conclusion}

    In this paper, two enhanced CS-based visual tracking algorithms, namely, RTCST and
    RTCST-B are proposed. A customized OMP algorithm is designed to facilitate the
    proposed tracking algorithms. Hash kernel and random projection are employed to reduce
    the feature dimension of tracking application. In RTCST-B, a CS-based background model
    , which is termed CSBM, is utilized instead of noise templates. The new trackers
    achieves significantly higher efficiency compared with their prototype---the $\ell_1$
    tracker. The remarkable speed growth, which is up to $6271$ times, makes
    CS-based visual trackers qualified for real-time applications. Meanwhile, our methods
    also obtain higher accuracy than off-the-shelf tracking algorithms, \ie, PF tracker
    and KMS tracker. Particularly, RTCST-B achieves consistently highest accuracy and
    robustness thanks to the exploitation of background information. In short words, the
    proposed RTCST and RTCST-B are sufficiently fast for real-time visual tracking and
    more accurate and robust than conventional trackers. 

    For future topics, we believe that one low-hanging fruit is employing the trick mentioned
    in \cite{Tropp_TIT_07_OMP} by Tropp \etal to accelerate the OMP procedure furthermore.
    Another promising direction is to take color information into consideration because in
    many scenarios, color-based classification is more discriminant than the
    intensity-based one. The third direction of future research is treating different
    part of the target, \eg left-top quarter and middle-bottom quarter, as different
    classes. As a result, a multiple classification is conduct within CS framework. The
    obtained likelihood for each particle then becomes a vector comprised of the
    confidences associated with various target parts. Because the time consumptions for
    binary and multiple classification are the same when using CS-based manner, we
    actually obtain more information at the same cost. If we can find a reasonable way to
    exploit the extra information for tracking, more accurate and robust result is likely
    to be obtained.

\footnotesize

\bibliographystyle{IEEEbib}

\end{document}